\documentclass[10pt,twocolumn,letterpaper]{article}

\usepackage{cvpr}
\usepackage{booktabs,multirow,colortbl}
\usepackage[T1]{fontenc}
\usepackage[utf8]{inputenc}
\usepackage[ruled,vlined,linesnumbered]{algorithm2e}
\usepackage{newtxtext,newtxmath}
\usepackage{microtype}
\usepackage[dvipsnames]{xcolor}
\definecolor{vlmblue}{HTML}{1F77B4}
\usepackage{multirow}
\usepackage[table]{xcolor}
\usepackage{graphicx}
\usepackage{booktabs,tabularx,xcolor}
\usepackage{siunitx}
\usepackage[percent]{overpic}
\usepackage{pifont} 
\usepackage{soul}
\newcommand{\R}{\mathbb{R}}

\newcommand{\secref}[1]{Sec.~\ref{#1}}
\newcommand{\tabref}[1]{Tab.~\ref{#1}}
\newcommand{\figref}[1]{Fig.~\ref{#1}}
\newcommand{\benchmark}[1]{\texttt{ViewBench}}
\newcommand{\supp}[1]{\textbf{the supplementary material}}
\newcommand{\inmain}[1]{\textbf{in the main paper}}
\newcommand{\ofmain}[1]{\textbf{of the main paper}}

\usepackage[most]{tcolorbox}
\tcbset{
  colback=blue!5,
  colframe=blue!40,
  boxrule=0.4pt,
  arc=1.5mm,
  left=1mm,right=1mm,top=1mm,bottom=1mm
}

\usepackage{adjustbox}

\usepackage{amsmath,amssymb,mathtools}
\numberwithin{equation}{section}

\definecolor{cvprblue}{rgb}{0.21,0.49,0.74}
\usepackage[pagebackref,breaklinks,colorlinks,allcolors=cvprblue]{hyperref}

\def\confName{CVPR}
\def\confYear{2026}

\title{Token Warping Helps MLLMs Look from Nearby Viewpoints}

\author{
Phillip Y. Lee\textsuperscript{$\ast$} $\;$
Chanho Park\textsuperscript{$\ast$} $\;$
Mingue Park $\;$
Seungwoo Yoo $\;$
Juil Koo $\;$
Minhyuk Sung \\[0.2em]
KAIST
}

\begin{document}

\twocolumn[{
\renewcommand\twocolumn[1][]{#1}
\maketitle
\begin{center}
    \centering
    \captionsetup{type=figure}
    \includegraphics[width=\textwidth]{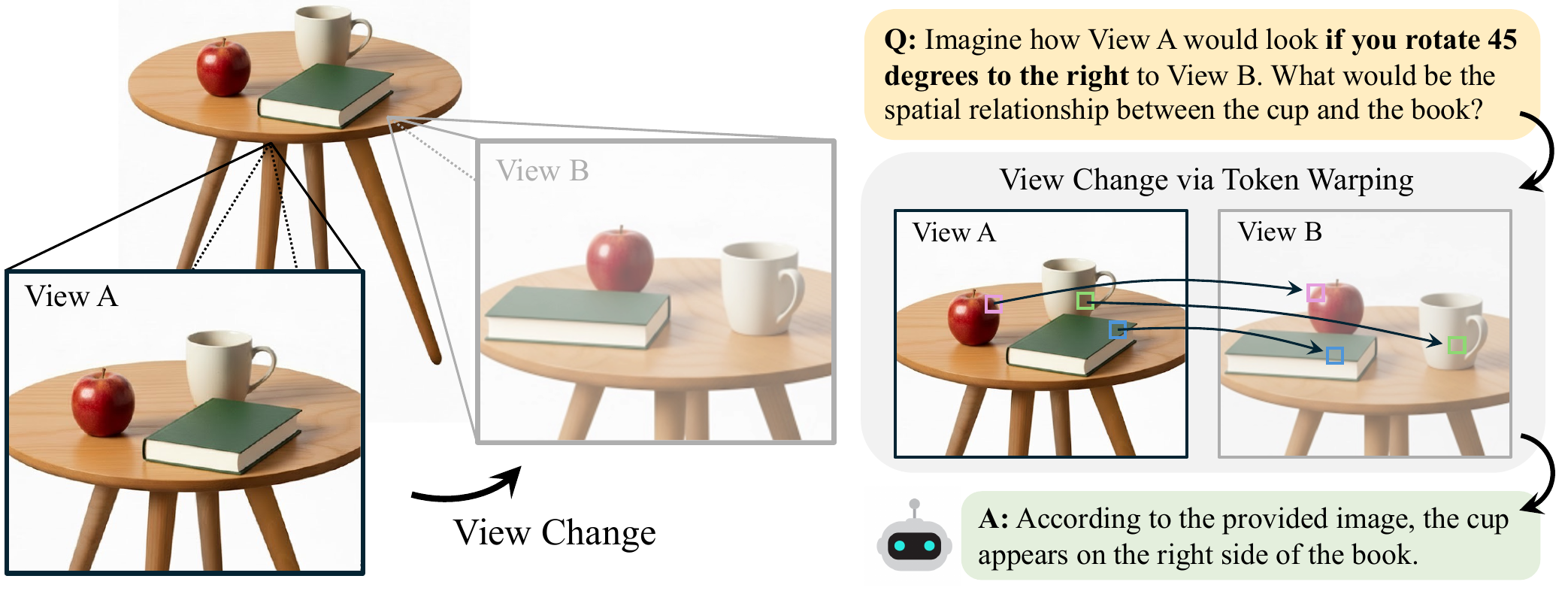}
    \vspace{-\baselineskip}
    \captionof{figure}{\textbf{Viewpoint Change via Token Warping.} We explore token warping as a means of enabling viewpoint changes for MLLMs and find that \emph{backward token warping} can reliably transfer source image content to novel viewpoints without synthesizing new pixels.}
    \label{fig:teaser}
\end{center}
}]

\def\thefootnote{*}\footnotetext{Equal contribution.}
\def\thefootnote{}\footnotetext{Correspondence: Phillip Y. Lee {\tt (phillip0701@kaist.ac.kr)} and Minhyuk Sung {\tt (mhsung@kaist.ac.kr)}}

\maketitle

\begin{abstract}
Can warping tokens, rather than pixels, help multimodal large language models (MLLMs) understand how a scene appears from a nearby viewpoint?
While MLLMs perform well on visual reasoning, they remain fragile to viewpoint changes, as pixel-wise warping is highly sensitive to small depth errors and often introduces geometric distortions. 
Drawing on theories of mental imagery that posit part-level structural representations as the basis for human perspective transformation, we examine whether image tokens in ViT-based MLLMs serve as an effective substrate for viewpoint changes.
We compare forward and backward warping, finding that backward token warping, which defines a dense grid on the target view and retrieves a corresponding source-view token for each grid point, achieves greater stability and better preserves semantic coherence under viewpoint shifts.
Experiments on our proposed~\benchmark{} benchmark demonstrate that token-level warping enables MLLMs to reason reliably from nearby viewpoints, consistently outperforming all baselines including pixel-wise warping approaches, spatially fine-tuned MLLMs, and a generative warping method.
Our project page is at \url{https://token-warping-mllm.github.io/}.
\end{abstract}
\section{Introduction}
A core aspect of spatial reasoning from images is understanding the scene’s three-dimensional structure. Although depth estimation has achieved near-perfect accuracy~\cite{bochkovskii2024depth, yang2024depth}, incorporating predicted depth into MLLMs does not yield genuine 3D understanding. 
Even for simple tasks such as describing the same scene from a different viewpoint (Fig.~\ref{fig:teaser}), MLLMs fine-tuned with explicit 3D supervision~\cite{ma2025_spatialreasoner} show little improvement.
Similar limitations arise in models~\cite{fan2025vlm3r, zheng2025vgllm} that incorporate 3D-aware features~\cite{wang2025continuous, wang2025vggt}, which still struggle to reason about viewpoint transformations.

Recent studies~\cite{lee2025apc, chen2025thinkwith3d, zhang2025spinbench, ramakrishnan2025spatialcog, yin2025mindcube} inspired by mental imagery~\cite{Paivio:1979Imagery, Kosslyn:1978VisualImages, Nanay2021, shepard1971mental, finke1989principles, Tolman1948CognitiveMI, hinton1979mental} suggest that perspective reasoning requires generating a virtual internal representation through explicit transformation. For instance, Lee~\etal{}~\cite{lee2025apc} model a scene using object-centric abstract representations and apply geometric transformations to them. While effective for object-level relational reasoning, such approaches often fail to capture fine-grained details and overall spatial coherence of the scene.

Classical research on mental imagery, from Shepard~\cite{shepard1971mental} to Minsky~\cite{minsky1974framework}, Pylyshyn~\cite{pylyshyn1973mind}, and Hinton~\cite{hinton1979mental}, proposes that mental images rely on structural descriptions defined at the \emph{part level} rather than at the holistic object level. From this perspective, the evolution of computer vision can be interpreted as the pursuit of machine-perceivable, part-level representations, which have recently converged in the form of \emph{image tokens} used by Transformer architectures~\cite{vaswani2017attention, dosovitskiy2021vit}. It is therefore natural to extend the concept of mental imagery to these perceptual atomic units rather than to object-level abstractions.

Motivated by this insight, we investigate whether transformations applied to image tokens can generate consistent internal representations of scenes under viewpoint changes, thereby improving spatial reasoning. We find that this is indeed the case. Unlike pixel-level warping, which amplifies even small depth errors into severe distortions, token-level transformations remain robust to geometric noise and yield more coherent viewpoint reasoning.

To systematically verify our hypothesis that image tokens form a robust substrate for viewpoint transformation, we first examine how sensitive recent MLLMs are to noise introduced during local patch retrieval. For each image token, we begin with the regular grid centers but intentionally fetch the corresponding image patch from a \emph{slightly perturbed} center position. By gradually increasing the perturbation magnitude, even to the point where the offset approaches the size of the patch, we observe that MLLMs remain surprisingly stable in their ability to recognize the underlying image content. This suggests that MLLMs are inherently tolerant to spatial noise during patch formation, providing strong evidence that when constructing image tokens from a different viewpoint using a predicted (and potentially imperfect) depth map, the geometric noise introduced during warping does not significantly undermine the model’s visual understanding.

Next, we investigate how to best implement token-level warping under viewpoint changes. Given an input image with its depth map and a target camera pose, there are two possible transformation strategies: \emph{forward} warping and \emph{backward} warping. In the forward approach, we first construct the image tokens from the input view and then map each token to the target viewpoint. In contrast, the backward approach begins by taking the regular grid centers of the target view and mapping each center back to the input image. Within backward warping, we consider two variants. The first, \emph{nearest fetching}, constructs all image tokens only once on the input view and then assigns to each mapped target location the nearest precomputed token. The second, \emph{adaptive fetching}, directly re-patchifies the input image at each mapped location by treating it as the patch center, rather than assigning the nearest precomputed token.

Through our experiments on~\benchmark{}, designed to evaluate MLLMs on spatial reasoning tasks involving viewpoint changes, we systematically explore the aforementioned axes of pipeline design. The results show that both the choice of representation to warp and the specific warping mechanism have substantial effects on performance. In particular, we find that backward token warping, which preserves dense and regularly spaced grids in the target view, outperforms all other variants. Remarkably, this approach, which incurs only minimal inference-time computation for warping, surpasses state-of-the-art specialist MLLMs fine-tuned on spatial reasoning datasets, as well as a generative warping technique that employs a camera-conditioned diffusion model to directly synthesize the target-view image.
\begin{figure}[t!]
  \centering
  \includegraphics[width=\linewidth]{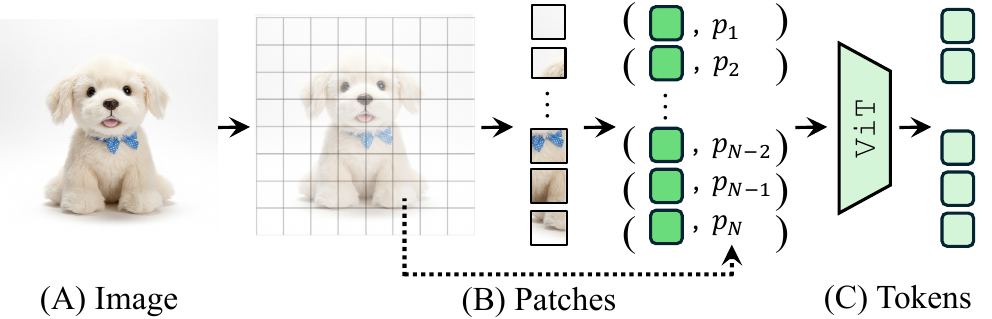}
  \vspace{-1.5\baselineskip}
  \caption{\textbf{Image Tokenization in MLLMs (Sec.~\ref{subsec:image_tokenization}).} 
  MLLMs process images by dividing them into fixed-size patches, embedding each patch, and passing them through a vision encoder (\eg, ViT) to obtain image tokens.}
  \label{fig:image_tok}
  \vspace{-1.5\baselineskip}
\end{figure}

\section{Related Work}
\subsection{Spatial Understanding in MLLMs}
\label{subsec:related_mllm_spatial_reasoning}
\vspace{-0.25\baselineskip}
The potential of multimodal LLMs (MLLMs) on real-world embodied tasks have sparked research interest on their spatial reasoning abilities~\cite{driess2023palm, mu2023embodiedgpt, zhang2025embodied, huang2023leo, hu2025_3dllmmem}.
Rich line of benchmarks and evaluation protocols pointed out that MLLMs often struggle at even basic spatial understanding~\cite{ma20253dsrbench, daxberger2025mmspatial, fu2024blinkmultimodallargelanguage, rahmanzadehgervi2024vision, ramakrishnan2025spatialcog, tang2024sparkle, wang2024picture, yu2025far}, and showed that their spatial cognition can be improved by well-curated data~\cite{chen2024spatialvlm, deitke2025molmo, li2024llavaonevision, liu2025spatialcot, song2024robospatial, ji2025robobrain}, introducing novel architecture designs~\cite{tong2024cambrian, ma2025spatialllm} and carefully designed training frameworks~\cite{wang2025ross3d, ma2025_spatialreasoner, li2025spatialladder, ouyang2025spacer}.
Another line of work suggest that integrating rich structural priors (\eg, depth maps~\cite{cai2025spatialbot}, segmentation masks~\cite{cheng2024spatialrgpt}, point clouds~\cite{hong2023_3dllm, deng2025_3dllava, chen2024ll3da}, or rich features from foundation models~\cite{fan2025vlm3r, huang20253drs, wu2025spatialmllm, zheng2025vgllm}) can assist MLLM's spatial reasoning on image, video and 3D inputs.
This can be implemented in either by training auxiliary encoders to project new modalities into the model~\cite{huang20253dr1, yuan2025_scener1, yu2025_inst3dllm}, or by designing novel prompting mechanisms~\cite{zhu2025struct2d, zhang2025spatialmind, qi2025gpt4scene, li2025see}.
Multiple works integrate 3D-aware features or positional embeddings into 2D MLLMs to enhance their 3D understanding~\cite{fu2025_scenellm, cheng2025sr3d, zhu2024llava3d, thai2025splattalk, zheng2025video3dllm}.
Moreover, other works focus on the LLM's reasoning skills, building agentic frameworks that tackle spatial tasks via program-like decomposition~\cite{marsili2025visual, ma2024spatialpin} or test-time scaling algorithms~\cite{yang2025mindjourney}.

\subsection{Viewpoint-Aware Reasoning}
\label{subsec:related_viewpoint_change}
\vspace{-0.25\baselineskip}
As MLLMs increasingly serve as the \emph{brains} of autonomous agents in open environments~\cite{qi2025vln, zhang2025embodied, yang2025embodiedbench, ferrag2025llm, wang2025vagen, ni2025embodied}, recent research has begun to examine their ability to handle \emph{viewpoint-aware} perception and cognition~\cite{zhang2024vision, song2024robospatial, lee2025apc, manh2025mind}.
Notably, COMFORT~\cite{zhang2024vision} draws on cognitive studies about \emph{frame of reference} for perspective-taking and shows that MLLMs are largely confined to the input camera's viewpoint. They struggle to adopt another person's or object's vantage point within the same scene, considered a core human cognitive skill.
Related works further propose finer-grained evaluation criteria~\cite{li2025viewspatial, zhang2025spinbench, goral2024seeingeyesevaluatingvisual, zhang2024sphere, ma20253dsrbench, linsley20243dpcbenchmarkvisualperspective} and suggest plug-in strategies inspired by human cognitive process to scaffold viewpoint reasoning~\cite{lee2025apc}.
When provided denser observations, either as multi-view images~\cite{yang2025mmsibench, zhang2025sparbench, xu2025_multispatialmllm, chen2025thinkwith3d} or videos~\cite{yang2025thinkinginspace, azuma2022scanqa, ma2022sqa3d, zhang2024llavanextvideo}, it is also essential to interpret the scene from a specific viewpoint (\eg, one of the frames).
For this, Mindcube~\cite{yin2025mindcube} generates a simple cognitive map to grasp the holistic structure of the scene, while ViLaSR~\cite{wu2025vilasr} uses drawing as a tool for reasoning in space.
We ask a new question: given a single image, can an MLLM \emph{look} from a nearby viewpoint? We investigate this by warping tokens, rather than synthesizing pixels or auxiliary data, to simulate viewpoint shifts efficiently and robustly.

\begin{figure}[b!]
  \centering
  \vspace{-0.5\baselineskip}
  \includegraphics[width=\linewidth]{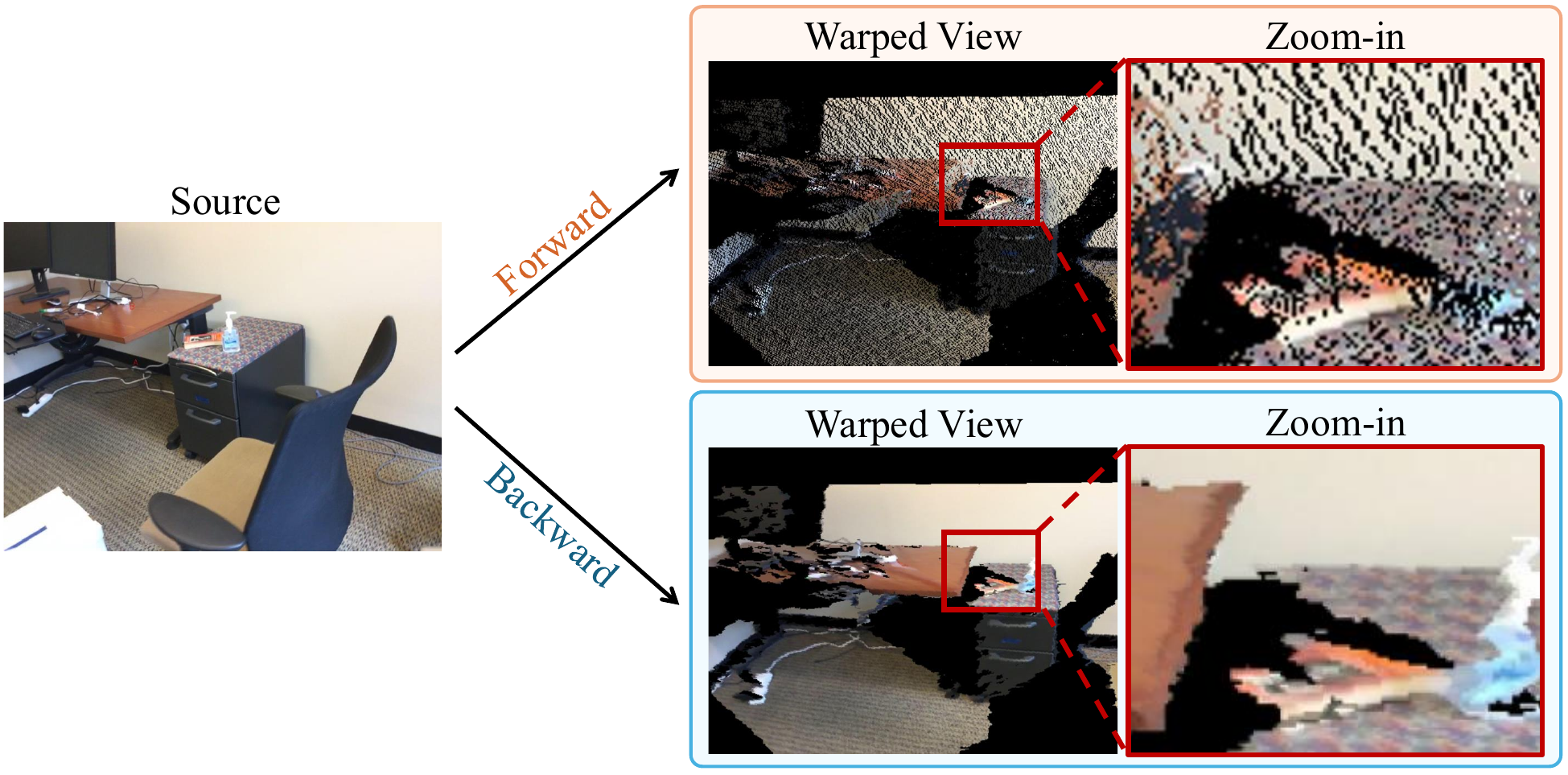}
  \vspace{-1.5\baselineskip}
  \caption{\textbf{Limitations of Pixel-Wise Warping.}  
    Pixel-wise warping to a target viewpoint often introduces local distortions and semantic degradation. In both \emph{forward} (top) and \emph{backward} (bottom) warping, the book from the source view appears significantly distorted after transformation (in the red box).}
  \label{fig:failture_modes_pixel}
  \vspace{-0.5\baselineskip}
\end{figure}

\subsection{Image as Tokens}
\label{subsec:related_image_tokenization}
\vspace{-0.25\baselineskip}
Since the introduction of Vision Transformers (ViT)~\cite{vaswani2017attention, dosovitskiy2021vit}, it has become standard to divide images into patch-wise \emph{tokens} as inputs to transformer-based vision models. 
Tokens serve as \emph{semantic primitives} that support both local detail and global context understanding, driving strong performance across computer vision tasks including classification~\cite{oquab2023dinov2}, detection~\cite{minderer2022simple, kim2023region}, segmentation~\cite{kirillov2023segment, ravi2024sam}, 3D reconstruction~\cite{wang2024dust3r, wang2025vggt}, multimodal understanding~\cite{liu2023visual}, and generation~\cite{rombach2022high, peebles2023scalable, flux2024}.
Building on this foundation, recent work explores deformable~\cite{xia2022vision} and adaptive~\cite{ronen2023vision, xia2022vision, rao2021dynamicvit, choudhury2025accelerating, chen2024subobject} tokenization techniques for improving semantic alignment and efficiency.
Others leverage tokens for image/video generation~\cite{li2023gligen, bai2025positional, lee2024groundit, qu2025tokenflow}, editing~\cite{geyer2023tokenflow, koo2025videohandles, xu2024headrouter}, or perception~\cite{fan2025vlm3r, yu2025introducing, bigverdi2025perception, lee2025molmoact} by introducing richer token types or directly manipulating tokens to steer model behavior.
In this work, we focus on the role of tokens as primary semantic units in MLLMs, and propose token warping as a lightweight and robust strategy to enable \emph{viewpoint-aware perception}.

\begin{figure*}[t!]
  \centering
  \includegraphics[width=\linewidth]{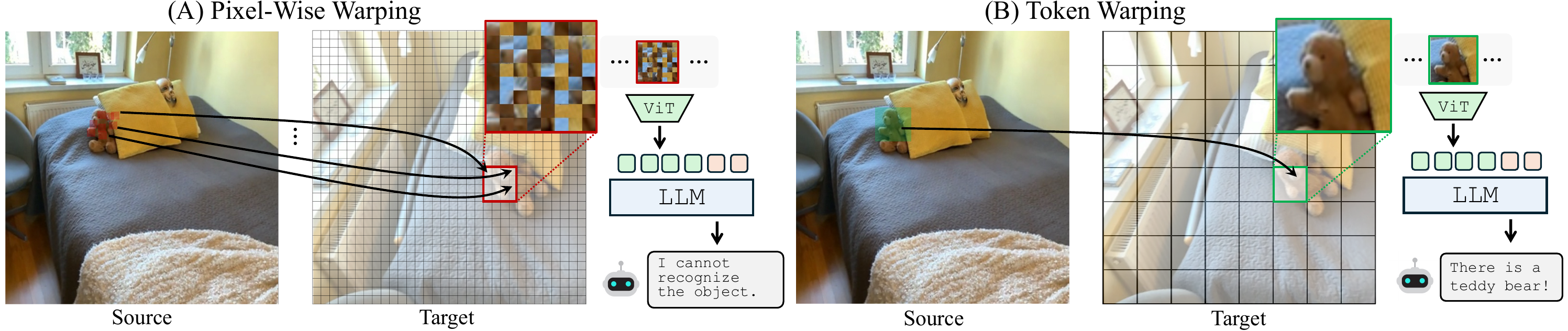}
  \vspace{-\baselineskip}
  \caption{\textbf{Pixel-Wise vs.\ Token Warping.}  
    Comparison of inverse warping strategies (Sec.~\ref{sec:token_warping}). 
    (A) \emph{Pixel-wise warping} retrieves pixels for each target coordinate, but patchifying the warped image introduces local distortions, resulting in degraded MLLM understanding. 
    (B) \emph{Token warping} directly retrieves intact tokens (or patches) from the source view, preserving semantics and improving viewpoint-aware perception.}
  \label{fig:method}
  \vspace{-0.5\baselineskip}
\end{figure*}

\section{Token Warping for Viewpoint Changes}
\label{sec:method_problem_definition}
Modern ViT-based MLLMs represent an image as a sequence of tokens obtained by dividing it into patches and embedding each into a latent vector.
These image tokens function as \emph{perceptual atoms} of the MLLM: localized, semantically meaningful units processed jointly with positional embeddings.
Inspired by cognitive theories of mental imagery~\cite{hinton1979mental, minsky1974framework, shepard1971mental, pylyshyn1973mind}, we investigate whether image tokens provide the appropriate part-level granularity for performing viewpoint transformations.
Object-level representations~\cite{lee2025apc} are too coarse, sacrificing important spatial and appearance details, while pixel-level representations are too fine-grained and sensitive to even small depth or geometric noise during warping (see Fig.~\ref{fig:failture_modes_pixel}). 
\emph{Image tokens} lie between these extremes, retaining rich visual detail while remaining robust to local perturbations.
We therefore posit that image tokens serve as an effective perceptual substrate for neural mental imagery and viewpoint transformation.

A key requirement for enabling such viewpoint transformations is robustness to positional perturbations introduced during patch retrieval, since even state-of-the-art depth estimation contains small errors that can cause significant distortion when pixels are warped directly.
To assess this, in Sec.~\ref{sec:toy}, we evaluate MLLM's sensitivity to retrieval-position noise by perturbing the regular grid center points used to fetch local patches. 
Specifically, we retrieve each patch from a slightly shifted center position, introducing a controlled offset during patch extraction.
This experiment reveals that image tokens are robust to positional noise, making them well suited for reliable geometric transformation under viewpoint changes.

Building on this insight, we search for the best token-level warping strategy in~\secref{sec:token_warping} by exploring several warping functions and analyzing how well each preserves structural coherence and semantic consistency under viewpoint shifts.

\subsection{Image Tokenization in MLLMs}
\label{subsec:image_tokenization}
In MLLMs, an image $\mathbf{I}$ is partitioned into a fixed, non-overlapping grid of patches $\left\{\mathbf{u}_i\right\}_{i=1}^{M}$ (\figref{fig:image_tok}).
Each patch $\mathbf{u}_i \in \mathbb{R}^{l \times l \times 3}$ corresponds to a square region of $\mathbf{I}$ associated with a grid-center coordinate $\mathbf{c}_i = (x_i, y_i)$ on $\mathbf{I}$'s lattice.
A shallow encoder $\mathcal{E}$ maps each patch to an embedding $\mathbf{e}_i = \mathcal{E} (\mathbf{u}_i)$.
These embeddings, together with their grid-center coordinates, are processed by a vision encoder $\mathcal{V}$ (\eg, ViT~\cite{vaswani2017attention, dosovitskiy2021vit}) to produce image tokens $\{ \mathbf{v}_i \}^{M}_{i=1} = \mathcal{V} \!\left( \{ (\mathbf{e}_i, \mathbf{c}_i) \}^{M}_{i=1} \right)$, which are then projected into the LLM’s latent space and processed alongside text tokens.
Notably, each token carries not only semantic information encoded from its pixel values but also positional information defined at the patch level as a whole. 
We hypothesize that transferring tokens rather than individual pixels is therefore more robust to noise in positional information, as we empirically demonstrate below.

\begin{figure}[b!]
  \centering
  \vspace{-\baselineskip}
  \includegraphics[width=\linewidth]{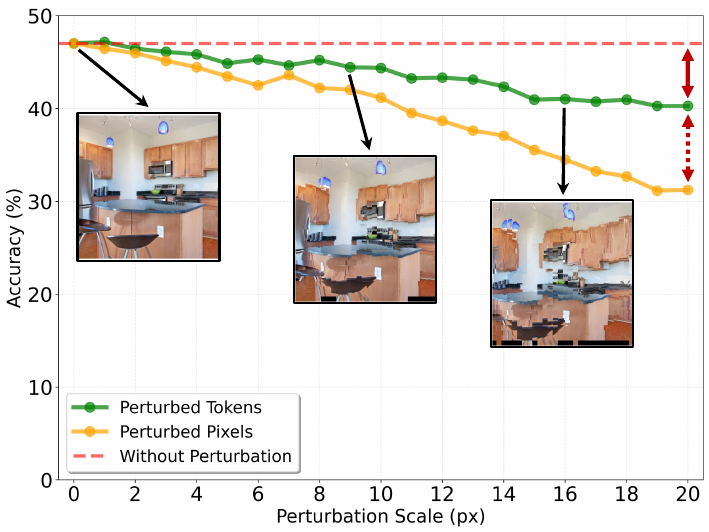}
  \vspace{-1.5\baselineskip}
  \caption{\textbf{Fetching Position Noise Sensitivity (Sec.~\ref{sec:toy}).}
    Through a toy experiment on CV-Bench-2D~\cite{tong2024cambrian}, where we emulate local positional perturbations and degradation introduced by warping, we find that token representations in MLLMs are highly robust to noise in the image positions from which tokens are fetched. This suggests that tokens are well suited for representing viewpoint changes.
    }
  \label{fig:toy_exp}
  \vspace{-0.5\baselineskip}
\end{figure}

\subsection{Fetching Position Noise Sensitivity Test}
\label{sec:toy}
\vspace{-0.25\baselineskip}
As hypothesized earlier, image tokens serve as perceptual atoms in MLLMs well suited for simulating viewpoint changes through warping: they naturally encode locality-aware features and propagate as coherent units during the warping operation.

To demonstrate this, we begin with a simple proof-of-concept experiment that perturbs the positional information of MLLM tokens via jittering. 
Further comparisons against pixel-based representations in actual viewpoint change scenarios are presented in~\secref{sec:evaluation}.
Specifically, consider each token $\mathbf{v}_i$ from image $\mathbf{I}$ together with its grid-center coordinate $\mathbf{c}_i$, which determines its positional embedding.
For each token, we sample a displacement vector $\mathbf{u}_i = (\Delta x_i, \Delta y_i)$ from standard Gaussian distribution and apply mean-filter smoothing over neighboring cells.
We then normalize all $\mathbf{u}_i$ by the global maximum magnitude and scale by a hyperparameter, the \emph{maximum displacement value}.
We vary this value from 0.0 to 20.0 and fix the smoothing neighborhood to 9 grid cells. This procedure is designed to emulate the noisy positional perturbations introduced during warping.
As a pixel-level baseline, we apply the same jittering process and add slight pixel-wise perturbation (\ie, 10\% of each maximum displacement value) to emulate pixel-level perturbations in pixel-wise warping.

Fig.~\ref{fig:toy_exp} shows Qwen2.5-VL's~\cite{bai2025qwen25vl} accuracy on CV-Bench-2D~\cite{tong2024cambrian} VQA tasks under varying maximum displacement values for token position perturbations (green plot).
The model maintains consistent performance across perturbation levels from 0 to 20.0. Notably, it exhibits only mild degradation in the large-perturbation regime (19.0-20.0 pixels), where the perturbation artifacts become visually apparent (top-right example in~\figref{fig:toy_exp}).
Compared with the pixel-level baseline (orange), token-level representations are clearly more robust under similar level of perturbations.
This result highlights the importance of preserving localized, semantically meaningful visual elements in perceptual tasks, consistent with classical discussions on part-level structures in mental imagery~\cite{hinton1979mental, shepard1971mental, minsky1974framework, pylyshyn1973mind}. 
Motivated by this finding, we adopt \emph{tokens} as the units during warping, as detailed in the following section.

\begin{figure*}[t!]
  \centering
  \includegraphics[width=\linewidth]{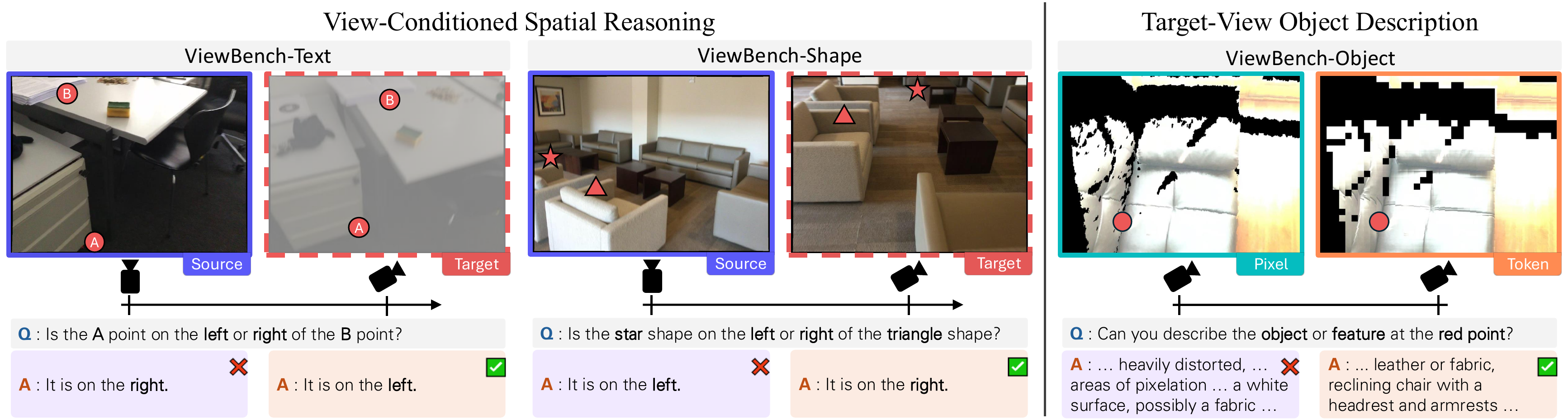}
  \vspace{-\baselineskip}
  \caption{\textbf{\benchmark{}.} Example source-target image pairs with corresponding questions and answers from our~\benchmark{} benchmark.
  The tasks evaluate MLLM's ability to infer spatial relationships from nearby viewpoints (\texttt{Text}, \texttt{Shape}), while also measuring robustness to view changes by asking to describe object properties visible in the warped target view (\texttt{Object}).}
  \label{fig:benchmark}
  \vspace{-0.5\baselineskip}
\end{figure*}

\subsection{Designing Token Warping Functions}
\label{sec:token_warping}
Building on our observation regarding the robustness of tokens, we now turn to a spatial reasoning task involving two viewpoints. In this setting, the model is given an observed~\emph{source} viewpoint and an unobserved~\emph{target} viewpoint, together with a question that requires imagining how the scene would appear from the target viewpoint in order to answer.
Formally, let $\mathbf{I} \in \mathbb{R}^{H \times W \times 3}$ denote the RGB image captured from the~\emph{source} viewpoint with camera pose matrix $\Pi_S \in \mathbb{R}^{4 \times 4}$, representing the world-to-camera transformation.
The question $Q$ is a natural language query about the scene depicted in $\mathbf{I}$, but posed from the perspective of a~\emph{target} viewpoint with camera pose $\Pi_T \in \mathbb{R}^{4 \times 4}$. 
We further assume that a depth map $\mathbf{D} \in \mathbb{R}^{H \times W \times 1}$ corresponding to $\mathbf{I}$ is available, either as ground truth or estimated via monocular depth estimation~\cite{yang2024depth}, along with the intrinsic matrix $\mathbf{K} \in \mathbb{R}^{4 \times 4}$.

Given the above, the most direct strategy for answering $Q$ is to~\emph{warp} the source image $\mathbf{I}$, along with the tokens encoded from it, into the target viewpoint using the depth map $\mathbf{D}$, the intrinsic matrix $\mathbf{K}$, and the relative pose $\Pi_{S \rightarrow T} = \Pi_T \Pi_S^{-1}$.
Let $\mathbf{c} \in \mathbb{R}^{(HW) \times 2}$ denote the grid-center coordinates of $\mathbf{I}$. The corresponding coordinates after warping, $\mathbf{c}^{*} \in \mathbb{R}^{(HW) \times 2}$, are computed as:
\begin{align}
    \mathbf{c}^{*} = f_{S \rightarrow T} (\mathbf{c}, \Pi_{S \rightarrow T}, \mathbf{K}, \mathbf{D}),
    \label{eq:forward_warp_func}
\end{align}
where $f_{S \rightarrow T}: \mathbb{R}^{(HW) \times 2} \rightarrow \mathbb{R}^{(HW) \times 2}$ denotes the forward-warping function that projects token positions from the source to the target viewpoint.
Conversely, we can define the backward mapping $f_{T \rightarrow S}$, which takes grid-center coordinates at the \emph{target} viewpoint and computes their corresponding coordinates on the \emph{source} image plane.

In this work, we explore both as candidates for token warping: either through direct forward projection ($f_{S \rightarrow T}$) or by fetching corresponding source tokens via backward projection ($f_{T \rightarrow S}$).
Beyond these two approaches for determining~\emph{which} coordinates to fetch, we further investigate~\emph{how} to fetch them, considering both nearest and adaptive fetching strategies.

\vspace{-0.5\baselineskip}
\paragraph{Forward vs.\ Backward Warping.}
\emph{Forward warping} projects tokens from $\mathbf{I}$ into the target viewpoint via $f_{S\rightarrow T}$ and computes their positional embeddings accordingly.
Despite its simplicity, this approach often yields irregular, sparse token distributions with large holes across the target image plane.
As we later show in~\secref{sec:evaluation}, such irregular and sparsely placed tokens are out-of-distribution inputs for an MLLM trained on dense, regularly spaced token grids, leading to substantial performance degradation.
\emph{Backward warping} takes the opposite strategy: we first define a dense, regular grid in the target view and retrieve the corresponding tokens from $\mathbf{I}$ via the mapping $f_{T \rightarrow S}$.
For this, we build a lightweight 3D proxy mesh from the source image's depth map and compute the mapping from each target grid to the source via ray casting. Implementation details are provided in~\supp{}.
Unlike forward warping, this approach produces tokens that are, by construction, regularly placed on the target image plane.
We thus adopt backward warping as our primary strategy, which consistently outperforms forward warping in our experiments (\secref{sec:evaluation}).
\begin{figure}[b!]
  \centering
  \includegraphics[width=\linewidth]{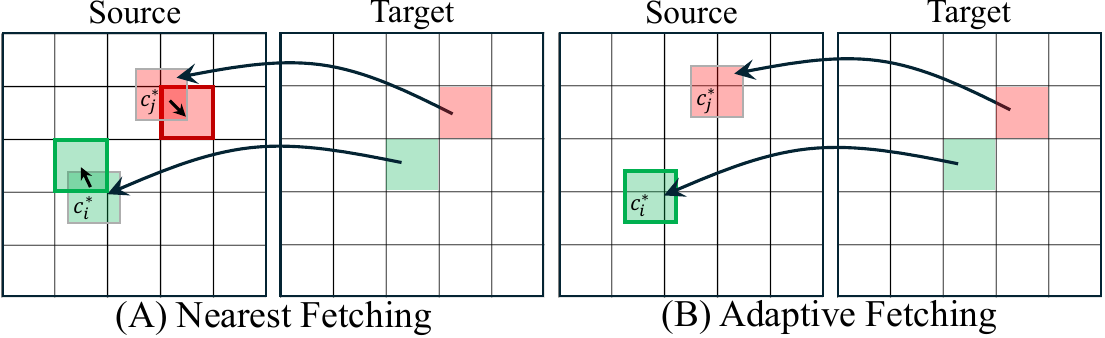}
  \vspace{-1.25\baselineskip}
  \caption{\textbf{Token Fetching Strategies.} (A) \emph{Nearest fetching} selects the closest existing token from the source image grid. (B) \emph{Adaptive fetching} dynamically crops a patch centered at the mapped coordinate to derive a token precisely centered at the target location.}
  \label{fig:inv_token_fetching}
  \vspace{-0.5\baselineskip}
\end{figure}

\begin{table*}[t!]
\smaller
\centering
\setlength{\tabcolsep}{3pt}
\begin{adjustbox}{width=\textwidth}
\begin{tabular}{l*{18}{c}@{}}
\toprule
 &
\multicolumn{6}{c}{\textbf{\texttt{ViewBench-Text} (\%)}} &
\multicolumn{6}{c}{\textbf{\texttt{ViewBench-Shape} (\%)}} &
\multicolumn{6}{c}{\textbf{\texttt{ViewBench-Object} (1-10)}} \\
\cmidrule(lr){2-7}\cmidrule(lr){8-13}\cmidrule(lr){14-19}
View Overlap (\%) 
& \multicolumn{2}{c}{5–15} & \multicolumn{2}{c}{15–25} & \multicolumn{2}{c}{25–35}
& \multicolumn{2}{c}{5–15} & \multicolumn{2}{c}{15–25} & \multicolumn{2}{c}{25–35}
& \multicolumn{2}{c}{5–15} & \multicolumn{2}{c}{15–25} & \multicolumn{2}{c}{25–35} \\
\cmidrule(lr){1-1} \cmidrule(lr){2-3}\cmidrule(lr){4-5}\cmidrule(lr){6-7}
\cmidrule(lr){8-9}\cmidrule(lr){10-11}\cmidrule(lr){12-13}
\cmidrule(lr){14-15}\cmidrule(lr){16-17}\cmidrule(lr){18-19}
Depth 
& GT & Pred. & GT & Pred. & GT & Pred.
& GT & Pred. & GT & Pred. & GT & Pred.
& GT & Pred. & GT & Pred. & GT & Pred. \\
\midrule

Target View (Oracle)
  & 100.00 & --& 100.00 & --& 100.00 & -
  & 100.00 & --& 100.00 & --& 100.00 & -
  & 6.64 & --& 7.31 & --& 7.43 & --\\

\midrule
\emph{Specialist MLLMs} &&&&&&&&&&&&&&&&&& \\

\cellcolor{blue!10} SpatialReasoner~\cite{ma2025_spatialreasoner}
  & 46.73 & --& 53.30 & --& 53.71 & --
  & 33.72 & --& 38.27 & --& 48.15 & --
  & --& --& --& --& --& --\\

\cellcolor{blue!10} VLM-3R~\cite{fan2025vlm3r}
  & 63.82 & --& 70.56 & --& 60.57 & --
  & 49.22 & --& 49.79 & --& 50.21 & --
  & --& --& --& --& --& --\\

\cellcolor{blue!10} ViLaSR~\cite{wu2025vilasr}
  & 44.22 & --& 52.28 & --& 48.00 & --
  & 22.87 & --& 23.05 & --& 34.57 & --
  & --& --& --& --& --& --\\

\cellcolor{blue!10} Qwen2.5-VL~\cite{bai2025qwen25vl}
  & 46.23 & --& 59.39 & --& 52.00 & --
  & 24.42 & --& 25.10 & --& 37.86 & --
  & --& --& --& --& --& --\\

\midrule
\emph{Novel View Synthesis} &&&&&&&&&&&&&&&&&& \\

\cellcolor{ForestGreen!15} GenWarp~\cite{seo2024genwarp}
  & 69.35 & --& 71.07 & --& 66.29 & --
  & 53.10 & --& 47.33 & --& 55.14 & --
  & 4.32 & --& 4.81 & --& 4.34 & --\\

\midrule
\emph{Pixel-Wise Warping} &&&&&&&&&&&&&&&&&& \\

\cellcolor{RubineRed!15} Forward
  & 70.85 & 69.35 & 73.60 & 73.10 & 62.86 & 67.43
  & 56.20 & 56.20 & 56.79 & 56.79 & 60.49 & 60.08
  & 3.22 & 3.22 & 4.04 & 3.87 & 4.78 & 4.54 \\

\cellcolor{RubineRed!15} Backward
  & 71.86 & 67.84 & 75.63 & 74.62 & 68.57 & 68.57
  & 62.40 & 58.14 & 58.02 & 56.79 & 66.67 & 64.20
  & 4.53 & 4.45 & \underline{5.52} & 5.48 & 5.94 & 5.89 \\

\midrule
\emph{Token Warping} &&&&&&&&&&&&&&&&&& \\

\cellcolor{Gray!15} Forward
  & 60.30 & 66.83 & 64.47 & 65.48 & 54.86 & 60.57
  & 55.04 & 56.98 & 55.14 & 60.91 & 53.09 & 56.38
  & 4.09 & 4.20 & 4.27 & 4.37 & 4.07 & 3.78 \\

\cellcolor{Gray!15} Backward-Nearest
  & \underline{74.87} & \textbf{75.38} & \textbf{80.71} & \textbf{81.73} & \underline{74.86} & \textbf{76.00}
  & \textbf{67.44} & \underline{63.95} & \underline{62.96} & \textbf{62.55} & \underline{73.25} & \textbf{75.31}
  & \underline{4.80} & \underline{4.86} & 5.39 & \underline{5.57} & \textbf{6.19} & \underline{5.97} \\

\cellcolor{Gray!15} Backward-Adaptive
  & \textbf{77.89} & \underline{73.37} & \underline{79.70} & \underline{80.71} & \textbf{78.86} & \underline{74.29}
  & \textbf{67.44} & \textbf{66.28} & \textbf{66.26} & \underline{61.32} & \textbf{75.72} & \underline{70.37}
  & \textbf{4.97} & \textbf{5.18} & \textbf{5.76} & \textbf{6.29} & \underline{6.11} & \textbf{6.14} \\

\bottomrule
\end{tabular}
\end{adjustbox}

\caption{\textbf{Quantitative Comparisons on~\benchmark{}.} 
The prediction accuracies of the models on the spatial reasoning tasks (\texttt{\benchmark{}-Text} and \texttt{\benchmark{}-Shape}) are reported in columns 2–13. The performance scores for the target-view object description task (\texttt{\benchmark{}-Object}), evaluated by Qwen2.5-VL 14B~\cite{bai2025qwen25vl} on a 1–10 scale, are summarized in columns 14–19. Across all tasks and setups, backward token-wise warping achieves the best performance.}
\label{tab:viewbench_real_gt_pred_pairs}
\vspace{-0.5\baselineskip}
\end{table*}

\paragraph{Nearest vs.\ Adaptive Fetching.}
A further design consideration is how to fetch tokens from the coordinates produced by $f_{T \rightarrow S}$, as these coordinates often fall between token grid centers on the source image plane.
Recall that each token originates from a fixed-grid patch (Sec.~\ref{subsec:image_tokenization}).
We explore two strategies to address this gap: nearest and adaptive fetching.
In \emph{nearest fetching}, given a mapped coordinate $\mathbf{c}^{*}_i$ from Eq.~\ref{eq:forward_warp_func}, we retrieve the token associated with the nearest grid-center point in Euclidean distance (\figref{fig:inv_token_fetching}-(A)).
In \emph{adaptive fetching}, the source image $\mathbf{I}$ is re-patchified according to the warped coordinates: for each $\mathbf{c}^{*}_i$, a patch centered at $\mathbf{c}^{*}_i$ is cropped and encoded through the same token encoding process shown in~\figref{fig:image_tok}.
This allows tokens to be centered at arbitrary locations beyond the constraints of a fixed patch grid.
\figref{fig:inv_token_fetching} provides a visual comparison of the two fetching strategies, and algorithmic details are provided in~\supp{}.
In our experiments (\secref{sec:evaluation}), we find that nearest fetching performs comparably to adaptive fetching, despite the latter requiring additional computation for re-patchification.

\definecolor{correctgreen}{RGB}{26,187,46}
\definecolor{wrongred}{RGB}{243,1,3}

\newcommand{\imgframe}[2]{
\begingroup \setlength{\fboxsep}{0pt}
\setlength{\fboxrule}{1.3pt}
\fcolorbox{#1}{white}{\includegraphics[width=1.0\linewidth,keepaspectratio]{#2}}
\endgroup }

\newcommand{\hcell}[1]{\makecell{\scriptsize #1}}
\begin{figure*}[t!]
\scriptsize
\centering
\renewcommand{\arraystretch}{1.0}
\setlength{\tabcolsep}{1pt}
\arrayrulecolor{black}
\setlength{\arrayrulewidth}{0.2pt}

\newcommand{\receptacle}[1]{
  \begingroup
  \setlength{\fboxsep}{2pt}
  \colorbox{olive!15}{#1}
  \endgroup
}
\newcommand{\targetobj}[1]{
  \begingroup
  \setlength{\fboxsep}{2pt}
  \colorbox{blue!15}{#1}
  \endgroup
}

\newcommand{\wrongimg}[2][]{
  \begin{overpic}[#1]{#2}
    \put(3,75){\color{wrongred}\fontsize{15}{15}\selectfont\bfseries\xmark}
  \end{overpic}
}

\newcommand{\correctimg}[2][]{
  \begin{overpic}[#1]{#2}
    \put(3,75){\color{correctgreen}\fontsize{15}{15}\selectfont\bfseries\cmark}
  \end{overpic}
}

\begin{tabularx}{\linewidth}{
    | >{\centering\arraybackslash}p{2.1cm} 
    | >{\centering\arraybackslash}p{2.1cm} 
    | >{\centering\arraybackslash}p{2.1cm} 
    | >{\centering\arraybackslash}p{2.1cm} 
    | >{\centering\arraybackslash}p{2.1cm} 
    | >{\centering\arraybackslash}p{2.1cm}
    | >{\centering\arraybackslash}p{2.1cm} 
    | >{\centering\arraybackslash}p{2.1cm} 
    |
}
\toprule

\multicolumn{1}{|c|}{\textbf{GT}} &
\multicolumn{2}{|c|}{\textbf{Pixel-Wise Warping}} &
\multicolumn{3}{|c|}{\textbf{Token Warping}} &
\multicolumn{1}{|c|}{\textbf{NVS}} &
\multicolumn{1}{|c|}{\textbf{GT}} \\
\midrule

Source$^\dagger$ ($I_S$) &
Forward & Backward &
Forward & Backward-Nearest & Backward-Adaptive &
GenWarp~\cite{seo2024genwarp} &
Target ($I_T$) \\
\midrule

\multicolumn{8}{|c|}{\texttt{[\benchmark{}-Text]} Question: ``\textit{Is the A point on the right or left of the B point?}'' Answer: ``\textit{left}''}\\[4pt]
\begin{minipage}{\linewidth}    
    \includegraphics[width=\linewidth]{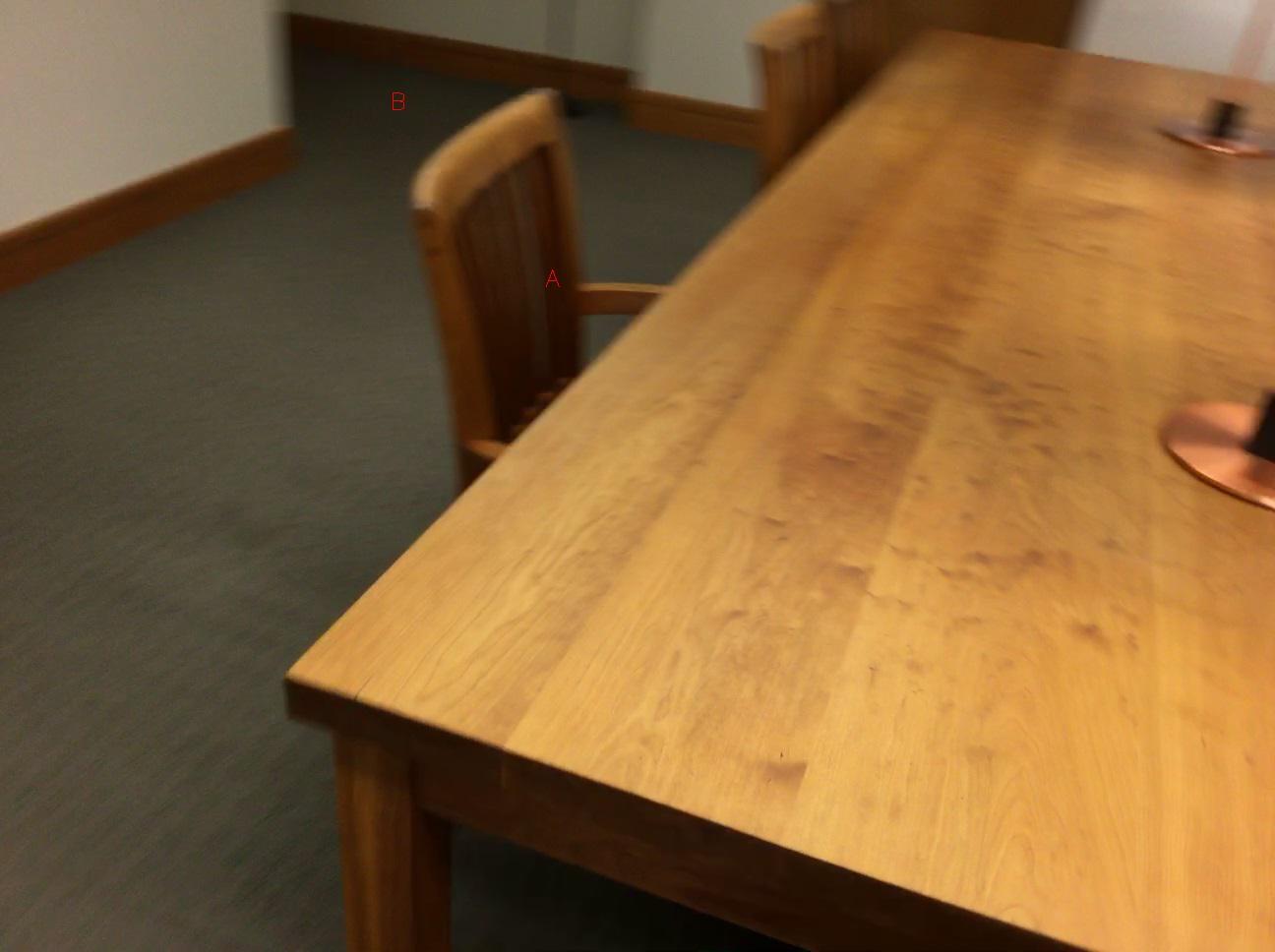}
    \centering Response: ``\textit{left}''
\end{minipage} &
\begin{minipage}{\linewidth}    
    \includegraphics[width=\linewidth]{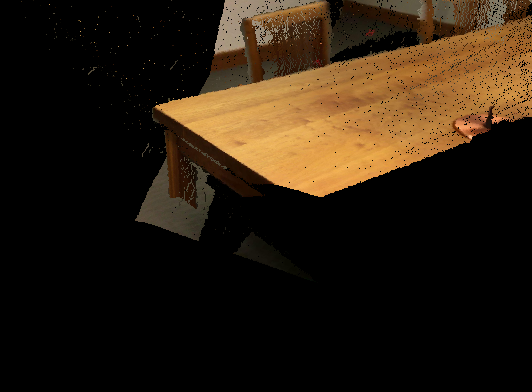}
    \centering Response: ``\textit{right}''
\end{minipage} &
\begin{minipage}{\linewidth}    
    \includegraphics[width=\linewidth]{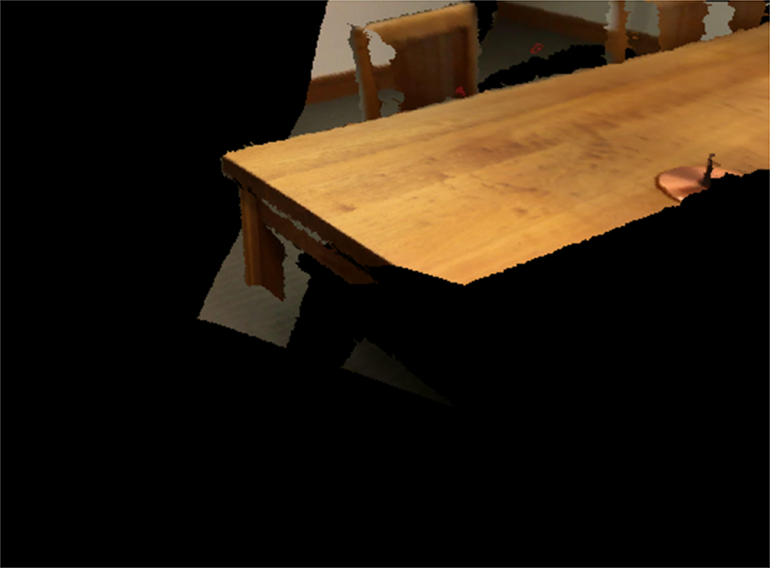}
    \centering Response: ``\textit{right}''
\end{minipage} &
\begin{minipage}{\linewidth}    
    \includegraphics[width=\linewidth]{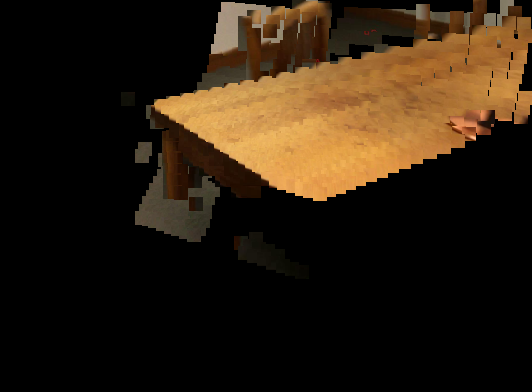}
    \centering Response: ``\textit{right}''
\end{minipage} &
\begin{minipage}{\linewidth}    
    \includegraphics[width=\linewidth]{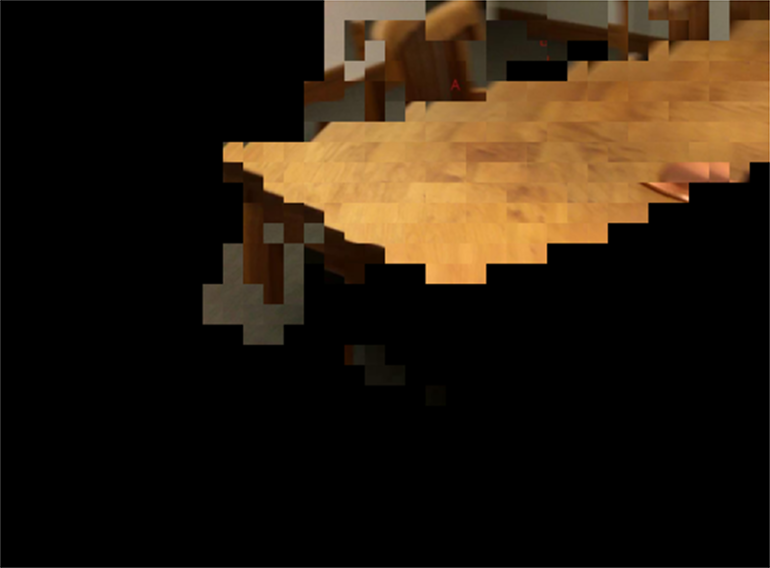}
    \centering Response: ``\textit{left}''
\end{minipage} &
\begin{minipage}{\linewidth}    
    \includegraphics[width=\linewidth]{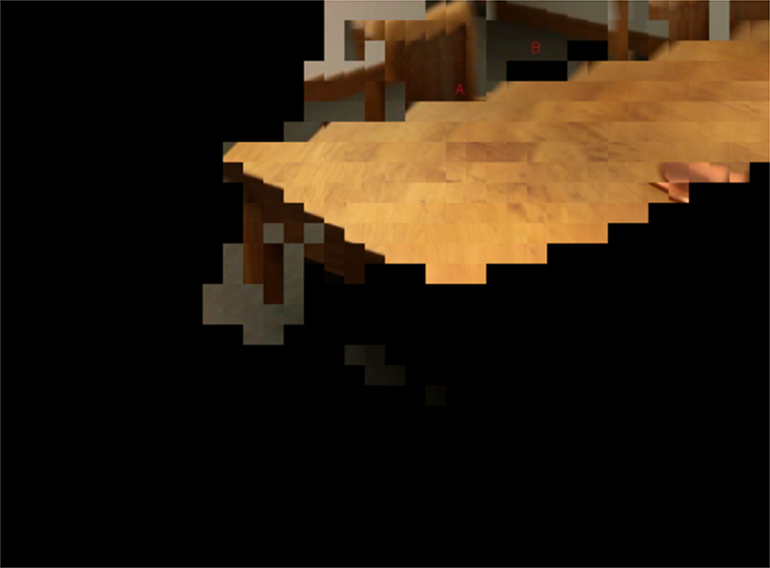}
    \centering Response: ``\textit{left}''
\end{minipage} &
\begin{minipage}{\linewidth}    
    \includegraphics[width=\linewidth]{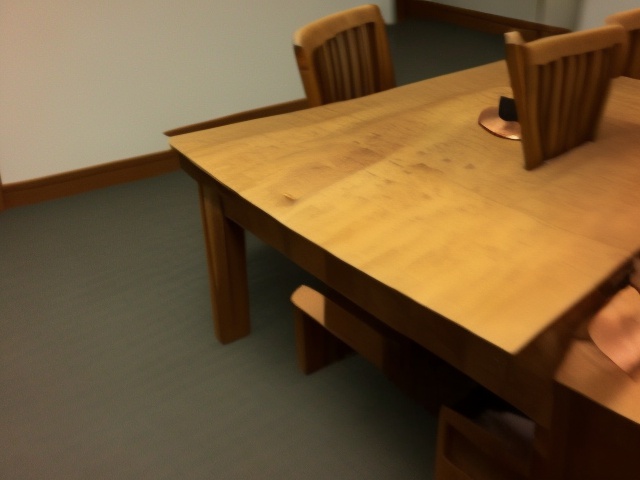}
    \centering Response: ``\textit{right}''
\end{minipage} &
\begin{minipage}{\linewidth}    
    \includegraphics[width=\linewidth]{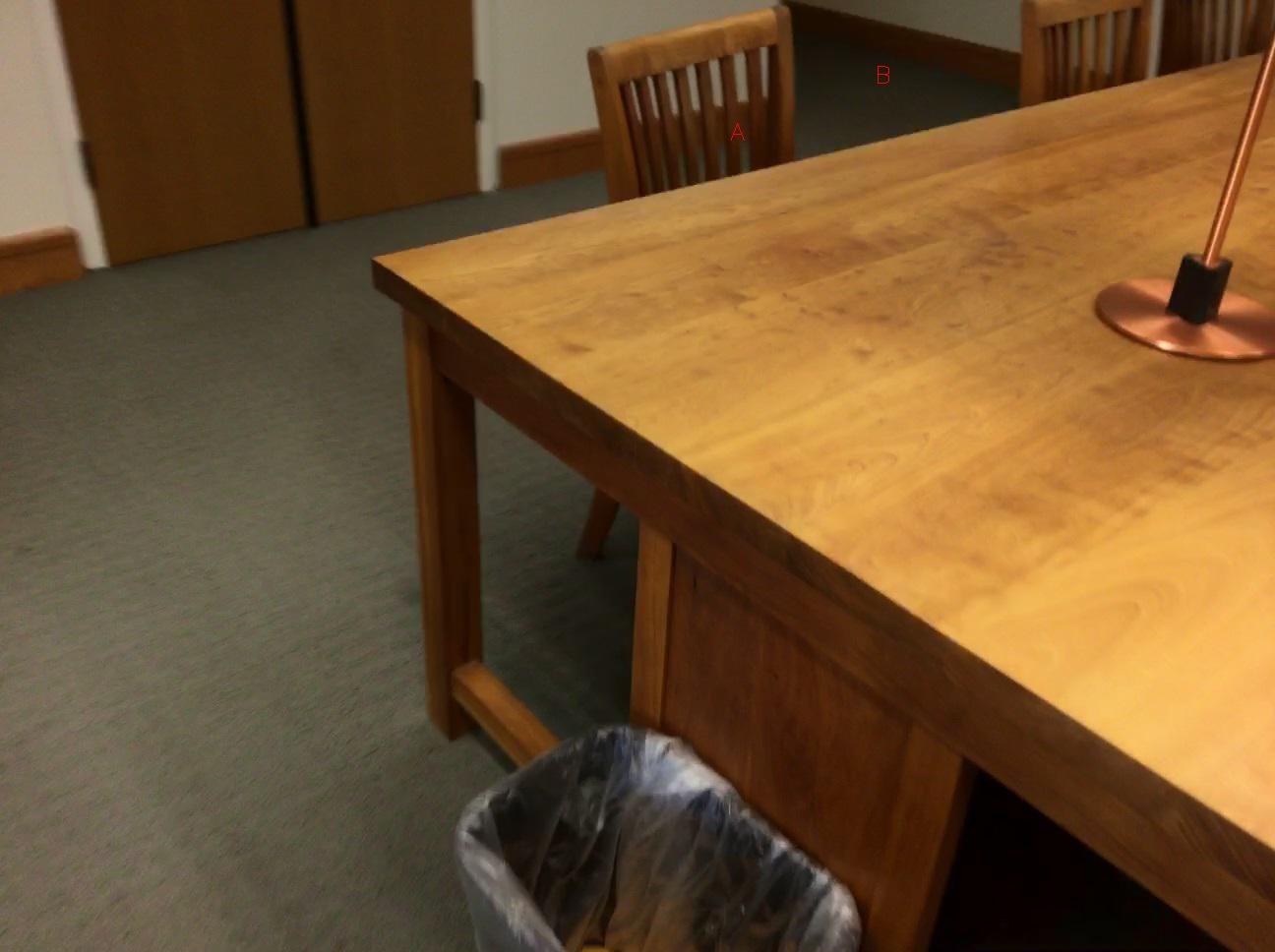}
    \centering Response: ``\textit{left}''
\end{minipage} \\

\midrule
\multicolumn{8}{|c|}{\texttt{[\benchmark{}-Text]} Question: ``\textit{Is the A point on the right or left of the B point?}'' Answer: ``\textit{right}''}\\[4pt]
\begin{minipage}{\linewidth}    
    \includegraphics[width=\linewidth]{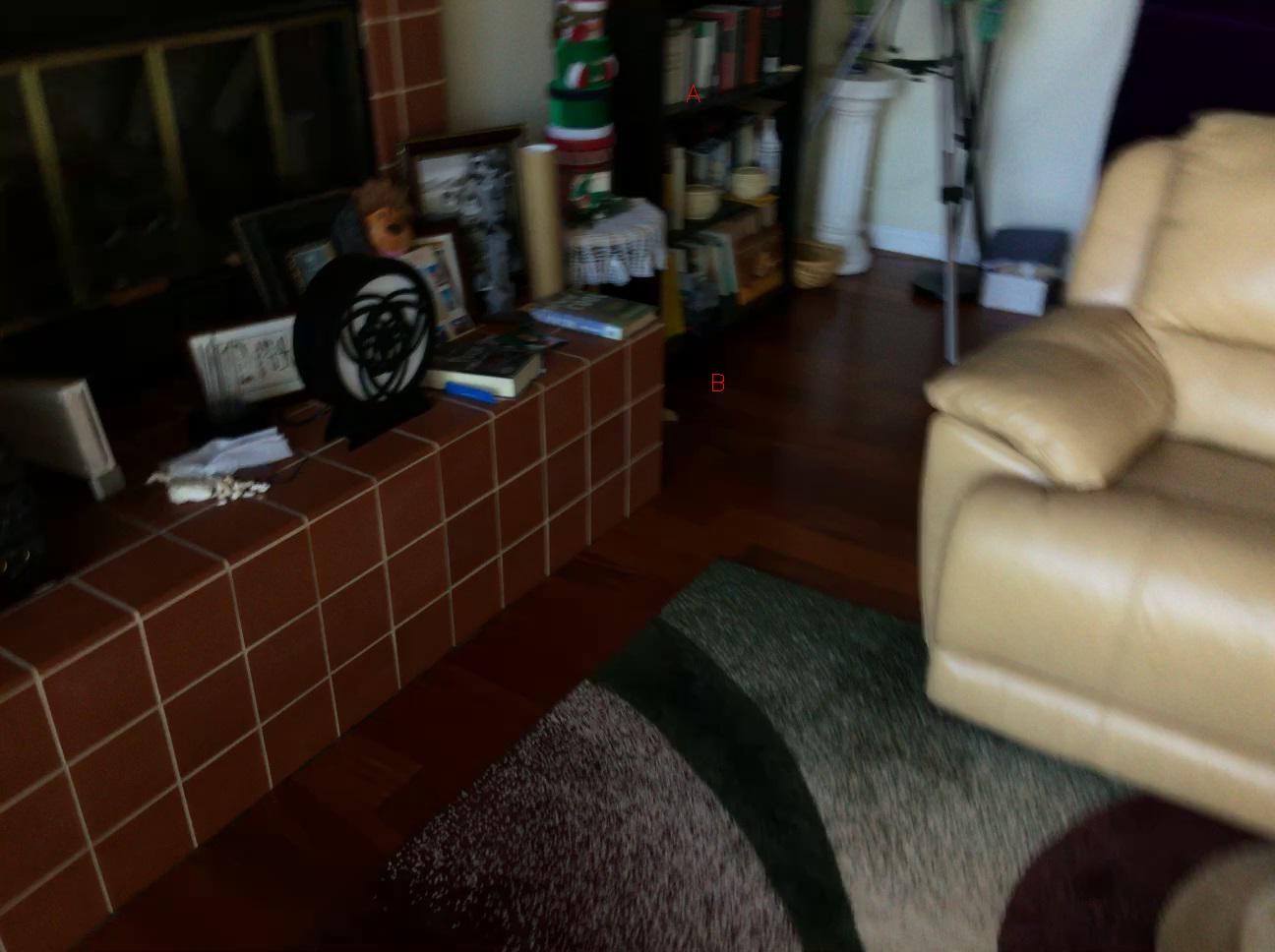}
    \centering Response: ``\textit{left}''
\end{minipage} &
\begin{minipage}{\linewidth}    
    \includegraphics[width=\linewidth]{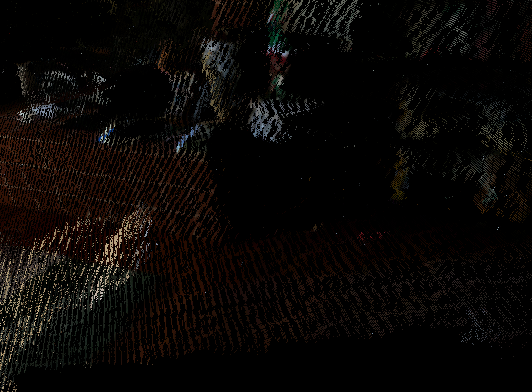}
    \centering Response: ``\textit{left}''
\end{minipage} &
\begin{minipage}{\linewidth}    
    \includegraphics[width=\linewidth]{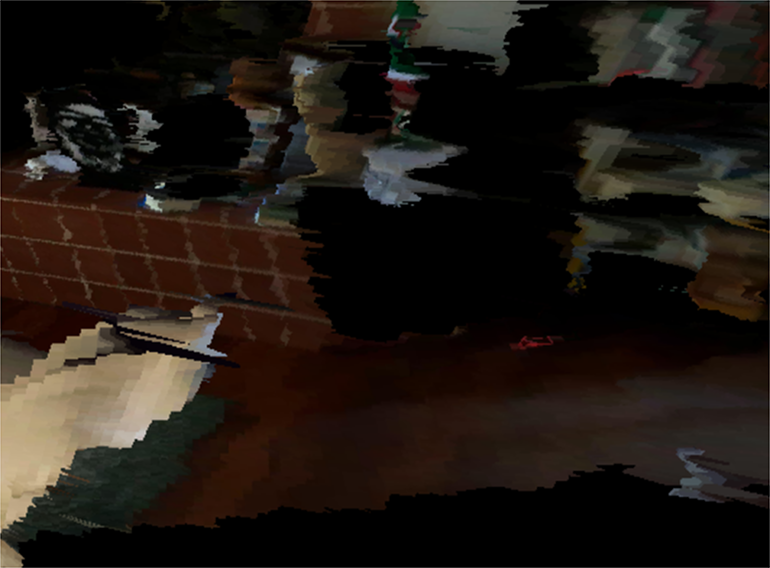}
    \centering Response: ``\textit{left}''
\end{minipage} &
\begin{minipage}{\linewidth}    
    \includegraphics[width=\linewidth]{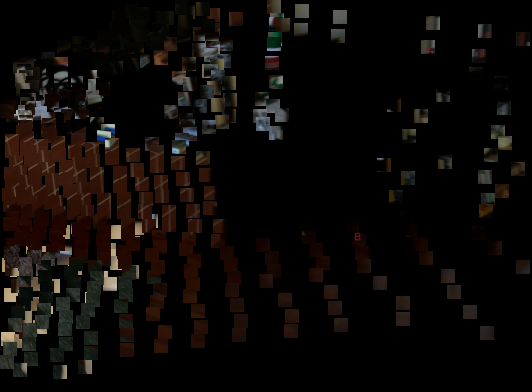}
    \centering Response: ``\textit{left}''
\end{minipage} &
\begin{minipage}{\linewidth}    
    \includegraphics[width=\linewidth]{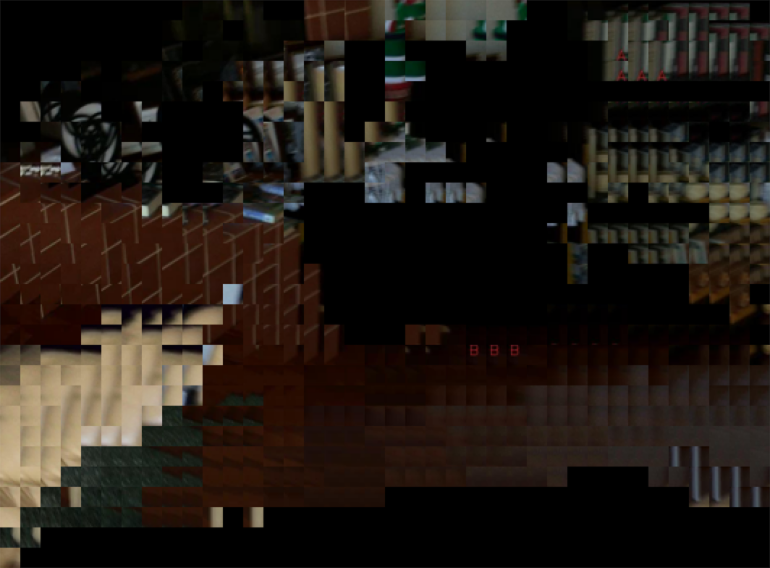}
    \centering Response: ``\textit{right}''
\end{minipage} &
\begin{minipage}{\linewidth}    
    \includegraphics[width=\linewidth]{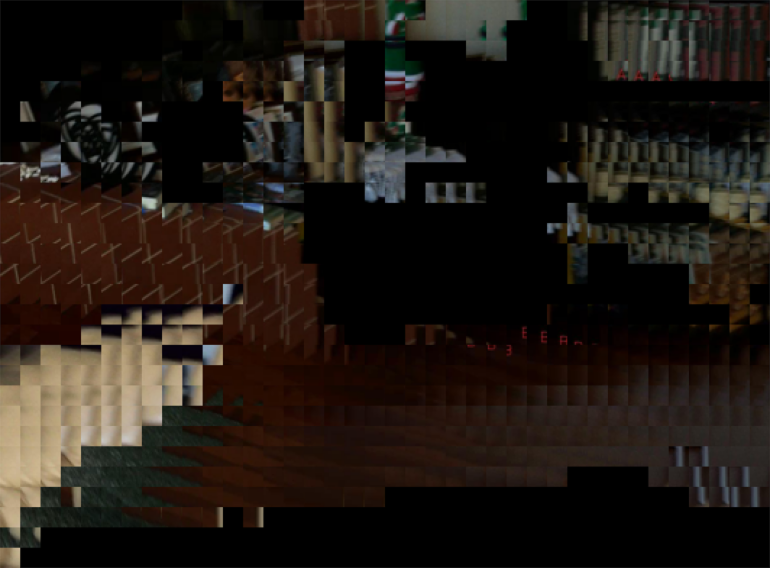}
    \centering Response: ``\textit{right}''
\end{minipage} &
\begin{minipage}{\linewidth}    
    \includegraphics[width=\linewidth]{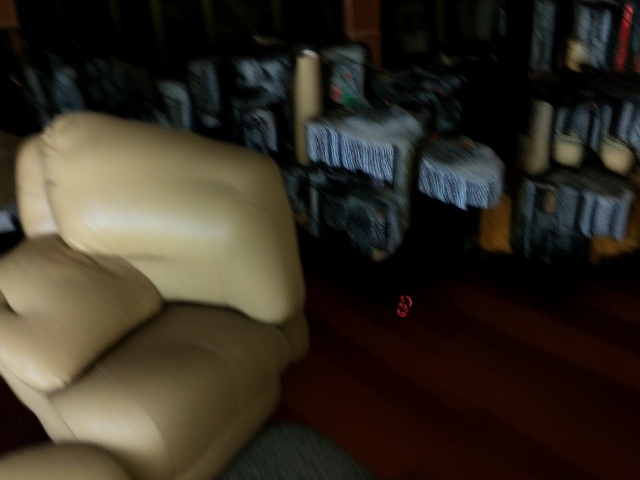}
    \centering Response: ``\textit{left}''
\end{minipage} &
\begin{minipage}{\linewidth}    
    \includegraphics[width=\linewidth]{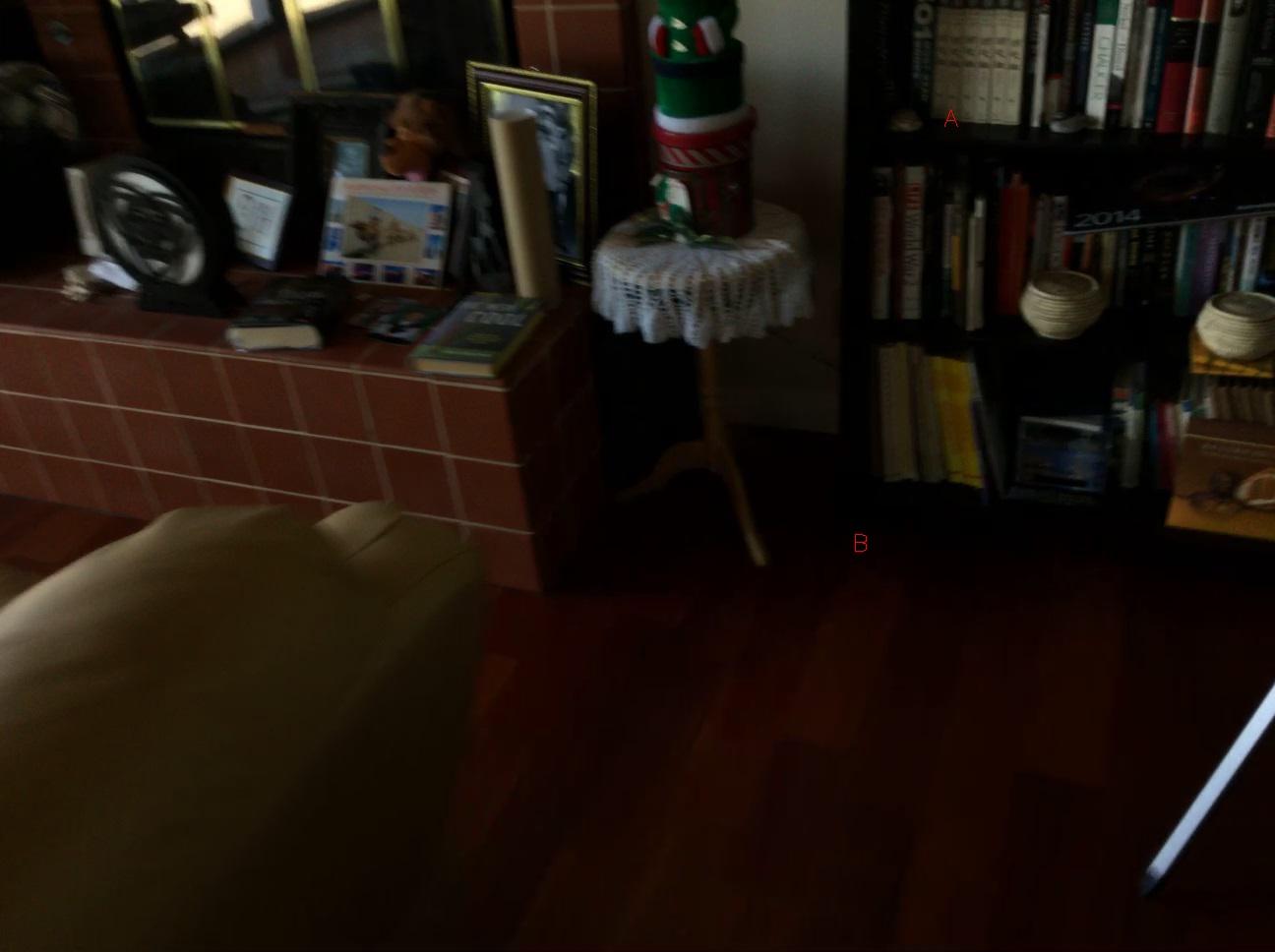}
    \centering Response: ``\textit{right}''
\end{minipage} \\

\midrule
\multicolumn{8}{|c|}{\texttt{[\benchmark{}-Text]} Question: ``\textit{Is the A point on the right or left of the B point?}'' Answer: ``\textit{right}''}\\[4pt]
\begin{minipage}{\linewidth}    
    \includegraphics[width=\linewidth]{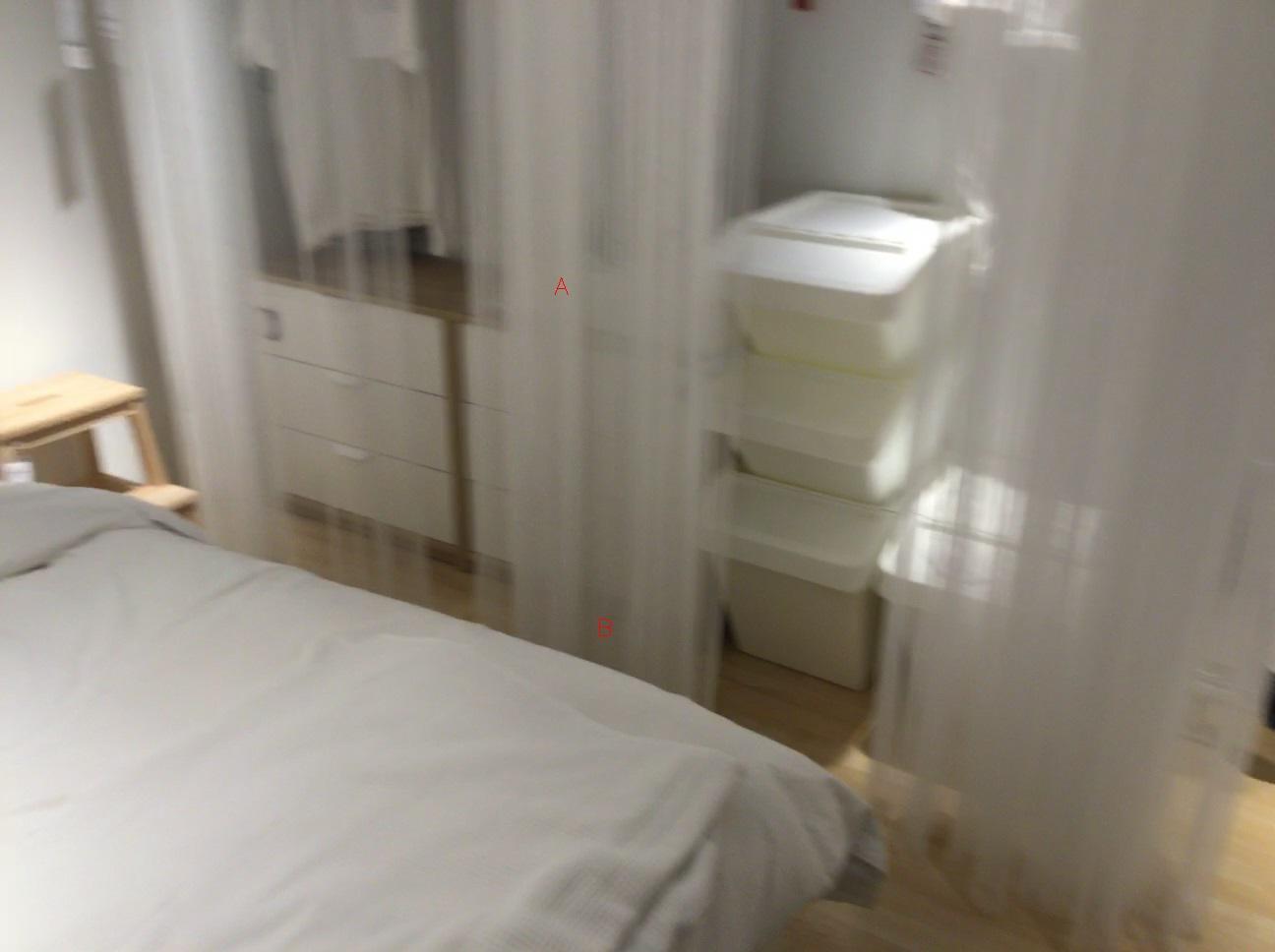}
    \centering Response: ``\textit{right}''
\end{minipage} &
\begin{minipage}{\linewidth}    
    \includegraphics[width=\linewidth]{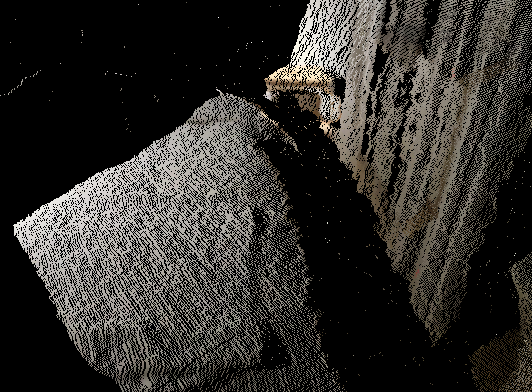}
    \centering Response: ``\textit{left}''
\end{minipage} &
\begin{minipage}{\linewidth}    
    \includegraphics[width=\linewidth]{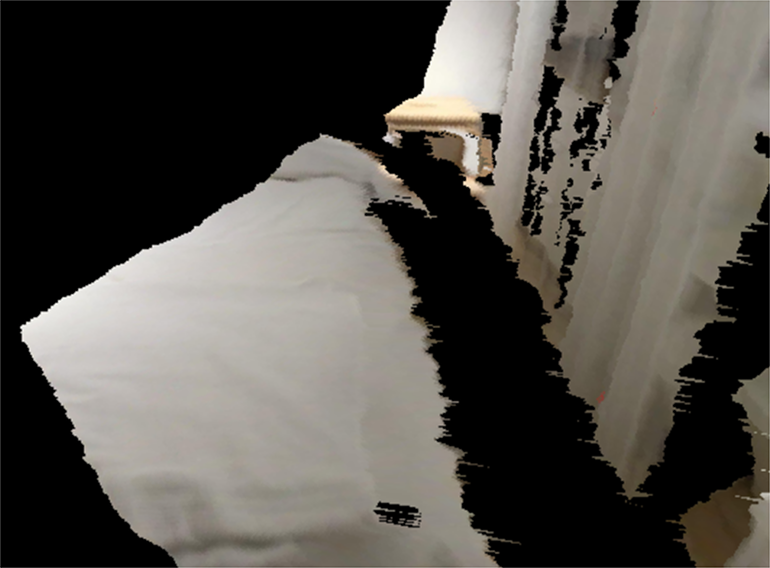}
    \centering Response: ``\textit{left}''
\end{minipage} &
\begin{minipage}{\linewidth}    
    \includegraphics[width=\linewidth]{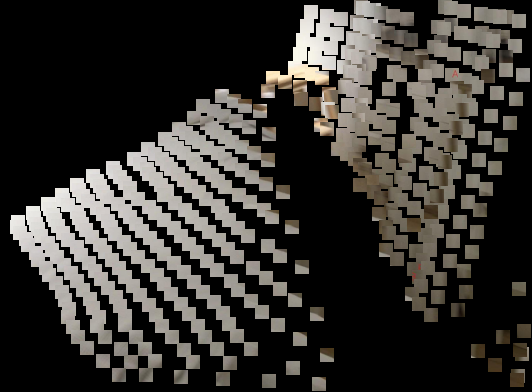}
    \centering Response: ``\textit{left}''
\end{minipage} &
\begin{minipage}{\linewidth}    
    \includegraphics[width=\linewidth]{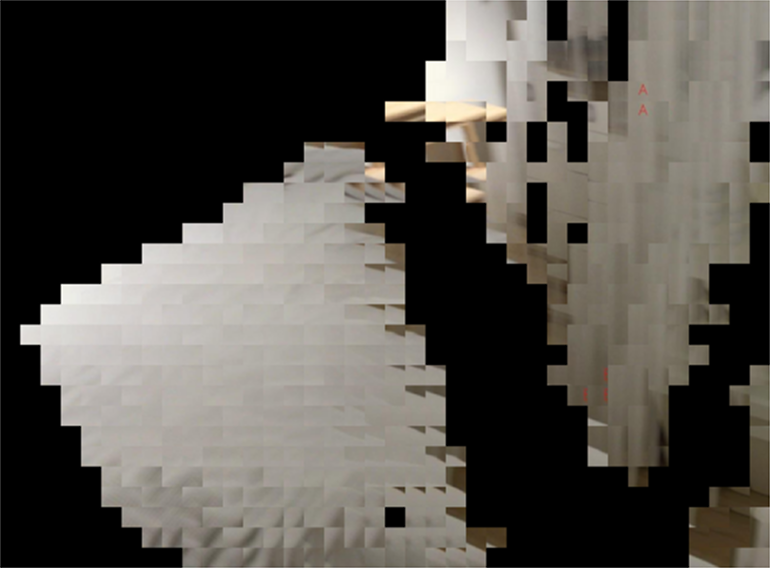}
    \centering Response: ``\textit{right}''
\end{minipage} &
\begin{minipage}{\linewidth}    
    \includegraphics[width=\linewidth]{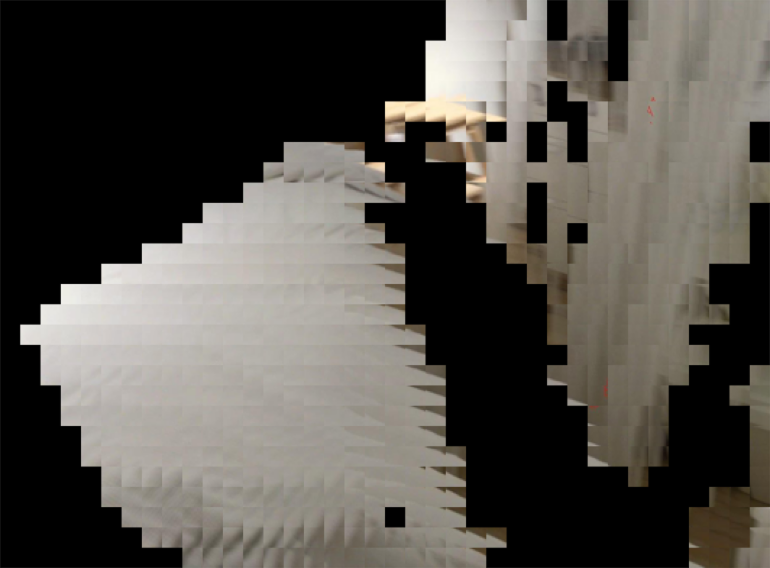}
    \centering Response: ``\textit{right}''
\end{minipage} &
\begin{minipage}{\linewidth}    
    \includegraphics[width=\linewidth]{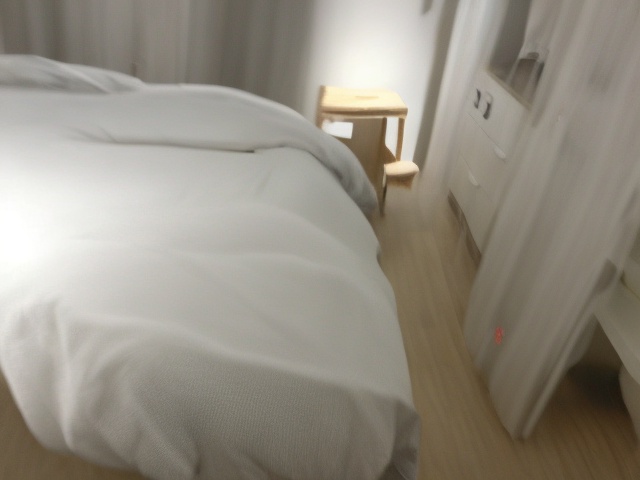}
    \centering Response: ``\textit{left}''
\end{minipage} &
\begin{minipage}{\linewidth}    
    \includegraphics[width=\linewidth]{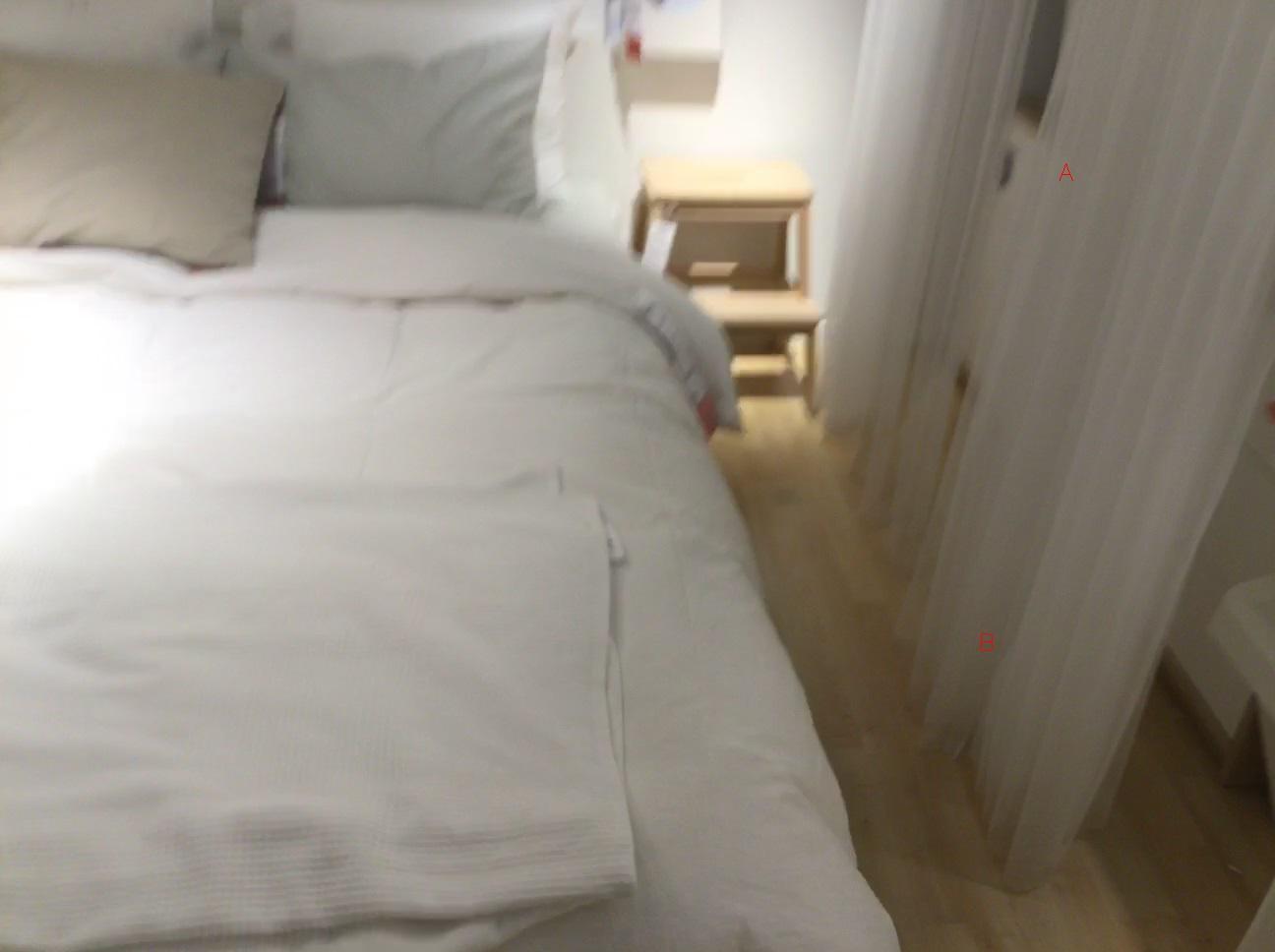}
    \centering Response: ``\textit{right}''
\end{minipage} \\

\midrule
\multicolumn{8}{|c|}{\texttt{[\benchmark{}-Shape]} Question: ``\textit{Is the star shape on the right or left of the triangle shape?}'' Answer: ``\textit{left}''}\\[4pt]
\begin{minipage}{\linewidth}    
    \includegraphics[width=\linewidth]{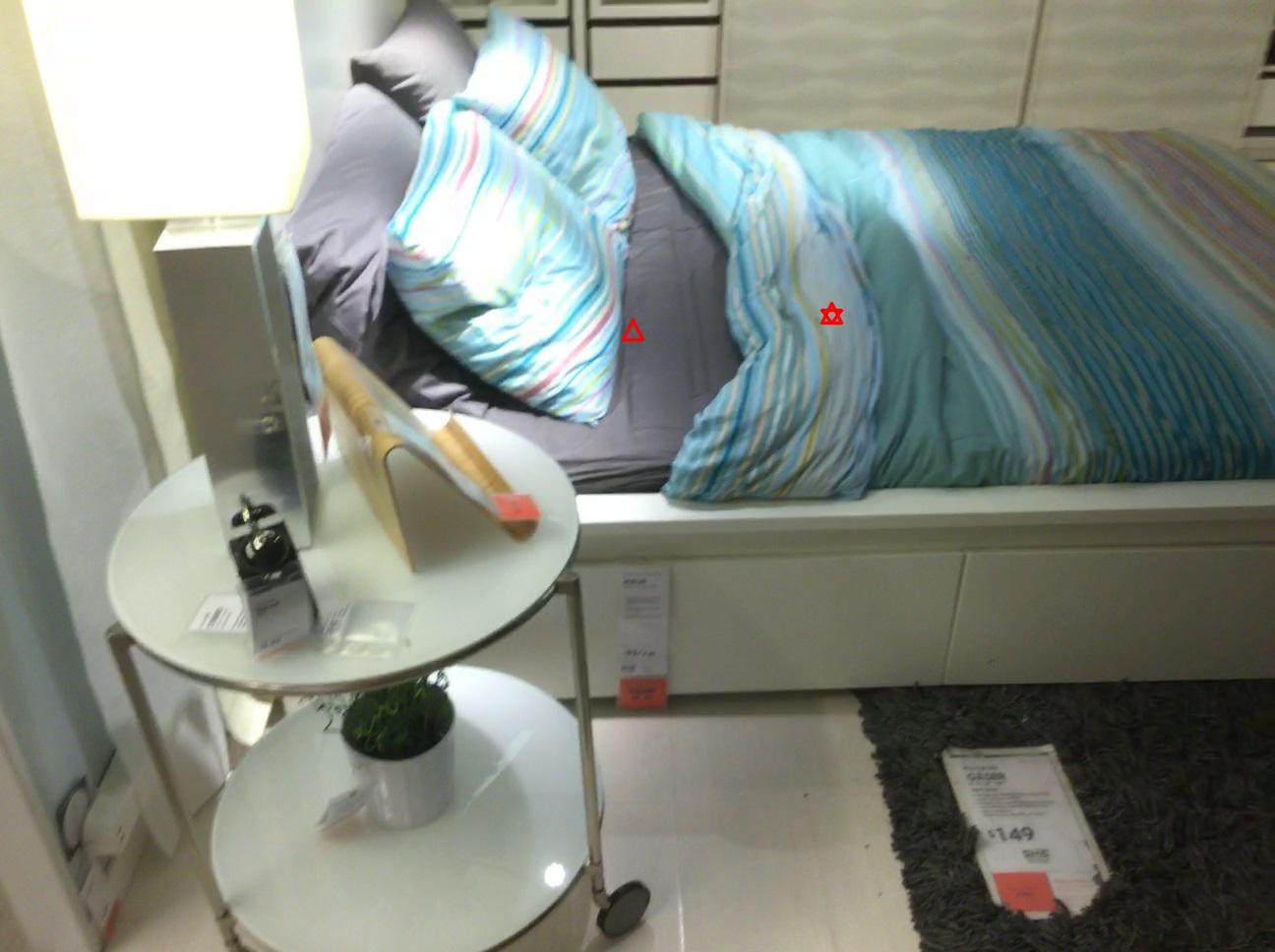}
    \centering Response: ``\textit{right}''
\end{minipage} &
\begin{minipage}{\linewidth}    
    \includegraphics[width=\linewidth]{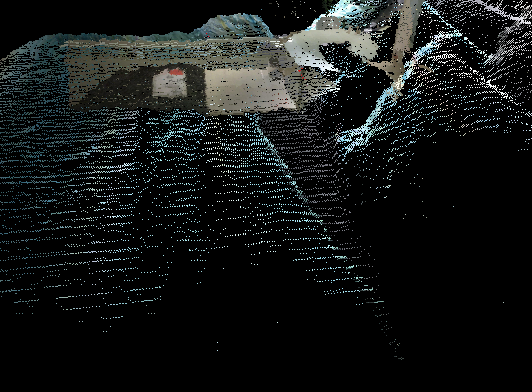}
    \centering Response: ``\textit{right}''
\end{minipage} &
\begin{minipage}{\linewidth}    
    \includegraphics[width=\linewidth]{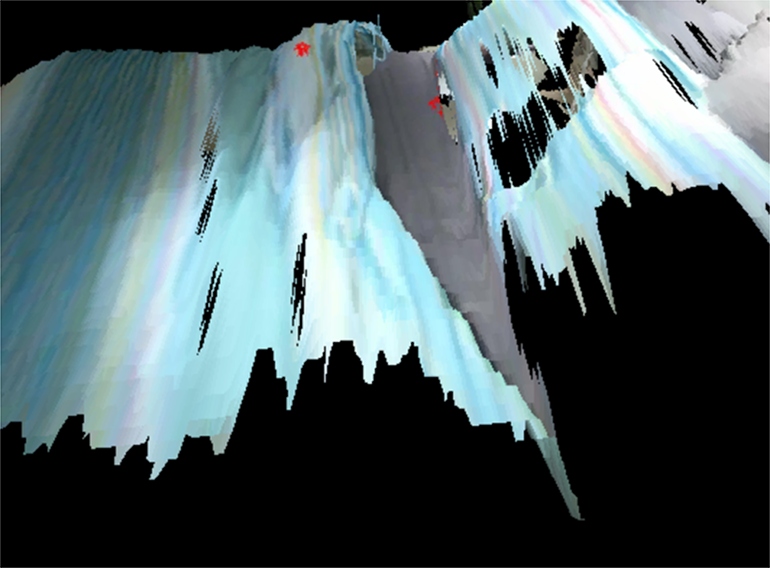}
    \centering Response: ``\textit{right}''
\end{minipage} &
\begin{minipage}{\linewidth}    
    \includegraphics[width=\linewidth]{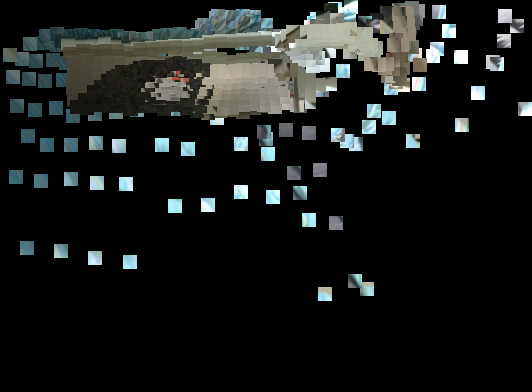}
    \centering Response: ``\textit{right}''
\end{minipage} &
\begin{minipage}{\linewidth}    
    \includegraphics[width=\linewidth]{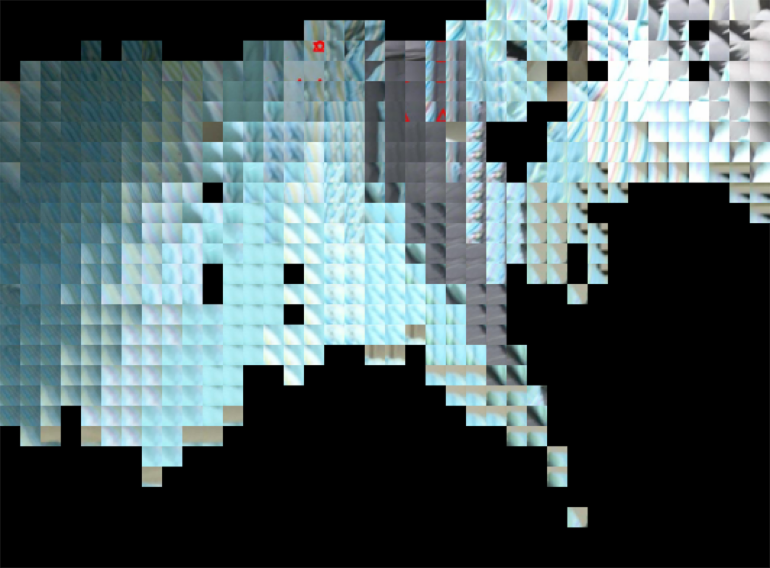}
    \centering Response: ``\textit{right}''
\end{minipage} &
\begin{minipage}{\linewidth}    
    \includegraphics[width=\linewidth]{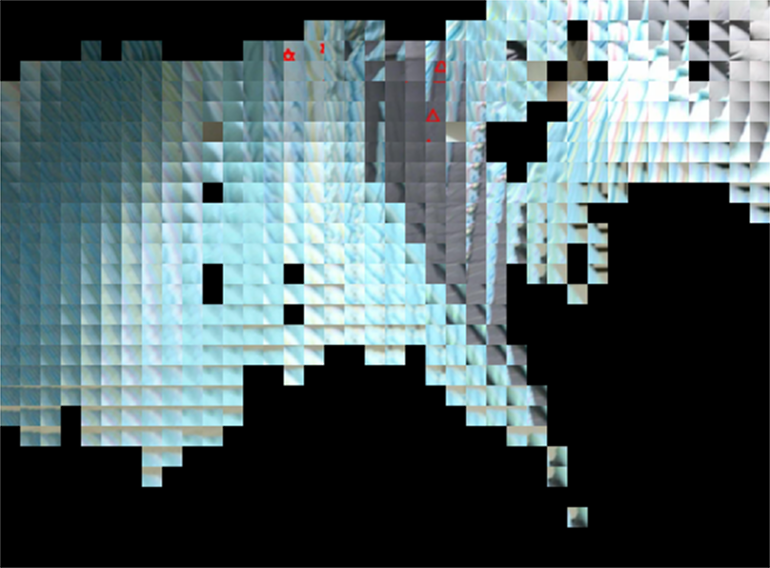}
    \centering Response: ``\textit{left}''
\end{minipage} &
\begin{minipage}{\linewidth}    
    \includegraphics[width=\linewidth]{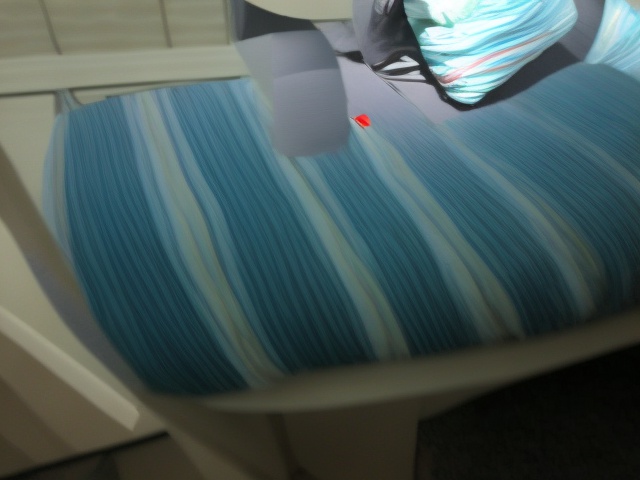}
    \centering Response: ``\textit{right}''
\end{minipage} &
\begin{minipage}{\linewidth}    
    \includegraphics[width=\linewidth]{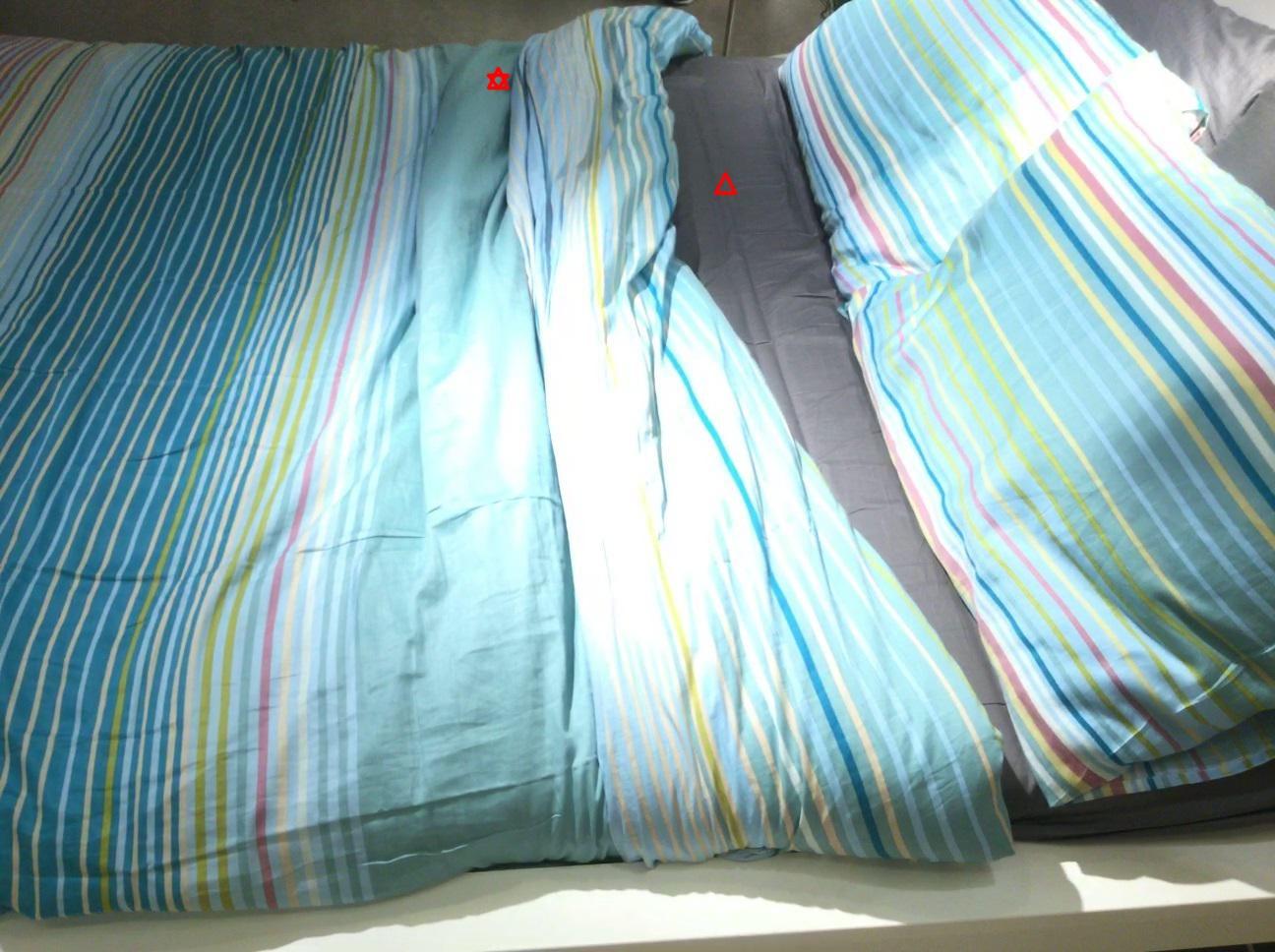}
    \centering Response: ``\textit{left}''
\end{minipage} \\

\midrule

\multicolumn{8}{|c|}{\texttt{[\benchmark{}-Shape]} Question: ``\textit{Is the star shape on the right or left of the triangle shape?}'' Answer: ``\textit{left}''}\\[4pt]
\begin{minipage}{\linewidth}    
    \includegraphics[width=\linewidth]{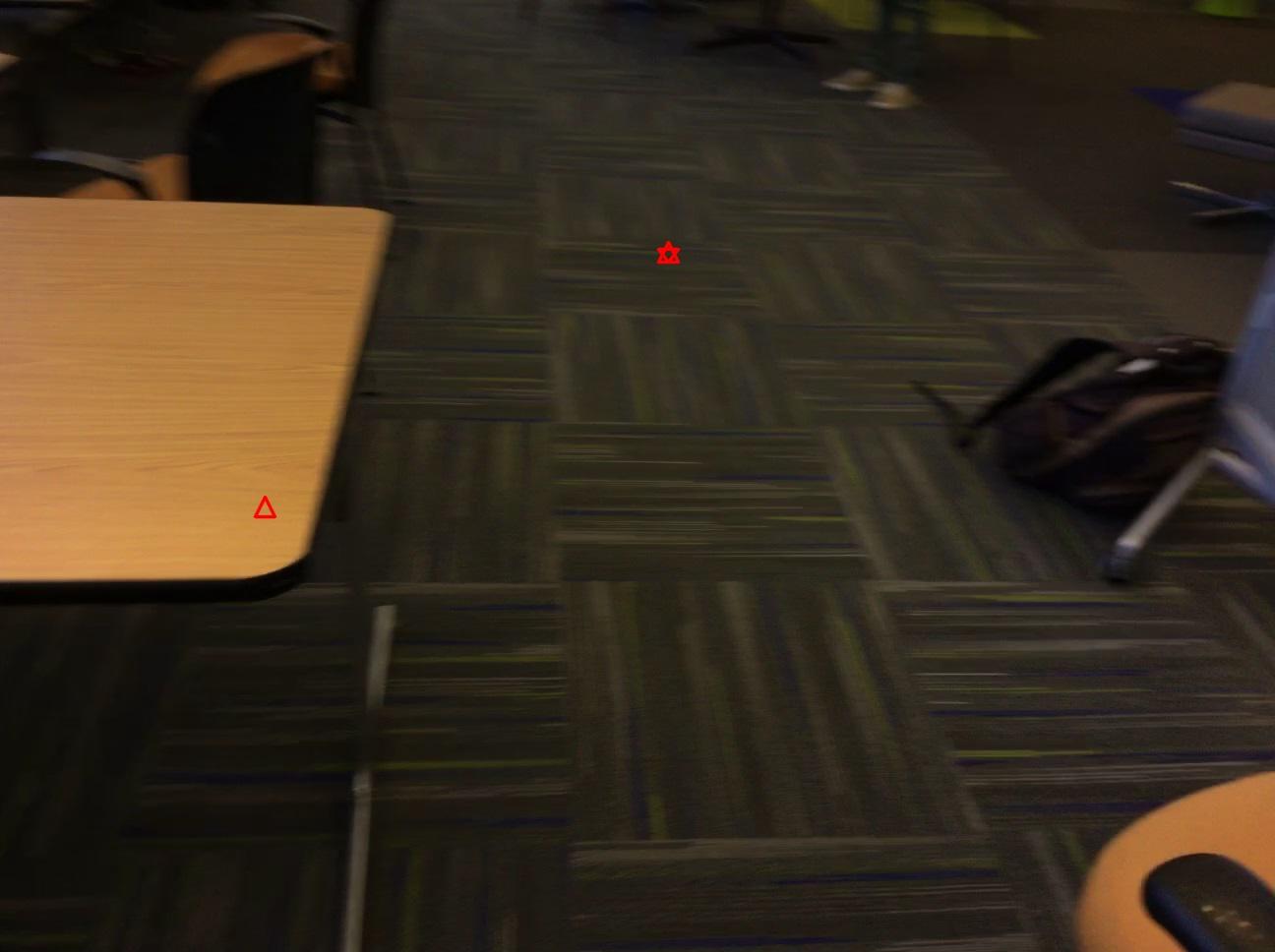}
    \centering Response: ``\textit{right}''
\end{minipage} &
\begin{minipage}{\linewidth}    
    \includegraphics[width=\linewidth]{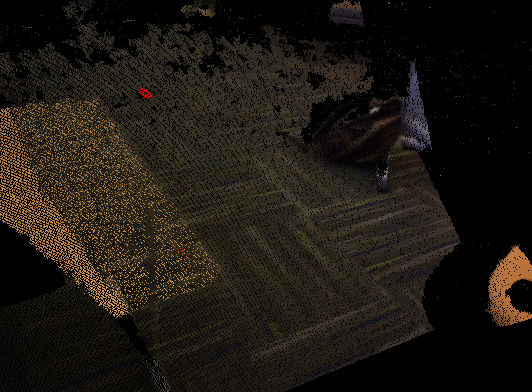}
    \centering Response: ``\textit{right}''
\end{minipage} &
\begin{minipage}{\linewidth}    
    \includegraphics[width=\linewidth]{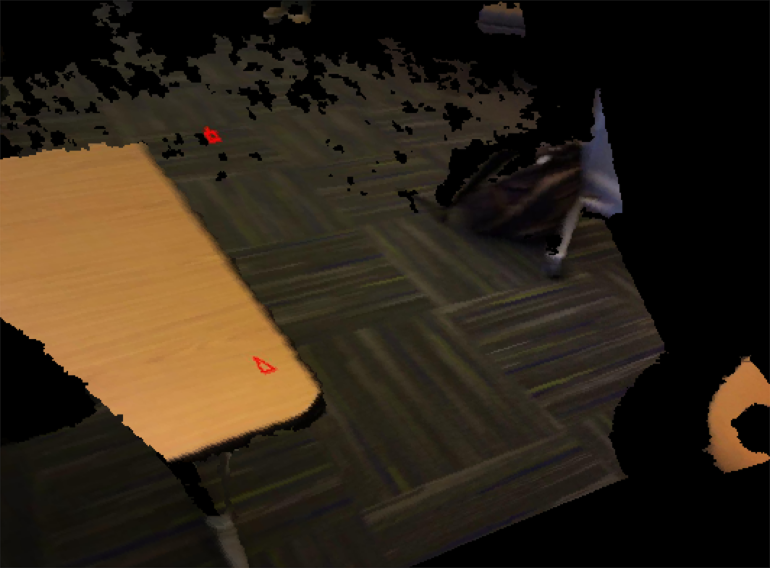}
    \centering Response: ``\textit{right}''
\end{minipage} &
\begin{minipage}{\linewidth}    
    \includegraphics[width=\linewidth]{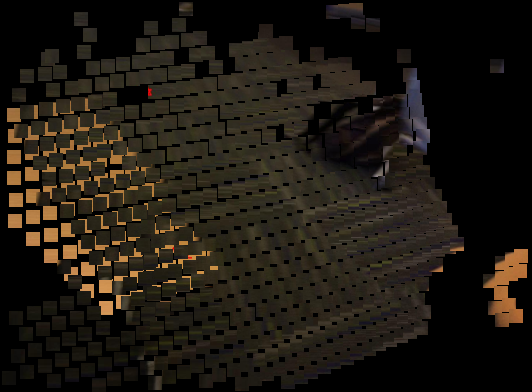}
    \centering Response: ``\textit{right}''
\end{minipage} &
\begin{minipage}{\linewidth}    
    \includegraphics[width=\linewidth]{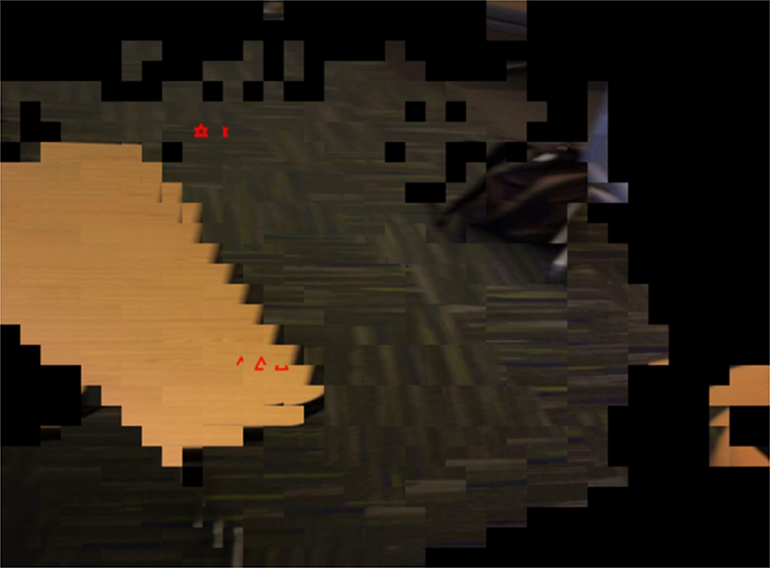}
    \centering Response: ``\textit{left}''
\end{minipage} &
\begin{minipage}{\linewidth}    
    \includegraphics[width=\linewidth]{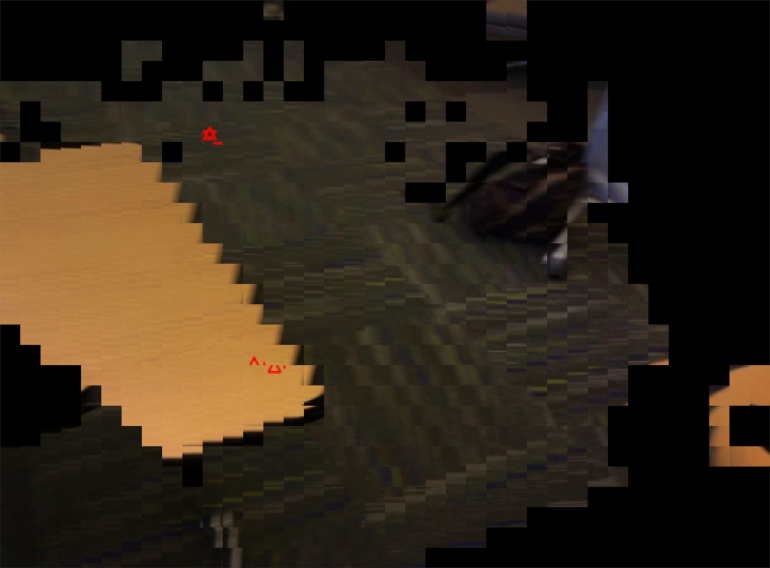}
    \centering Response: ``\textit{left}''
\end{minipage} &
\begin{minipage}{\linewidth}    
    \includegraphics[width=\linewidth]{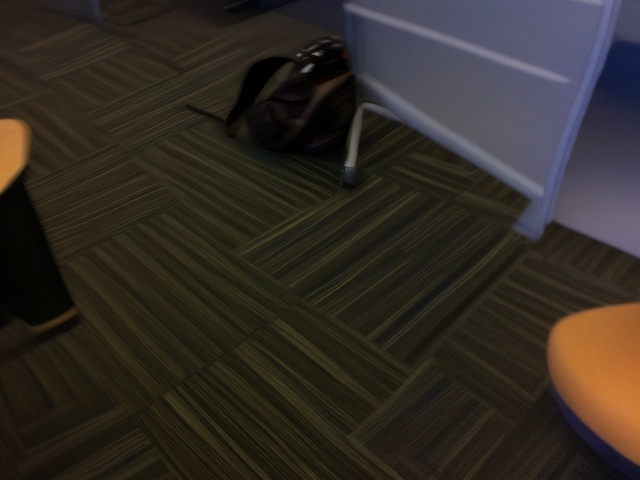}
    \centering Response: ``\textit{None}''
\end{minipage} &
\begin{minipage}{\linewidth}    
    \includegraphics[width=\linewidth]{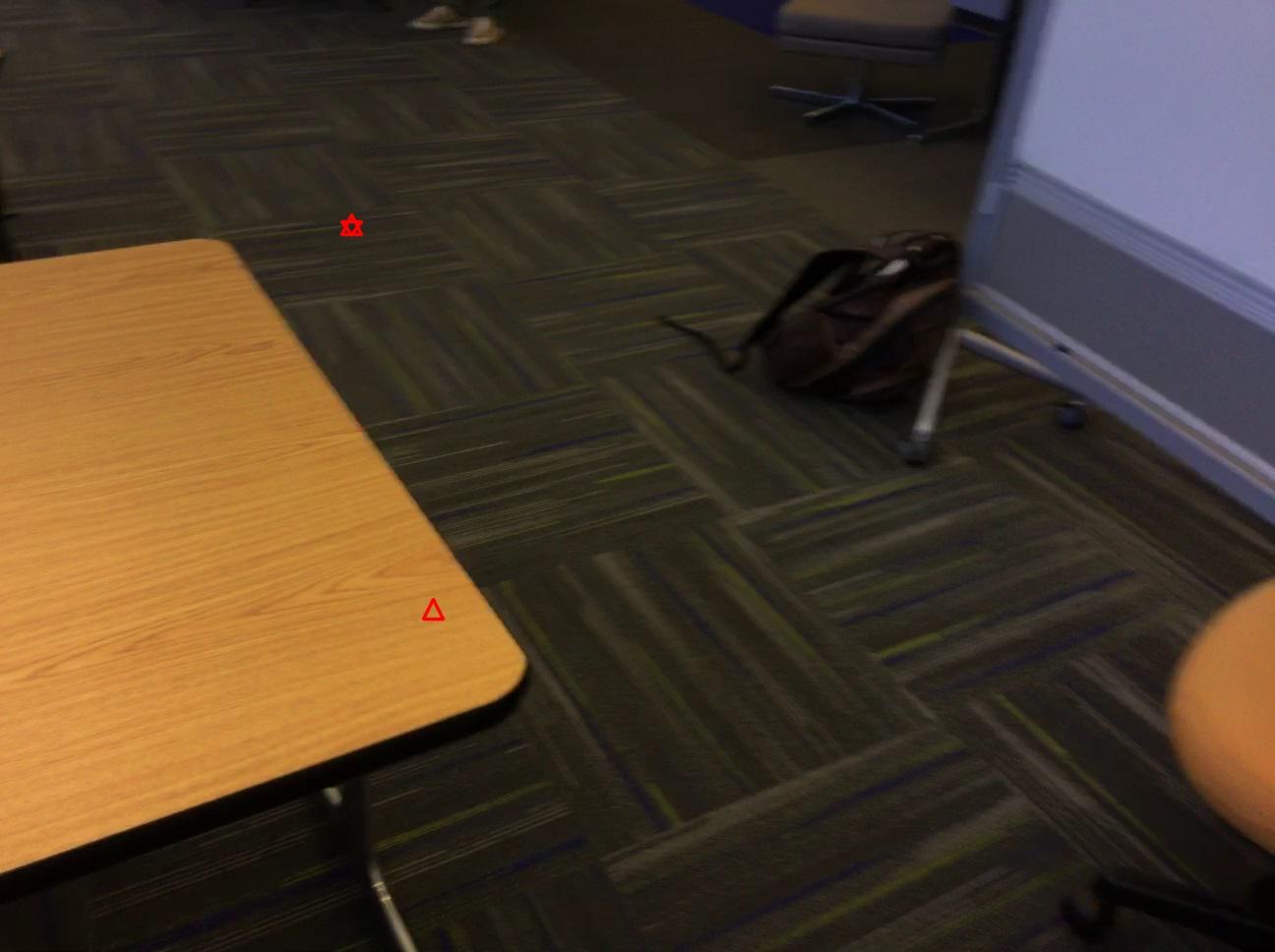}
    \centering Response: ``\textit{left}''
\end{minipage} \\

\midrule

\multicolumn{8}{|c|}{\texttt{[\benchmark{}-Shape]} Question: ``\textit{Is the star shape on the left or right of the triangle shape?}'' Answer: ``\textit{right}''}\\[4pt]
\begin{minipage}{\linewidth}    
    \includegraphics[width=\linewidth]{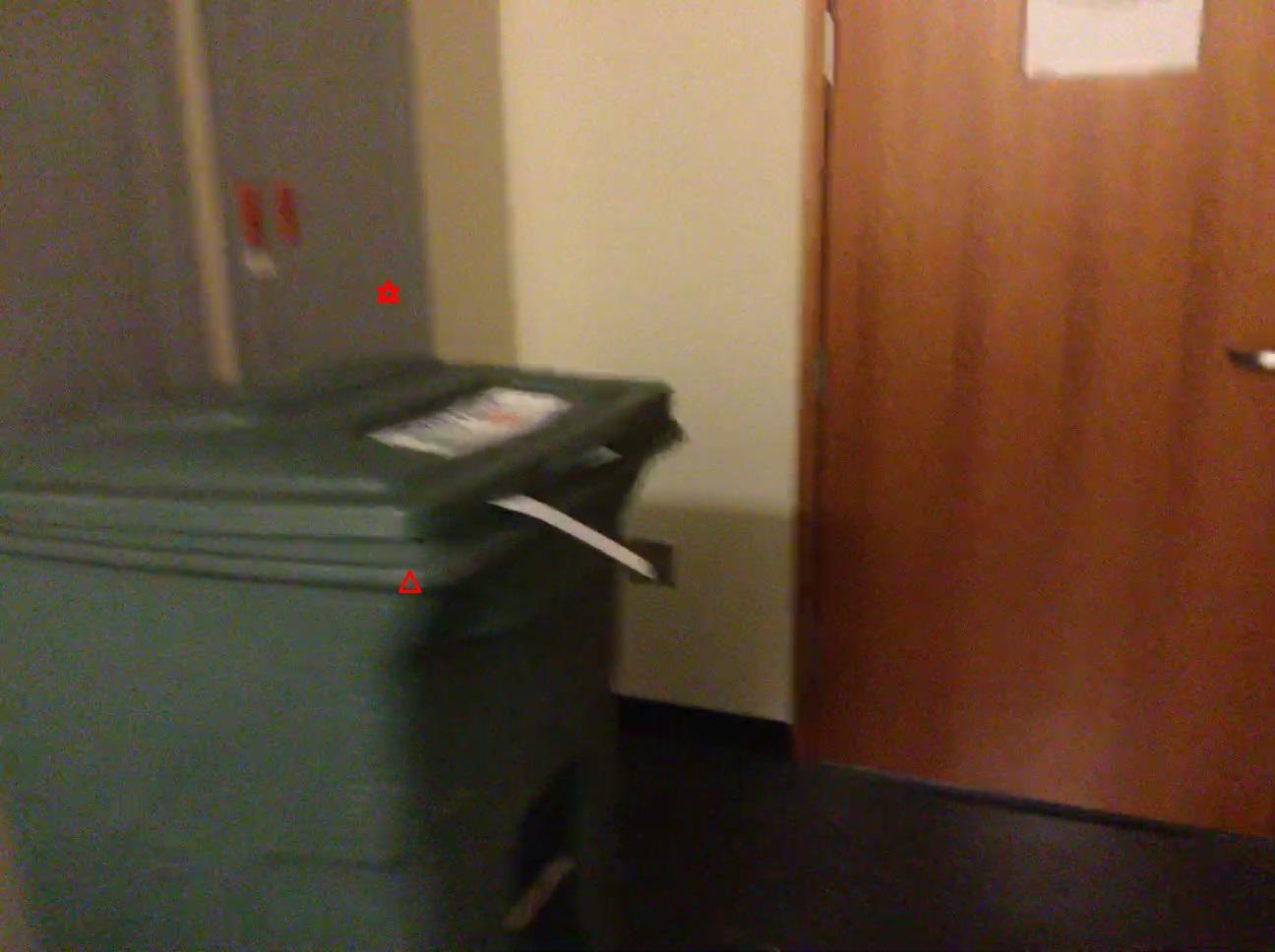}
    \centering Response: ``\textit{left}''
\end{minipage} &
\begin{minipage}{\linewidth}    
    \includegraphics[width=\linewidth]{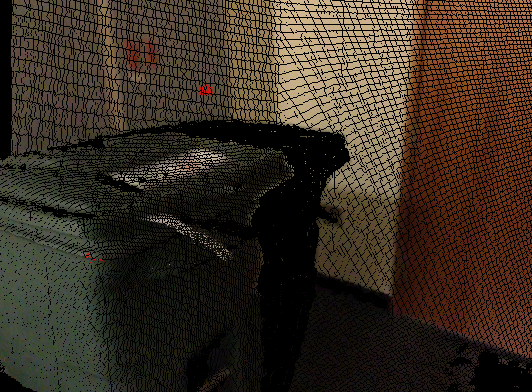}
    \centering Response: ``\textit{left}''
\end{minipage} &
\begin{minipage}{\linewidth}    
    \includegraphics[width=\linewidth]{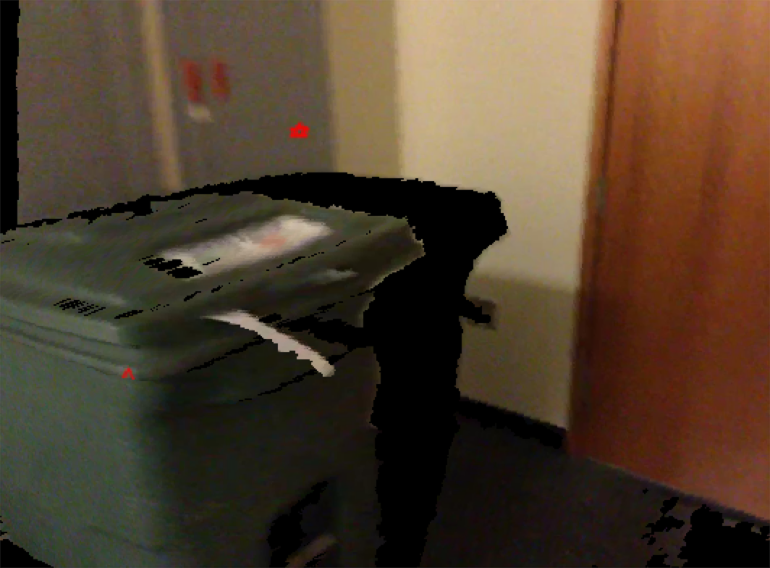}
    \centering Response: ``\textit{left}''
\end{minipage} &
\begin{minipage}{\linewidth}    
    \includegraphics[width=\linewidth]{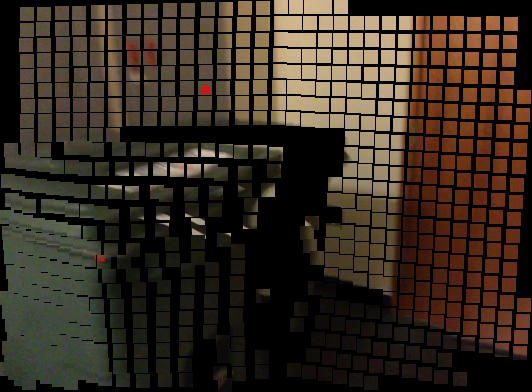}
    \centering Response: ``\textit{left}''
\end{minipage} &
\begin{minipage}{\linewidth}    
    \includegraphics[width=\linewidth]{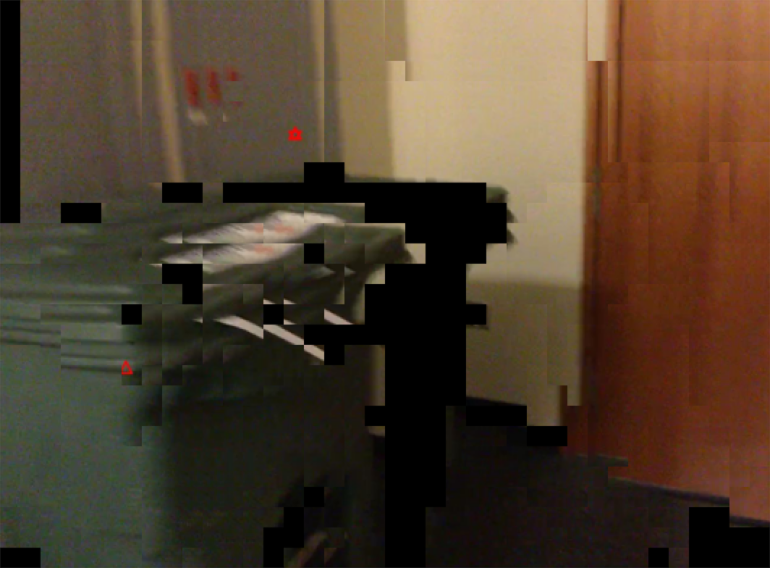}
    \centering Response: ``\textit{left}''
\end{minipage} &
\begin{minipage}{\linewidth}    
    \includegraphics[width=\linewidth]{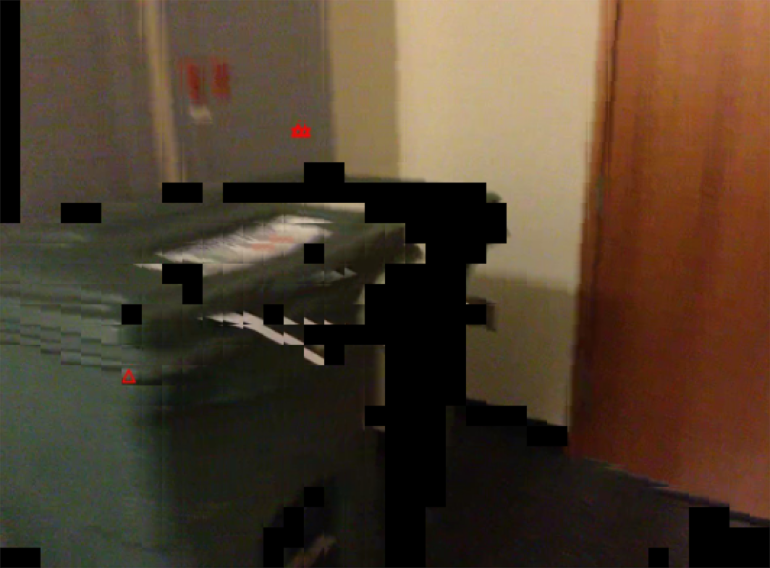}
    \centering Response: ``\textit{right}''
\end{minipage} &
\begin{minipage}{\linewidth}    
    \includegraphics[width=\linewidth]{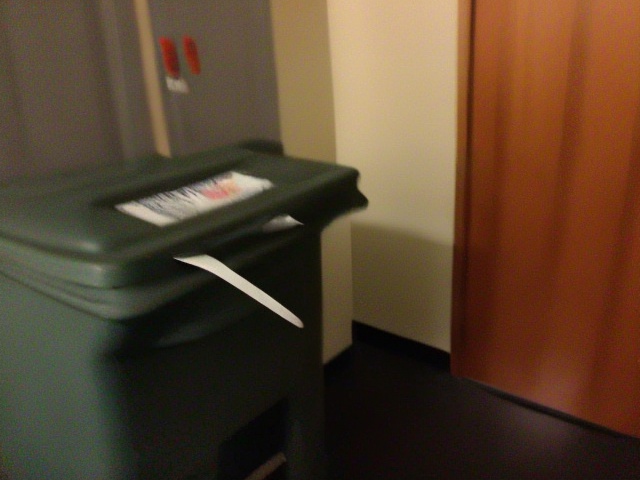}
    \centering Response: ``\textit{left}''
\end{minipage} &
\begin{minipage}{\linewidth}    
    \includegraphics[width=\linewidth]{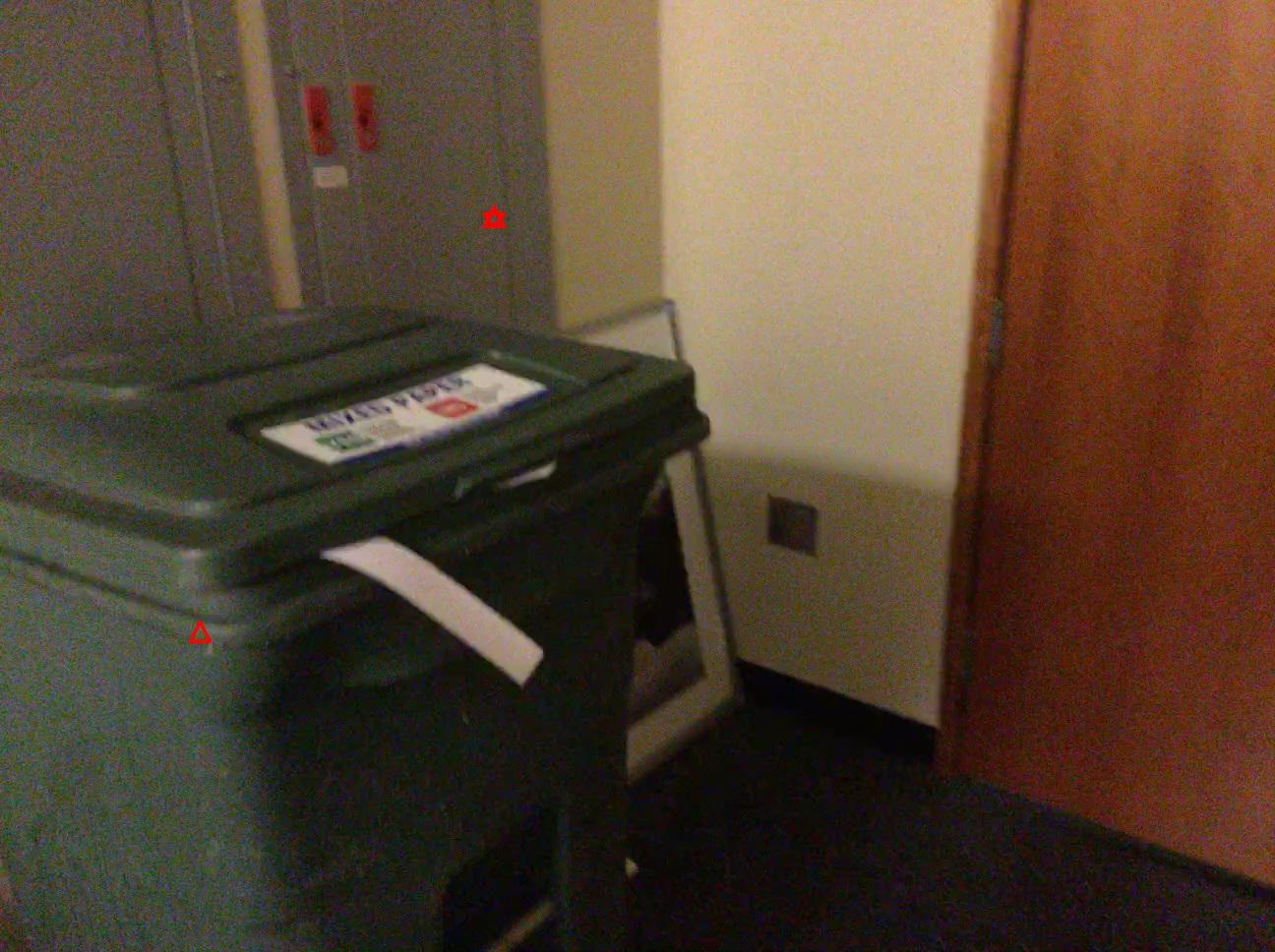}
    \centering Response: ``\textit{right}''
\end{minipage} \\

\bottomrule
\end{tabularx}
\vspace{-0.25\baselineskip}
\caption{\textbf{Warping Visualizations.} We compare the warped results of pixel-wise warping, token warping, and the generative NVS output~\cite{seo2024genwarp}. The rightmost image shows the ground-truth target viewpoint. For token warping, we visualize the RGB image patches corresponding to each token for illustration only. Above each row, we provide the question $Q$ from \texttt{ViewBench}, and below each image we show the response from Qwen2.5-VL~\cite{bai2025qwen25vl} when given the corresponding warped result. $^\dagger$The camera motion from the source view to the target view is additionally supplied as part of the prompt.}
\label{fig:quali_benchmark}
\vspace{-0.5\baselineskip}
\end{figure*}

\section{ViewBench}
\label{sec:viewbench}
In this section, we introduce~\benchmark{}, a benchmark designed to assess MLLMs' ability to perform spatial reasoning tasks that require imagining a scene from alternative viewpoints while accurately transferring fine-grained details from the observed viewpoint.

\vspace{-0.25\baselineskip}
\paragraph{Data.}
To construct source–target viewpoint pairs for generating VQAs, we collect image pairs captured from adjacent viewpoints with overlapping fields of view, drawn from real-world scans in ScanNet~\cite{dai2017scannet}.
The collected pairs are divided into difficulty levels based on their overlap ratios~\cite{xu2025_multispatialmllm}, which reflect the amount of shared content between the two views.
For each pair, one viewpoint is designated as the source, with image $\mathbf{I}_S$ and pose $\Pi_S$, and the other as the target, with image $\mathbf{I}_T$ and pose $\Pi_T$.
We then generate a question $Q$ answerable only from the target viewpoint, using information available in the source view together with an instruction describing the relative pose change between the two viewpoints. 
Importantly, we ensure that $Q$ refers only to regions visible in both views, avoiding content that is occluded or unseen from the target view.

\vspace{-0.25\baselineskip}
\paragraph{Tasks.}
The form of $Q$ depends on the specific task.
We design two tasks, both tailored to evaluate an MLLM's ability to simulate viewpoint changes for spatial reasoning: (1) view-conditioned spatial reasoning and (2) target-view object description.

\begin{itemize}
    \item \textbf{View-Conditioned Spatial Reasoning.}  
    This task evaluates whether an MLLM can reason about spatial relationships from a transformed viewpoint. To construct $Q$, we identify two points visible in both $\mathbf{I}_S$ and $\mathbf{I}_T$ whose left-right spatial relationship is reversed after the viewpoint change.
    These points are annotated using either text labels (\texttt{\benchmark{}-Text}) or simple geometric shapes (\texttt{\benchmark{}-Shape}), and $Q$ asks whether one point appears to the left or right of the other when viewed from the target viewpoint.
        \item \textbf{Target-View Object Description.}  
    This task assesses whether an MLLM can accurately describe an object from the source image as it would appear from the target viewpoint, testing its ability to preserve semantic fidelity and fine-grained visual details---a capability that is often challenging to achieve with pixel-wise warping. As in the previous task, we identify two points visible in both $\mathbf{I}_S$ and $\mathbf{I}_T$ to construct $Q$, which asks the MLLM to describe an object, or a specific visual attribute of it, at the annotated position.
\end{itemize}
Examples from our~\benchmark{} are shown in Fig.~\ref{fig:benchmark}. Additional details on the benchmark construction are provided in~\supp{}.

\vspace{-0.5\baselineskip}
\paragraph{Metrics.}
For quantitative evaluation in the view-conditioned spatial reasoning task with binary labels—left or right—we report accuracy (\%), defined as the proportion of correctly answered VQA queries.
For the target-view object description task, we employ Qwen2.5-14B~\cite{bai2025qwen25vl} as an evaluator and ask it to rate the generated responses on a scale from 1 to 10. We compute the score for each example in our benchmark suite and report the average score as the performance metric. 
As a barometer for the reported metrics, we additionally compute and report an oracle performance metric obtained by using the ground-truth target-view image when answering the VQA queries.
Specifically, for the \texttt{Text} and \texttt{Shape} subsets, we retained only those data pairs on which the oracle was correct, yielding 571 and 744 pairs, respectively, for evaluation. We used 300 pairs for \texttt{Object}.

\section{Evaluation}
\label{sec:evaluation}
We evaluate the token warping techniques from~\secref{sec:method_problem_definition} on~\benchmark{}, with baselines summarized in~\secref{subsec:baselines}.
Results for view-conditioned spatial reasoning and target-view object description are presented in~\secref{subsec:view_conditioned_result} and~\secref{subsec:target_view_object_result}, respectively.

\subsection{Baselines}
\label{subsec:baselines}
We compare token warping against pixel-wise warping variants and external baselines.
For our framework, its variants, and the generative warping baseline, we use Qwen2.5-VL-7B~\cite{bai2025qwen25vl} as the base MLLM.
We implement both forward and backward pixel-wise warping, along with three variants of token warping, denoted \emph{Forward}, \emph{Backward-Nearest}, and \emph{Backward-Adaptive}, respectively.
These methods introduce minimal inference-time overhead for warping during inference without requiring extra fine-tuning.
In addition, we include specialized MLLMs fine-tuned on spatial reasoning datasets, such as SpatialReasoner~\cite{ma2025_spatialreasoner}, ViLaSR~\cite{wu2025vilasr}, and VLM-3R~\cite{fan2025vlm3r}. 
For these models, we provide the original source view together with an additional text prompt that explicitly describes the relative camera motion from the source to the target view.
Lastly, we employ GenWarp~\cite{seo2024genwarp}, a camera-conditioned diffusion model that uses implicit warping for novel view synthesis, to directly generate an RGB image at the target viewpoint and then pass it to Qwen2.5-VL~\cite{bai2025qwen25vl} for querying.
We provide comparisons against additional baselines in~\supp{}.

\subsection{View-Conditioned Spatial Reasoning}
\label{subsec:view_conditioned_result}
The quantitative results for the view-conditioned spatial reasoning task, including~\texttt{\benchmark{}-Text} and~\texttt{\benchmark{}-Shape}, are presented in columns 2–13 of~\tabref{tab:viewbench_real_gt_pred_pairs}.
As shown in the rows highlighted in~\textcolor{Gray}{gray}, backward token warping, regardless of the fetching strategy (nearest or adaptive), consistently outperforms forward token warping across all overlapping ratios.
For example, in the most challenging settings, \texttt{\benchmark{}-Text} (5–15) and \texttt{\benchmark{}-Shape} (5–15), where the source and target viewpoints share only minimal overlap, the Backward-Nearest variant improves accuracy by 14.57\%p and 12.4\%p, respectively, when ground-truth depth maps are used for warping. Similar trends are observed across all other configurations, highlighting that providing dense and regular positional embeddings to MLLMs is crucial for maintaining high performance under viewpoint changes, consistent with our analysis in~\secref{sec:token_warping}.
In addition, we observe that the simple nearest-fetching strategy performs on par with the adaptive variant—an effect we attribute to the robustness of token-level representations, which naturally preserve local semantics by treating groups of pixels as coherent units.

When compared against the pixel-wise warping variants (rows highlighted in~\textcolor{RubineRed}{red}), the specialized MLLMs (rows highlighted in~\textcolor{blue}{blue}), and the generative warping baseline (row highlighted in~\textcolor{ForestGreen}{green}), our token-wise warping approach consistently outperforms all of them.
Notably, VLM-3R~\cite{fan2025vlm3r}, which incorporates features from CUT3R~\cite{wang2025continuous}, still remains behind backward token warping, indicating that rich features alone do not equip models with the capacity to mentally shift viewpoints.

Qualitative examples in rows 1–4 of~\figref{fig:quali_benchmark} provide a visual explanation of this trend. Note that the pixelated images in the \textbf{Token Warping} columns are displayed solely for visualization; our framework operates entirely on token embeddings. In contrast, pixel-wise warping baselines feed the warped images, such as those illustrated in the figure, into the MLLM’s vision encoder.
As shown in row 2 of~\figref{fig:quali_benchmark}, pixel-wise warping introduces severe visual artifacts during both forward and backward warping, yielding incorrect predictions (\eg, ``left''). Even a generative approach~\cite{seo2024genwarp} does not fully resolve these issues, as it may hallucinate non-existent objects or lose existing ones. 
For instance, Fig.~\ref{fig:quali_benchmark} row 5 shows that the simple shapes in the input image are omitted in the output of GenWarp, therefore leading to the response ``none''.
In contrast, backward token-warping–based approaches consistently produce the correct answer.
We provide qualitative results for the warped images as well as the descriptions generated by MLLMs in~\supp{}.

\subsection{Target-View Object Description}
\label{subsec:target_view_object_result}
We summarize the quantitative results for the target-view object description task (\texttt{\benchmark{}-Object}) in columns 14--19 of~\tabref{tab:viewbench_real_gt_pred_pairs}.
Consistent with our analysis in~\secref{subsec:view_conditioned_result}, among the token-warping methods highlighted in~\textcolor{gray}{gray} rows, backward warping approaches outperform their forward-warping counterpart, as reflected in higher scores from the MLLM evaluator. The same trend holds when comparing token-warping approaches against the pixel-wise warping baselines (shown in~\textcolor{RubineRed}{red}) and the generative warping baseline (shown in~\textcolor{ForestGreen}{green}).
We report qualitative results for warped images, and the descriptions generated by MLLMs in~\supp{}.

\section{Conclusion}
In this work, inspired by classic discussions on part-based representations for mental imagery~\cite{shepard1971mental, minsky1974framework, pylyshyn1973mind, hinton1979mental}, we explored token warping as a simple yet effective strategy for transferring source view observations to nearby novel viewpoints. By comparing different token warping directions (\emph{forward} vs.\ \emph{backward}) and backward token fetching techniques (\emph{adaptive} vs.\ \emph{nearest}), we found that constructing a regular, dense grid of tokens via backward warping is crucial for robust MLLM performance. Notably, simple nearest fetching performs comparably to the more sophisticated adaptive fetching, offering a practical and efficient solution.

\newpage
\section*{Acknowledgements}
We thank Daehyeon Choi and Sangwoo Youn for their valuable discussions.
This work was supported by the National Research Foundation of Korea (NRF) (RS-2026-25486000); the Institute of Information \& Communications Technology Planning \& Evaluation (IITP) grants (RS-2019-II190075, RS-2022-00156435, RS-2024-00399817, RS-2025-25441313, RS-2025-25443318), funded by the Korean government (MSIT); the Industrial Technology Innovation Program (RS-2025-02317326), funded by the Korean government (MOTIE); the National Supercomputing Center (KSC-2025-CRE-0475); and the DRB-KAIST SketchTheFuture Research Center.

\clearpage
{
    \small
    \bibliographystyle{ieeenat_fullname}
    \bibliography{main}
}

\clearpage
\newpage

\def\confName{CVPR}
\def\confYear{2026}

\maketitlesupplementary

\appendix
\renewcommand{\thesection}{\Alph{section}}
\numberwithin{table}{section}
\renewcommand{\thetable}{\thesection\arabic{table}}
\numberwithin{figure}{section}
\renewcommand{\thefigure}{\thesection\arabic{figure}}

\noindent
In this supplementary material, we report additional experimental results with more baseline MLLMs and showcase qualitative examples of warped visualizations with corresponding MLLM responses (\secref{supp_sec:additional_results}).
We then present implementation and algorithmic details of \emph{backward token warping} with \emph{nearest} and \emph{adaptive} fetching (\secref{supp_sec:impl_details}).
Finally, we describe the step-by-step data construction pipeline of \texttt{ViewBench} in~\secref{supp_sec:viewbench_details}.
\section{Additional Results}
\label{supp_sec:additional_results}
This section presents additional experiments: extended comparisons with specialized MLLMs (\secref{supp_subsec:additional_baselines}), robustness analysis under estimated geometry (\secref{supp_subsec:robustness}), evaluation under extreme viewpoint shifts and occlusion (\secref{supp_subsec:large_shift}), a geometry-based oracle analysis (\secref{supp_subsec:oracle}), and qualitative examples (\secref{supp_subsec:additional_qualitatives}).

\begin{table*}[t!]
\smaller
\centering
\setlength{\tabcolsep}{3pt}
\begin{adjustbox}{width=\textwidth}
\begin{tabular}{l*{18}{c}@{}}
\toprule
 &
\multicolumn{6}{c}{\textbf{\texttt{ViewBench-Text} (\%)}} &
\multicolumn{6}{c}{\textbf{\texttt{ViewBench-Shape} (\%)}} &
\multicolumn{6}{c}{\textbf{\texttt{ViewBench-Object} (1-10)}} \\
\cmidrule(lr){2-7}\cmidrule(lr){8-13}\cmidrule(lr){14-19}
View Overlap (\%) 
& \multicolumn{2}{c}{5–15} & \multicolumn{2}{c}{15–25} & \multicolumn{2}{c}{25–35}
& \multicolumn{2}{c}{5–15} & \multicolumn{2}{c}{15–25} & \multicolumn{2}{c}{25–35}
& \multicolumn{2}{c}{5–15} & \multicolumn{2}{c}{15–25} & \multicolumn{2}{c}{25–35} \\
\cmidrule(lr){1-1} \cmidrule(lr){2-3}\cmidrule(lr){4-5}\cmidrule(lr){6-7}
\cmidrule(lr){8-9}\cmidrule(lr){10-11}\cmidrule(lr){12-13}
\cmidrule(lr){14-15}\cmidrule(lr){16-17}\cmidrule(lr){18-19}
Depth 
& GT & Pred. & GT & Pred. & GT & Pred.
& GT & Pred. & GT & Pred. & GT & Pred.
& GT & Pred. & GT & Pred. & GT & Pred. \\
\midrule

Target View (Oracle)
  & 100.00 & --& 100.00 & --& 100.00 & -
  & 100.00 & --& 100.00 & --& 100.00 & -
  & 6.64 & --& 7.31 & --& 7.43 & --\\

\midrule
\emph{Specialist MLLMs} &&&&&&&&&&&&&&&&&& \\

\cellcolor{blue!10} SpatialReasoner~\cite{ma2025_spatialreasoner}
  & 46.73 & --& 53.30 & --& 53.71 & --
  & 33.72 & --& 38.27 & --& 48.15 & --
  & --& --& --& --& --& --\\

\cellcolor{blue!10} VLM-3R~\cite{fan2025vlm3r}
  & 63.82 & --& 70.56 & --& 60.57 & --
  & 49.22 & --& 49.79 & --& 50.21 & --
  & --& --& --& --& --& --\\

\cellcolor{blue!10} ViLaSR~\cite{wu2025vilasr}
  & 44.22 & --& 52.28 & --& 48.00 & --
  & 22.87 & --& 23.05 & --& 34.57 & --
  & --& --& --& --& --& --\\

\cellcolor{blue!10} Qwen2.5-VL~\cite{bai2025qwen25vl}
  & 46.23 & --& 59.39 & --& 52.00 & --
  & 24.42 & --& 25.10 & --& 37.86 & --
  & --& --& --& --& --& --\\

\midrule
\cellcolor{orange!20} Qwen3-VL~\cite{yang2025qwen3}
& 41.71	& -- & 47.21 & -- & 45.14 &	--	
& 18.60	& -- & 22.22 & -- & 35.80 & --
& --& --& --& --& --& --\\

\cellcolor{orange!20} InternVL3~\cite{zhu2025internvl3}
& 56.28	& --	& 64.47	& --	& 61.71	& --
& 32.17	& --	& 38.68	& --	& 51.85	& --
& --& --& --& --& --& --\\

\cellcolor{orange!20} Cambrian-1~\cite{tong2024cambrian}
& 9.05	& --	& 11.68	& --	& 9.71	& --	
& 34.88	& --	& 34.57	& --	& 44.03	& --
& --& --& --& --& --& --\\

\cellcolor{orange!20} LLaVA-OneVision-1.5~\cite{an2025llavaonevision15}
& 48.24	& --	& 51.27	& --	& 61.71	& --
& 27.52	& --	& 30.04	& --	& 38.27	& --
& --& --& --& --& --& --\\

\cellcolor{orange!20} Kimi-VL-Thinking~\cite{team2025kimivl}
& 49.25	& --	& 54.31	& --	& 52.00	& --	
& 31.78	& --	& 37.86	& --	& 43.21	& --
& --& --& --& --& --& --\\

\cellcolor{orange!20} RoboBrain-2.0~\cite{team2025robobrain2}
& 37.69 &	--	& 43.65	& -- & 49.71 & -- 
& 22.48	& -- & 29.63 & -- & 39.92 & --
& --& --& --& --& --& --\\

\cellcolor{orange!20} VeBrain~\cite{luo2025vebrain}
& 49.25	& --	& 54.31 &	--	& 54.29	& --
& 29.84	& --	& 32.51	& --	& 47.33	& --
& --& --& --& --& --& --\\

\cellcolor{orange!20} SpaceQwen~\cite{chen2024spatialvlm}
& 68.34	& --	& 72.69	& --	& 62.86	 & --
& 48.06	& --	& 46.50	& --	& 49.38	& --
& --& --& --& --& --& --\\

\cellcolor{orange!20} SpaceThinker~\cite{chen2024spatialvlm}
& 48.74	& --	& 51.27	& --	& 48.57	& --
& 46.51	& --	& 47.74	& --	& 48.15	& --
& --& --& --& --& --& --\\

\cellcolor{orange!20} MindCube~\cite{yin2025mindcube}
& 59.30	& --	& 59.39	& --	& 57.14 & 	-- 
& 46.90	& --	& 47.74	& --	& 47.33	& --
& --& --& --& --& --& --\\

\cellcolor{orange!20} VST-RL~\cite{yang2025visualspatialtuning}
& 28.14	& --	& 34.01	& --	& 38.29	& --	
& 28.29	& --	& 26.34	& --	& 43.62	& --
& --& --& --& --& --& --\\

\cellcolor{orange!20} VST-SFT~\cite{yang2025visualspatialtuning}
& 42.71	& --	& 47.72	& --	& 46.29	& --	
& 28.29	& --	& 26.34	& --	& 43.62	& --
& --& --& --& --& --& --\\

\cellcolor{orange!20} SpaceR-SFT-7B~\cite{ouyang2025spacer}
& 67.84	& --	& 73.10	& --	& 64.00	& --	
& 44.96	& --	& 48.15	& --	& 53.09	& --
& --& --& --& --& --& --\\

\cellcolor{orange!20} SpatialLadder~\cite{li2025spatialladder}
& 70.35 & 	--	& 74.11	& --	& 67.43	& --
& 50.00	& --	& 49.38	& --	& 50.21	& --
& --& --& --& --& --& --\\

\cellcolor{orange!20} VG-LLM~\cite{zheng2025vgllm}
& 5.93	& --	& 13.20	& --	& 9.71	& --	
& 13.18	& --	& 14.40	& --	& 24.69	& --
& --& --& --& --& --& --\\

\midrule
\emph{Novel View Synthesis} &&&&&&&&&&&&&&&&&& \\

\cellcolor{ForestGreen!15} GenWarp~\cite{seo2024genwarp}
  & 69.35 & --& 71.07 & --& 66.29 & --
  & 53.10 & --& 47.33 & --& 55.14 & --
  & 4.32 & --& 4.81 & --& 4.34 & --\\

\midrule
\emph{Pixel-Wise Warping} &&&&&&&&&&&&&&&&&& \\

\cellcolor{RubineRed!15} Forward
  & 70.85 & 69.35 & 73.60 & 73.10 & 62.86 & 67.43
  & 56.20 & 56.20 & 56.79 & 56.79 & 60.49 & 60.08
  & 3.22 & 3.22 & 4.04 & 3.87 & 4.78 & 4.54 \\

\cellcolor{RubineRed!15} Backward
  & 71.86 & 67.84 & 75.63 & 74.62 & 68.57 & 68.57
  & 62.40 & 58.14 & 58.02 & 56.79 & 66.67 & 64.20
  & 4.53 & 4.45 & \underline{5.52} & 5.48 & 5.94 & 5.89 \\

\midrule
\emph{Token Warping} &&&&&&&&&&&&&&&&&& \\

\cellcolor{Gray!15} Forward
  & 60.30 & 66.83 & 64.47 & 65.48 & 54.86 & 60.57
  & 55.04 & 56.98 & 55.14 & 60.91 & 53.09 & 56.38
  & 4.09 & 4.20 & 4.27 & 4.37 & 4.07 & 3.78 \\

\cellcolor{Gray!15} Backward-Nearest
  & \underline{74.87} & \textbf{75.38} & \textbf{80.71} & \textbf{81.73} & \underline{74.86} & \textbf{76.00}
  & \textbf{67.44} & \underline{63.95} & \underline{62.96} & \textbf{62.55} & \underline{73.25} & \textbf{75.31}
  & \underline{4.80} & \underline{4.86} & 5.39 & \underline{5.57} & \textbf{6.19} & \underline{5.97} \\

\cellcolor{Gray!15} Backward-Adaptive
  & \textbf{77.89} & \underline{73.37} & \underline{79.70} & \underline{80.71} & \textbf{78.86} & \underline{74.29}
  & \textbf{67.44} & \textbf{66.28} & \textbf{66.26} & \underline{61.32} & \textbf{75.72} & \underline{70.37}
  & \textbf{4.97} & \textbf{5.18} & \textbf{5.76} & \textbf{6.29} & \underline{6.11} & \textbf{6.14} \\

\bottomrule
\end{tabular}
\end{adjustbox}

\caption{\textbf{Additional Quantitative Comparisons on~\benchmark{}.}
Extended table of Tab. 1~\inmain{}, with additional baseline MLLMs included in orange (\colorbox{orange!20}{\phantom{a}}). 
Columns 2-13 report accuracy (\%) on spatial reasoning tasks (\texttt{\benchmark{}-Text} and \texttt{\benchmark{}-Shape}), and columns 14-19 report target-view object description scores (\texttt{\benchmark{}-Object}), evaluated by Qwen2.5-VL-14B~\cite{bai2025qwen25vl} on a 1–10 scale. Across all tasks and setups, backward token-wise warping achieves the best performance.}
\label{supp_tab:additional_quantitative}

\vspace{-0.5\baselineskip}
\end{table*}

\subsection{Comparison with Additional Baselines}
\label{supp_subsec:additional_baselines}
Extending Tab.~1~\ofmain{}, we report a more extensive quantitative comparison against a wider range of specialist and general-purpose MLLMs.

\paragraph{Baselines.}
We include recent open-source MLLMs: \emph{Qwen3-VL}~\cite{yang2025qwen3}, \emph{InternVL3}~\cite{zhu2025internvl3}, \emph{Cambrian-1}~\cite{tong2024cambrian}, \emph{LLaVA-OneVision-1.5}~\cite{an2025llavaonevision15}, and \emph{Kimi-VL-Thinking}~\cite{team2025kimivl}.
We further include models explicitly fine-tuned for spatial reasoning via SFT and/or GRPO~\cite{guo2025deepseek}. 
\emph{RoboBrain-2.0}~\cite{team2025robobrain2} and \emph{VeBrain}~\cite{luo2025vebrain} extend Qwen2.5-VL~\cite{bai2025qwen25vl} with rich spatial task suites, while \emph{SpaceQwen}~\cite{spaceqwen} and \emph{SpaceThinker}~\cite{spacethinker} are Qwen2.5-VL variants fine-tuned on spatial VQA data~\cite{vqasynth} following data synthesis protocol of SpatialVLM~\cite{chen2024spatialvlm}.
For \emph{MindCube}~\cite{yin2025mindcube}, we used the \texttt{Plain-CGMap-FFR-Out} SFT variant, reported as the best-performing configuration by the authors.

We include models from \emph{VST}~\cite{yang2025visualspatialtuning}, a concurrent work that fine-tunes Qwen2.5-VL on a curated dataset spanning over 19 spatial tasks, comparing both their SFT \emph{(VST-SFT)} and RL-tuned \emph{(VST-RL)} variants. 
We further compare with a SFT variant of \emph{SpaceR}~\cite{ouyang2025spacer} and \emph{SpatialLadder}~\cite{li2025spatialladder}, a concurrent work employing a progressive SFT+GRPO training schedule for spatial reasoning.
Finally, we evaluate \emph{VG-LLM}~\cite{zheng2025vgllm}, which integrates a 3D geometry encoder initialized from VGGT~\cite{wang2025vggt} into an MLLM, similar to VLM-3R~\cite{fan2025vlm3r}~\inmain{}, which integrates CUT3R~\cite{wang2025continuous} features to provide strong 3D priors.

\paragraph{Results.}
Full results are shown in~\tabref{supp_tab:additional_quantitative}.
Consistent with Tab.~1~\ofmain{}, our backward token warping methods (\ie, \emph{Backward-Nearest} and \emph{Backward-Adaptive}) achieve the best performance on both \texttt{ViewBench-Text} and \texttt{ViewBench-Shape}, outperforming all baselines including the newly added models. 
Notably, recent state-of-the-art general MLLMs (\eg, Qwen3-VL~\cite{yang2025qwen3}, InternVL3~\cite{zhu2025internvl3}) still struggle to internally shift viewpoint to solve our tasks. Likewise, MindCube~\cite{yin2025mindcube}, despite being designed for multi-view spatial reasoning, shows clear limitations when required to reason about a single view from a nearby target viewpoint.
SpatialLadder~\cite{li2025spatialladder}, despite its carefully designed training curriculum, still underperforms our backward token warping, which explicitly and reliably transfers source-view information to the target viewpoint.

Lastly, VG-LLM~\cite{zheng2025vgllm}, which integrates rich 3D features from VGGT~\cite{wang2025vggt}, exhibits highly degraded behavior: the model frequently outputs multiple-choice labels (\eg, \emph{``A''}, \emph{``B''}) even when prompted to answer with \emph{``left''} or \emph{``right''}. 
We hypothesize that the VGGT-based fine-tuning phase may have compromised the base MLLM's general capabilities, whereas our token warping approach leaves the underlying MLLM unchanged, better preserving its original abilities.

\subsection{Robustness Analysis on Estimated Geometry}
\label{supp_subsec:robustness}
Our token warping framework relies on the depth map $\mathbf{D}$ and relative camera pose $\mathbf{\Pi}_{T \rightarrow S}$ to compute the backward warping function $f_{T \rightarrow S}$ (Eq.~\ref{eq:backward_warping}).
A natural concern is whether the method remains effective when geometric inputs are estimated rather than ground-truth. 
We evaluate this on \texttt{ViewBench-Shape} by replacing the ground-truth geometry with predictions from off-the-shelf models.

\paragraph{Depth Estimation.}
We compare ground-truth depth (GT) against predictions from two monocular depth estimators: Depth Anything v2 (DA-V2)~\cite{yang2024depth} and Depth Pro (DP)~\cite{bochkovskii2024depth}. 
We additionally include a no-warping reference baseline (Ref.) using the base Qwen2.5-VL~\cite{bai2025qwen25vl} on the source image.
As shown in~\tabref{supp_tab:depth_pose_robust}, backward token warping with adaptive fetching achieves 65.84\% with DA-V2 and 67.74\% with DP, compared to 70.99\% with GT depth. 
Pixel-wise backward warping follows the same trend, dropping from 62.35\% (GT) to 60.49\% (DA-V2) and 62.76\% (DP).
In both cases, warping with estimated geometry substantially outperforms the no-warping baseline, confirming that the gains from warping persist even without ground-truth geometry.
Importantly, the performance gap between token warping and pixel-wise warping is preserved regardless of the depth source, indicating that the advantage of operating in token space is orthogonal to improvements in depth estimation quality.

\begin{table}[b!]
\vspace{-\baselineskip}
\centering
\small
\setlength{\tabcolsep}{3pt}
\resizebox{\columnwidth}{!}{%
\begin{tabular}{l|c|cc|cc|c}
\toprule
 & \textbf{GT} & \multicolumn{2}{c|}{\cellcolor{Green!10} \textbf{Depth}} & \multicolumn{2}{c|}{\cellcolor{Blue!10} \textbf{Depth \& Pose}} & \textbf{Ref.} \\
 & & \textbf{DA-V2} & \textbf{DP} & \textbf{VGGT} & \textbf{DUSt3R} & \\
\midrule
\cellcolor{red!8}
Pixel-Wise Warp. & 62.35 & 60.49 & 62.76 & 63.58 & 61.29 & \multirow{2}{*}{31.48} \\
\cellcolor{gray!15}
\textbf{Token Warp.} & 70.99 & 65.84 & 67.74 & 68.95 & 65.05 & \\
\bottomrule
\end{tabular}%
}
\vspace{-0.5\baselineskip}
\caption{\textbf{Robustness to Estimated Geometry.} Accuracy (\%) on \texttt{ViewBench-Shape} (averaged across all overlap levels). \emph{Ref.}  is a no-warping baseline with base Qwen2.5-VL~\cite{bai2025qwen25vl}.}
\label{supp_tab:depth_pose_robust}
\end{table}

\paragraph{Joint Depth and Pose Estimation.}
We further evaluate a more challenging setting where \emph{both} depth and relative pose are predicted from an image pair, using VGGT~\cite{wang2025vggt} and DUSt3R~\cite{wang2024dust3r}. 
As reported in~\tabref{supp_tab:depth_pose_robust}, token warping with VGGT-estimated geometry achieves 68.95\%, compared to 63.58\% for pixel-wise warping under the same conditions. 
With DUSt3R, both methods decline further, yet token warping still outperforms pixel-wise warping.
These results confirm that the conclusions of Tab.~1~\ofmain{} hold under realistic conditions where ground-truth geometry is unavailable.

\subsection{Larger Viewpoint Shifts and Occlusion}
\label{supp_subsec:large_shift}
To stress-test our method beyond the overlap ranges in Sec.~5~\ofmain{} (5--35\%), we construct two additional evaluation splits targeting extreme viewpoint shifts and occlusion.

\paragraph{Larger Viewpoint Shifts.}
We sample source--target pairs from ScanNet~\cite{dai2017scannet} with very low overlap (2--5\%), representing nearly disjoint views where only a small portion of the scene is shared. 
As shown in~\tabref{supp_tab:large_shift}, backward token warping with adaptive fetching achieves 65.08\% with GT depth and 66.14\% with estimated depth, substantially outperforming pixel-wise backward warping (61.90\% / 61.38\%) and the no-warping baseline (34.39\%).
The consistent trend across all overlap levels suggests that the advantages of token-level warping are not confined to moderate viewpoint changes.

\begin{table}[h!]
\centering
\small
\setlength{\tabcolsep}{6pt}
\begin{tabular}{l|cc}
\toprule
 \multicolumn{1}{c}{Depth}
 & GT & Pred. \\
\midrule
\cellcolor{blue!8}
Qwen2.5-VL~\cite{bai2025qwen25vl} & 34.39 & -- \\
\cellcolor{red!8}
Pixel-Wise Warp. & 61.90 & 61.38 \\
\cellcolor{gray!15}
\textbf{Token Warp.} & 65.08 & 66.14 \\
\bottomrule
\end{tabular}
\caption{\textbf{Larger Viewpoint Shift (2--5\% Overlap).} Accuracy (\%) on a stress-test split with extremely low view overlap, where the source and target views share only 2--5\% of visible scene content.}
\label{supp_tab:large_shift}
\end{table}

\begin{figure}[h!]
  \centering
  \includegraphics[width=\columnwidth]{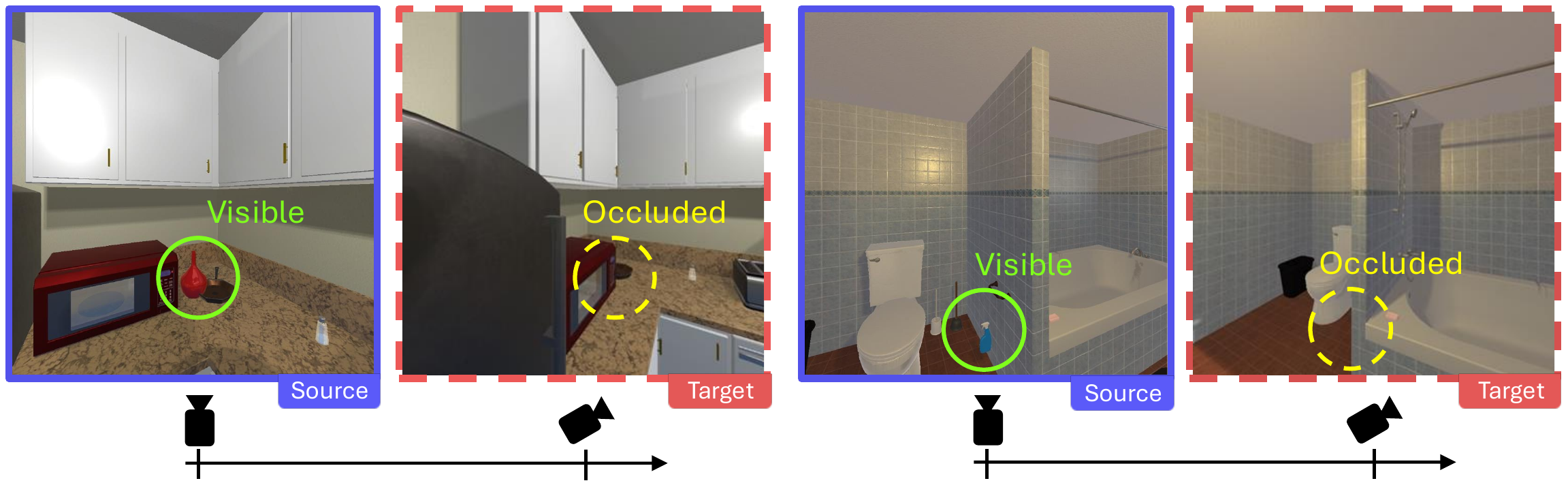}
  \vspace{-\baselineskip}
  \caption{\textbf{Occlusion Evaluation.} Example source--target pairs from the ProcTHOR-based~\cite{deitke2022procthor} occlusion split, where a visible object in the source view becomes fully occluded in the target view.}
  \label{supp_fig:occlusion}
\end{figure}

\paragraph{Occlusion.}
We also collect synthetic image pairs using ProcTHOR~\cite{deitke2022procthor} where an object visible from the source view becomes \emph{fully occluded} at the target viewpoint.
This tests whether warping helps the model reason about visibility changes under viewpoint shifts.
As shown in Fig.~\ref{supp_fig:occlusion}, token warping achieves 46\% accuracy with GT depth, compared to 38\% for pixel-wise warping and 32\% for the base Qwen2.5-VL~\cite{bai2025qwen25vl}, evaluated on 50 pairs with GT depth.
While absolute accuracies are lower due to the difficulty of reasoning under full occlusion, the relative ordering is consistent with our main findings: token warping provides a more reliable basis for viewpoint reasoning even under significant visibility changes.

\begin{table}[h!]
\centering
\small
\setlength{\tabcolsep}{6pt}
\begin{tabular}{l|c}
\toprule \multicolumn{1}{c}{Depth} & GT \\
\midrule
\cellcolor{blue!8}
Qwen2.5-VL~\cite{bai2025qwen25vl} & 32.00 \\
\cellcolor{red!8}
Pixel-Wise Warp. & 38.00 \\
\cellcolor{gray!15}
\textbf{Token Warp.} & 46.00 \\
\bottomrule
\end{tabular}
\caption{\textbf{Occlusion Evaluation.} Accuracy (\%) on a ProcTHOR-based~\cite{deitke2022procthor} split where the queried object is fully occluded in the target view. Token warping consistently outperforms pixel-wise warping and the base Qwen2.5-VL~\cite{bai2025qwen25vl}.}
\label{supp_tab:occlusion}
\end{table}

\subsection{Geometry-Based Oracle}
\label{supp_subsec:oracle}
To verify the reliability of the geometric pipeline underlying our token warping, we implement a \emph{geometry-based oracle} that bypasses the MLLM entirely.
Given a source--target pair, the oracle applies the backward warping function $f_{T \rightarrow S}$ (Eq.~\ref{eq:backward_warping}) to the two annotated keypoints in the source image and determines their left--right ordering by directly comparing the $x$-coordinates of the warped points, without querying the MLLM.

As shown in Tab.~\ref{supp_tab:geo_oracle}, the geometry-based oracle achieves above 93\% across all overlap levels for both \texttt{ViewBench-Text} and \texttt{ViewBench-Shape}. 
The small gap from 100\% is attributable to occasional depth noise near object boundaries and edge cases where the two keypoints project to nearly identical $x$-coordinates in the target view. 
These results confirm that the warping geometry is highly accurate, and that the remaining gap between our token warping methods (Tab.~1~\ofmain{}) and the oracle is primarily due to limitations in the MLLM's perception and reasoning capabilities rather than geometric errors.

\begin{table}[h!]
\centering
\small
\setlength{\tabcolsep}{4pt}
\resizebox{\columnwidth}{!}{%
\begin{tabular}{lcccccc}
\toprule
 & \multicolumn{3}{c}{\textbf{\texttt{Text (\%)}}} & \multicolumn{3}{c}{\textbf{\texttt{Shape (\%)}}} \\
 \cmidrule(lr){2-4}\cmidrule(lr){5-7}
View Overlap (\%) & 5--15 & 15--25 & 25--35 & 5--15 & 15--25 & 25--35 \\
\midrule
\textbf{Oracle (Geometry)} & 93.78 & 93.81 & 93.64 & 95.26 & 94.54 & 93.75 \\
\bottomrule
\end{tabular}%
}
\vspace{-0.5\baselineskip}
\caption{\textbf{Geometry-Based Oracle.} Accuracy (\%) of a geometry-only baseline that determines left--right ordering by comparing $x$-coordinates of the warped source keypoints.}
\label{supp_tab:geo_oracle}
\end{table}

\subsection{Additional Qualitative Results}
\label{supp_subsec:additional_qualitatives}
\begin{figure*}[h!]
  \centering
  \includegraphics[width=\linewidth]{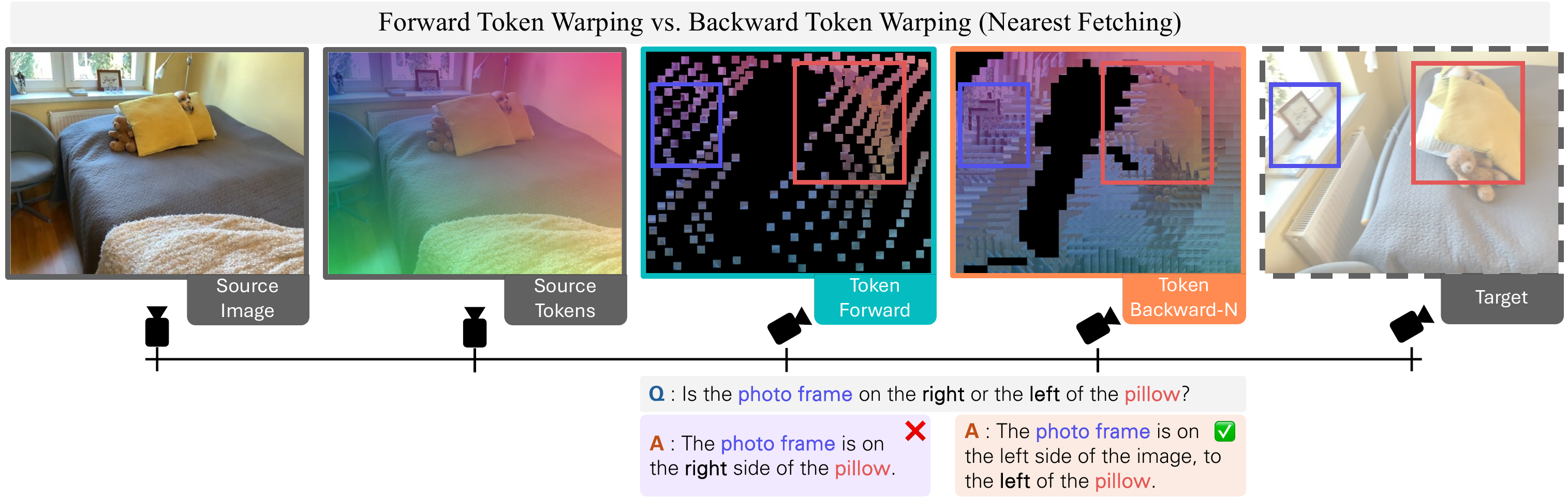}
  \caption{\textbf{Qualitative Sample 1.} 
    Given the source image (leftmost), the question asks for the spatial relationship between the photo frame (blue box) and the pillow (red box) as viewed \emph{from the target viewpoint} (rightmost). \textbf{To visualize tokens, we color-code each source token by its $(x,y)$ position in the source image, and preserve this color after warping, so the color of each token in the warped views indicates its source location.} With \textbf{forward token warping}, the projected tokens become sparse and irregular, leading the MLLM to answer incorrectly. In contrast, \textbf{backward token warping} with \textbf{nearest fetching} produces a dense, regular target token grid, allowing the model to correctly infer the spatial relationship from the target view. (Source and target images are from ARKitScenes~\cite{baruch2021arkitscenes}.)}
  \label{fig:supp_qualitative_1}
\end{figure*}

\begin{figure*}[t!]
  \centering
  \includegraphics[width=\linewidth]{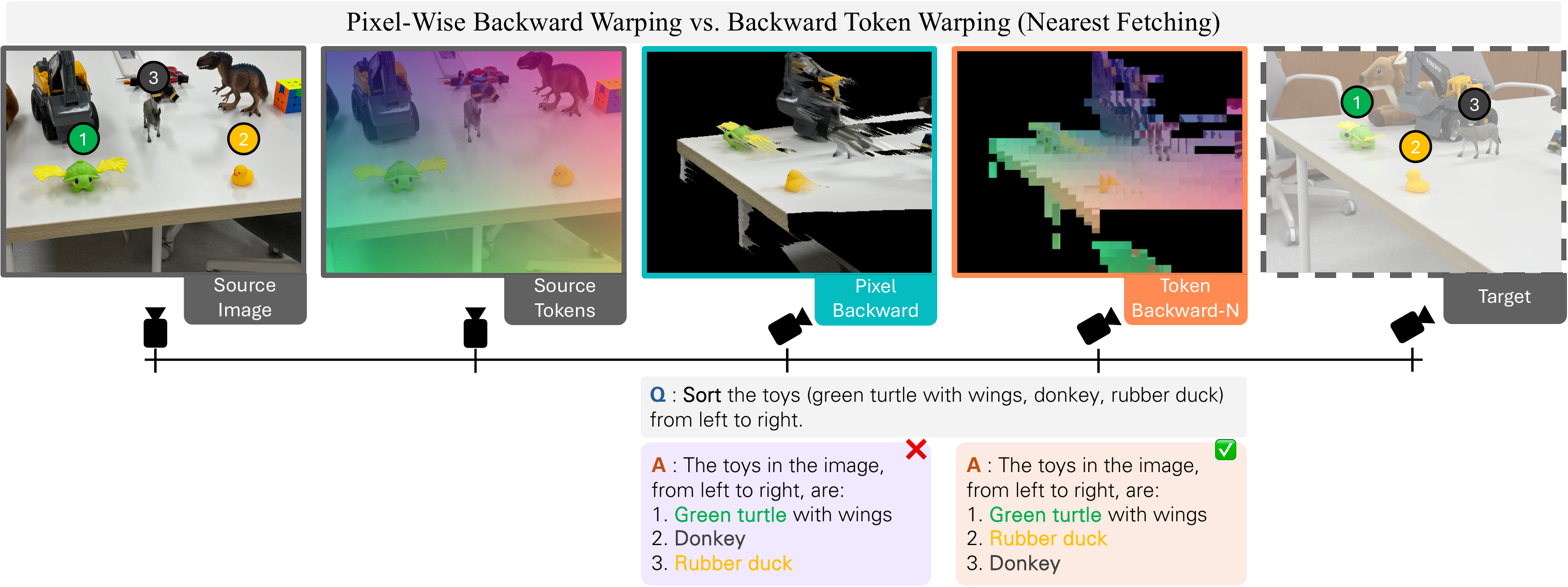}
  \caption{\textbf{Qualitative Sample 2.}
    Given the source image (leftmost), the question asks for the order of the toys from left to right \emph{as seen from the target viewpoint} (rightmost). \textbf{To visualize tokens, we color-code each source token by its $(x,y)$ position in the source image, and preserve this color after warping, so the color of each token in the warped views indicates its source location.} With \textbf{pixel-wise backward warping}, the target-view image suffers from local pixel distortions caused by depth noise, leading the MLLM to answer incorrectly. In contrast, \textbf{backward token warping} with \textbf{nearest fetching} preserves the semantic content while shifting viewpoint, allowing the MLLM to produce the correct ordering of the toys. (Source and target images were captured manually.)}
  \label{fig:supp_qualitative_2}
\end{figure*}

\begin{figure*}[t!]
  \centering
  \includegraphics[width=\linewidth]{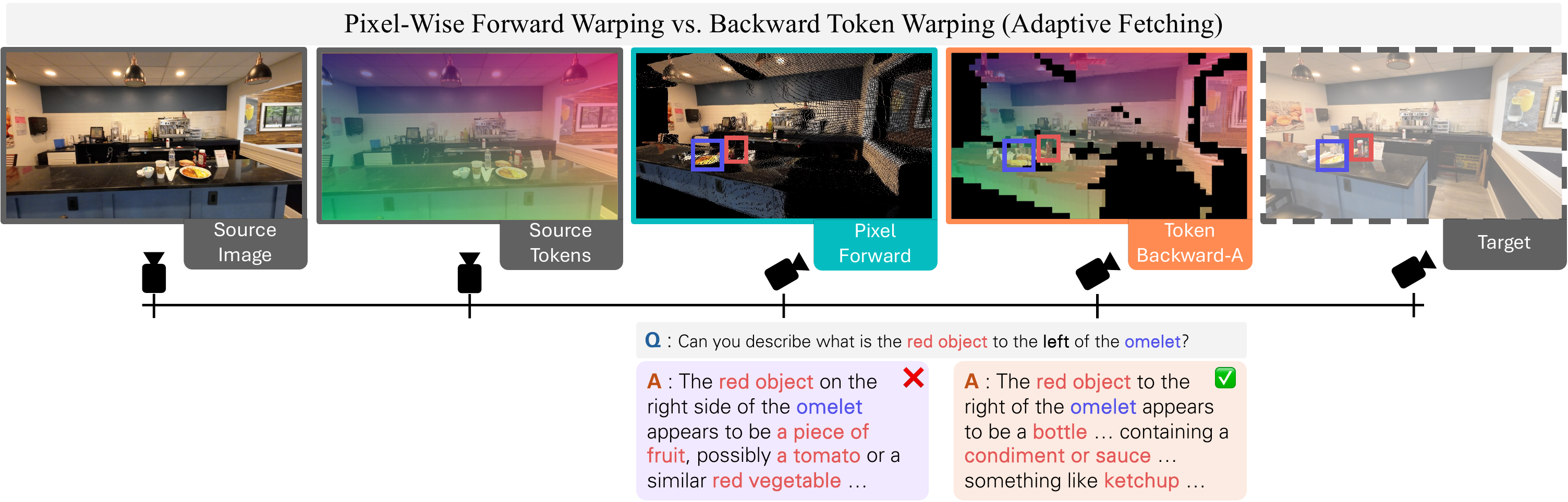}
  \caption{\textbf{Qualitative Sample 3.}
    Given the source image (leftmost), the question asks to describe the \emph{red object} (red box) placed on the left side of the omelet (blue box) \emph{when viewed from the target viewpoint} (rightmost). \textbf{To visualize tokens, we color-code each source token by its $(x,y)$ position in the source image, and preserve this color after warping, so the color of each token in the warped views indicates its source location.} When using \textbf{pixel-wise forward warping}, the warped image exhibits local pixel distortions due to depth prediction noise and holes caused by magnification. Consequently, given this warped RGB image, the MLLM incorrectly answers that the object is \emph{“a piece of fruit”}. In contrast, with \textbf{backward token warping} and \textbf{adaptive fetching}, the MLLM correctly identifies the object as a \emph{“bottle”}, more specifically \emph{“containing a condiment or sauce”} and \emph{“ketchup”}. This further highlights the advantage of warping in token space rather than pixel space when transferring source content to a target view. (Source and target images are from DL3DV-10K~\cite{ling2024dl3dv}.)}
  \label{fig:supp_qualitative_3}
\end{figure*}

\begin{figure*}[t!]
  \centering
  \includegraphics[width=\linewidth]{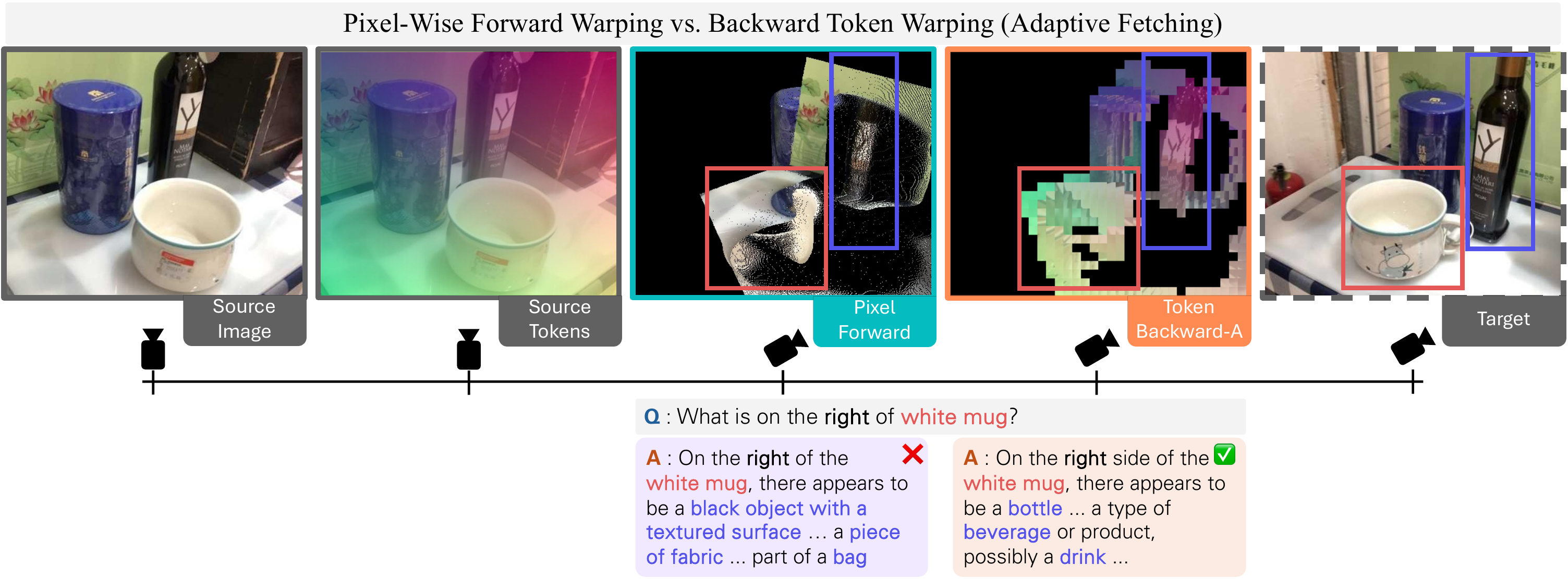}
  \caption{\textbf{Qualitative Sample 4.}
    Given the source image (leftmost), the question asks to describe the object that is located on the right side of the white mug (red box) \emph{when viewed from the target viewpoint} (rightmost). \textbf{To visualize tokens, we color-code each source token by its $(x,y)$ position in the source image, and preserve this color after warping, so the color of each token in the warped views indicates its source location.} With \textbf{pixel-wise forward warping}, the warped image shows distorted local details due as the forward warping distributes the source image pixels to a sparse grid in the target image. Consequently, the MLLM fails to accurately describe the bottle on the right side, and instead replies \emph{``piece of fabric''} and \emph{``part of a bag''}, which are not visible in the target image. On the other hand, when using \textbf{backward token warping} with \textbf{adaptive fetching}, the MLLM describes the specified object as \emph{``a bottle''} and \emph{``a type of beverage''} which is accurate when seen from the target image. These results again show that our proposes backward token warping can provide a robust way of transferring source image information to the target viewpoint. (Source and target images are from BLINK~\cite{fu2024blinkmultimodallargelanguage}.)}
  \label{fig:supp_qualitative_4}
\end{figure*}

We provide additional qualitative comparisons of our backward token warping with multiple baselines---including pixel-wise warping and forward token warping---on single-view VQA examples that require reasoning under viewpoint changes.
The visualizations are shown in Figs.~\ref{fig:supp_qualitative_1}--~\ref{fig:supp_qualitative_4}, with brief descriptions provided below.
For each case, we are given the source image, its depth map, the relative camera pose from source to target, and the camera intrinsics. 
To obtain the depth and poses, we run VGGT~\cite{wang2025vggt} on the source and target view images.

\clearpage
\newpage

\clearpage
\newpage

\section{Implementation Details}
\label{supp_sec:impl_details}
This section extends Sec.~3.3~\ofmain{} and details the implementation of our backward token warping framework, which enables MLLMs to reason under viewpoints changes from a single source image, its depth map, and relative camera pose.
For clarity, in this section we use ``$\mathbf{c}$'' to denote coordinates in the \emph{source} view and ``$\mathbf{g}$'' to denote coordinates in the \emph{target} view.

\subsection{Details on Backward Token Warping}
\label{supp_subsec:backward_warping}
Recall that in backward warping, we define a dense, regular grid in the target view and  fetch the corresponding tokens from the source image $\mathbf{I}$ via the target-to-source mapping $f_{T \rightarrow S}$.

\paragraph{Target Grid.}
For an image $\mathbf{I} \in \mathbb{R}^{H \times W \times 3}$, we impose a regular patch grid of size $l \times l$, yielding $M = (HW)/l^2$ patches\footnote{We assume $H$ and $W$ are divisible by $l$.}.
We denote by $\mathbf{g} \in \mathbb{R}^{M \times 2}$ the set of target-grid centers on the image plane, where each $\mathbf{g}_j$ specifies a location at which we wish to place a token sampled from the source image.
In backward token warping, our goal is to assign exactly one token to each grid center. 
For simplicity, we assume the target image has the same resolution as the source.

\paragraph{Source Proxy from Depth.}
Because the target-view image is unobserved, we cannot directly compute target-to-source correspondences.
Instead, we construct a lightweight 3D triangle mesh $\mathcal{M}_S$ from the source depth map $\mathbf{D}\in\R^{H\times W\times 1}$. 
Specifically, for each pixel $\mathbf{p}_i = (u_i, v_i)$ in $\mathbf{I}$ with its depth $d_i$ from $\mathbf{D}$, we unproject it using the $3\times 3$ intrinsic matrix $\mathbf{K}_{3\times 3}$ to obtain a 3D point:
\begin{align}
\mathbf{x}_i = d_i \mathbf{K}^{-1}_{3\times 3} \tilde{\mathbf{p}}_i , 
\quad \text{where } \tilde{\mathbf{p}}_i = [u_i, v_i, 1]^\top.
\end{align}
Here, $\mathbf{x}_i = [ x_i, y_i, z_i ]^\top$. 
We then triangulate every $2 \times 2$ pixel cell into two triangles, forming $\mathcal{M}_S$ in the source camera frame.

\paragraph{Backward Mapping via Ray Casting.}
For each target grid center $\mathbf{g}_j$, we cast a ray from the target camera using its pose $\Pi_T \in \mathbb{R}^{4 \times 4}$ and intrinsics $\mathbf{K} \in \mathbb{R}^{4 \times 4}$, and intersect it with the proxy mesh $\mathcal{M}_S$, obtaining a 3D hit point in the target frame, $\mathbf{x}_j^{\ast} \in \mathbb{R}^3$. We then express this point in homogeneous coordinates and project it back into the source image using the relative pose $\Pi_{T\rightarrow S} = \Pi_S \Pi_T^{-1}$ and intrinsics $\mathbf{K}$:
\begin{align}
\tilde{\mathbf{p}}_j^{*} &= \mathbf{K} \, \Pi_{T\rightarrow S} \, \tilde{\mathbf{x}}_j^{\ast},  \quad \text{where} \; \tilde{\mathbf{x}}_j^{\ast} = [\mathbf{x}_j^{\ast}, 1]^\top, \\
\mathbf{g}_j^{\ast} &= \pi\!\left( \tilde{\mathbf{p}}_j^{*} \right),
\label{eq:ray_casting}
\end{align}
where $\pi([u,v,w,1]^\top) = (u/w, v/w)^\top$ denotes perspective projection. The resulting $\mathbf{g}_j^{\ast} \in \mathbb{R}^2$ is a coordinate on $\mathbf{I}$ and serves as the backward mapping from target to source.
If no valid intersection is found (e.g., due to occlusion or field-of-view mismatch), we mark $\mathbf{g}_j^{\ast}$ as invalid and omit the corresponding patch.

By applying Eq.~\ref{eq:ray_casting} for every target grid center $\mathbf{g}_j \in \mathbf{g}$, we obtain the set of backward-warped coordinates on the source image, $\mathbf{g}^{\ast} \in \mathbb{R}^{M \times 2}$.  
Consistent with Eq.~3.1~\inmain{}, we denote this backward warping process as
\begin{align}
\mathbf{g}^{\ast} = f_{T\rightarrow S}\!\left(\mathbf{g}, \Pi_{T\rightarrow S}, \mathbf{K}, \mathbf{D}\right).
\label{eq:backward_warping}
\end{align}
Given $f_{T\rightarrow S}$, which provides a coordinate for every target grid center, the final step is to \emph{fetch} the corresponding tokens from the source image at these locations. We provide details on the fetching strategies in the next section.

\subsection{Nearest vs.\ Adaptive Fetching}
\label{supp_subsec:fetching}
We now detail the \emph{nearest} and \emph{adaptive} token fetching strategies used in the final step of backward token warping.

\paragraph{Nearest Fetching.}
Recall from Sec. 3.1~\inmain{} that source image $I$ is partitioned into a fixed, non-overlapping grid of patches $\{ \mathbf{u}_i \}_{i=1}^{M}$.
Let the source image $\mathbf{I}$ be patchified on a fixed grid, and let $\mathbf{c} \in \mathbb{R}^{M \times 2}$ denote the set of source grid centers, where $M$ is the number of patches.
Given a target grid center $\mathbf{g}_j$ and its backward-warped source coordinate $\mathbf{g}_j^{\ast}$ from $f_{T\rightarrow S}$, \emph{nearest fetching} selects the existing source patch whose center is closest to $\mathbf{g}_j^{\ast}$ in Euclidean distance:
\begin{align}
    i' = \operatorname*{\arg\min}_i \big\| \; \mathbf{g}_j^{\ast} - \mathbf{c}_i \; \big\|_2.
\end{align}
We then assign to $\mathbf{g}_j$ the token that was derived from the patch $\mathbf{u}_{i'}$ centered at $\mathbf{c}_{i'}$. While this introduces a small mismatch as $\mathbf{g}_j^{\ast}$ may not not coincide with any $\mathbf{c}_i$ in most cases, it allows us to reuse the original, efficient fixed-grid patchification for the source image.

\paragraph{Adaptive Fetching.}
Alternatively, we further implement \emph{adaptive fetching}, which re-patchifies the source image $\mathbf{I}$ according to the backward-warped coordinates $\mathbf{g}^{\ast}$ so that each patch is centered exactly at $\mathbf{g}_j^{\ast}$ with size $l \times l$.  
For each $\mathbf{g}_j^{\ast}$, we obtain a patch $\bar{\mathbf{u}}_j$ via
\begin{align}
    \bar{\mathbf{u}}_j = \mathrm{Crop}(\mathbf{I}, \mathbf{g}_j^{\ast}),
    \quad \bar{\mathbf{u}}_j \in \mathbb{R}^{l \times l \times 3},
\end{align}
where $\mathrm{Crop}(\mathbf{I}, \mathbf{g}_j^{\ast})$ extracts an $l \times l$ patch from $\mathbf{I}$ centered at $\mathbf{g}_j^{\ast}$.
Applying this to all $\mathbf{g}_j^{\ast} \in \mathbf{g}^{\ast}$ yields a new set of \emph{adaptive} patches $\{ \bar{\mathbf{u}}_j \}_{j=1}^{M}$ that replaces the original fixed-grid patches $\{ \mathbf{u}_i \}_{i=1}^{M}$. Finally, we assign to each target grid center $\mathbf{g}_j$ the token derived from its corresponding adaptive patch $\bar{\mathbf{u}}_j$, which is explicitly centered at $\mathbf{g}_j^{\ast}$.
Intuitively, this approach more faithfully respects the precise backward mappings in $f_{T\rightarrow S}$, at the cost of re-patchifying the image rather than relying on the original, efficient fixed-grid partitioning.

\section{Details on ViewBench}
\label{supp_sec:viewbench_details}
In this section, we provide additional details on the data synthesis protocol and evaluation metrics for \texttt{ViewBench}, introduced in Sec.~4~\inmain{}.

\subsection{Benchmark Construction}
\label{supp_subsec:viewbench_construction}
We construct \texttt{ViewBench} from real indoor scenes in ScanNet~\cite{dai2017scannet}, which provides dense RGB-D frames along with ground-truth depth, camera poses, and intrinsics.
For evaluations with estimated depth maps, we use Depth Anything v2~\cite{yang2024depth}.
To sample two-view pairs with controlled overlap, we use the MultiSPA data engine from~\citet{xu2025_multispatialmllm}, originally introduced for generating multi-view VQA data.
We adopt the same notions of \emph{visible points} and \emph{overlap ratio} as in MultiSPA and use them to construct \texttt{ViewBench} questions.
Below, we detail the benchmark construction procedure, following the notation of MultiSpa~\cite{xu2025_multispatialmllm}.

\paragraph{Overlap Computation.}
For each ScanNet scene~\cite{dai2017scannet}, we are given a 3D point cloud
\begin{align}
    \mathbf{P}_{\text{scene}} = \{ \mathbf{p}^w \}, 
    \quad \text{where } \mathbf{p}^w = [x^w, y^w, z^w]^\top,
\end{align}
with each point $\mathbf{p}^w$ expressed in the world coordinate system.
Each RGB frame $\mathbf{I}_i \in \mathbb{R}^{H \times W \times 3}$ is associated with a depth map $\mathbf{D}_i \in \mathbb{R}^{H \times W \times 1}$, an extrinsic matrix $\mathbf{E}_i \in \mathbb{R}^{4 \times 4}$, and an intrinsic matrix $\mathbf{K}_i \in \mathbb{R}^{4 \times 4}$.  

The extrinsic matrix is defined as
\begin{align}
    \mathbf{E}_i :=
    \begin{bmatrix}
        \mathbf{R}_i & \mathbf{t}_i \\
        \mathbf{0}^\top & 1
    \end{bmatrix},
    \quad \mathbf{R}_i \in \mathbb{R}^{3 \times 3},\; \mathbf{t}_i \in \mathbb{R}^{3 \times 1},
\end{align}
where $\mathbf{R}_i$ and $\mathbf{t}_i$ denote the camera rotation and translation, respectively.

Following MultiSPA~\cite{xu2025_multispatialmllm}, we map each world point $\mathbf{p}^w$ into the $i$-th camera coordinate system via
\begin{align}
    \tilde{\mathbf{p}}_i^c = (\mathbf{E}_i)^{-1} \tilde{\mathbf{p}}^w,
    \quad \text{where } \tilde{\mathbf{p}}^w = [\mathbf{p}^w, 1]^\top,
\end{align}
and denote $\tilde{\mathbf{p}}_i^c = [x_i^c, y_i^c, z_i^c, 1]^\top$.
We then project this point to the image plane :
\begin{align}
    \begin{bmatrix}
    u\\
    v\\
    1
    \end{bmatrix} = \frac{\mathbf{K}_i}{z_i^c} \begin{bmatrix}
        x_i^c \\
        y_i^c \\
        z_i^c
    \end{bmatrix},
    \quad 
    \label{eq:camera_projection}
\end{align}
We define the set of \emph{visible points} in frame $i$ as:
\begin{align}
    \mathcal{V}_i
    =
    \left\{
        \mathbf{p}^w \in \mathbf{P}_{\text{scene}}
        \;\middle|\;
        0 < z_i^c < d_i(u, v)
    \right\},
\end{align}
where $d_i(u, v)$ is the depth value at pixel $(u, v)$ from $\mathbf{D}_i$.  
This captures points whose projections fall inside $\mathbf{I}_i$ and are not occluded according to $\mathbf{D}_i$, which is identical to the visibility criterion of MultiSPA~\cite{xu2025_multispatialmllm}.

Finally, given two frames $\mathbf{I}_i$ and $\mathbf{I}_j$, we measure how much of the scene they see in common using the IoU of their visible point sets, defining the \emph{overlap ratio}~\cite{xu2025_multispatialmllm}:
\begin{align}
    \mathrm{Overlap}(i,j)
    =
    \frac{\left| \mathcal{V}_i \cap \mathcal{V}_j \right|}
         {\left| \mathcal{V}_i \cup \mathcal{V}_j \right|}.
\end{align}
We use this overlap ratio to create controlled splits in \texttt{ViewBench}.

\paragraph{Two-View Pair Selection.}
For each ScanNet scene, we enumerate candidate frame pairs and compute the overlap ratio defined above.  
We retain two-view a pair $(\mathbf{I}_S, \mathbf{I}_T)$ as a candidate if $\mathrm{Overlap}(S,T)$ lies in a moderate range (approximately 5--35\%), so that the two views are neither nearly identical nor almost disjoint.  
Following the overlap-aware sampling strategy of MultiSPA~\cite{xu2025_multispatialmllm}, we bin all non-zero-overlap pairs by their overlap ratio and sample an approximately equal number of pairs from each bin to mitigate the natural long-tailed bias toward small overlaps.
We then group the selected pairs into three overlap levels: \textbf{5--15\%}, \textbf{15--25\%}, and \textbf{25--35\%}. This categorization allows us to systematically study how viewpoint-conditioned reasoning changes as the amount of shared scene content varies.

\paragraph{Point Annotation.}
For each selected source–target pair $(\mathbf{I}_S, \mathbf{I}_T)$, we focus on the points that are visible in \emph{both} views, that is, the co-visible set
\(
\mathcal{V}_S \cap \mathcal{V}_T.
\)
For any $\mathbf{p}^w$ in this intersection, we obtain its camera-frame coordinates in each view via
\begin{align}
    \tilde{\mathbf{p}}_S^c &= (\mathbf{E}_S)^{-1} \tilde{\mathbf{p}}^w, \quad
    \tilde{\mathbf{p}}_T^c = (\mathbf{E}_T)^{-1} \tilde{\mathbf{p}}^w,
\end{align}
and then project them to the image planes using the same camera model as in Eq.~\ref{eq:camera_projection}.
These co-visible projections form the pool of candidate keypoints used to construct task-specific questions, analogous to the \emph{visual correspondence} subset construction in MultiSPA~\cite{xu2025_multispatialmllm}.

For \texttt{ViewBench-Text}, we randomly sample two co-visible points and annotate them with alphabet labels (\ie, A/B).
For \texttt{ViewBench-Shape}, we instead mark them with simple geometric symbols (e.g., triangle, star).
In all cases, annotations in the two views are guaranteed to correspond to the same underlying 3D locations.
============

\paragraph{Selecting View-Dependent Point Pairs.}
Given a source–target pair $(\mathbf{I}_S, \mathbf{I}_T)$ and its co-visible point set $\mathcal{V}_S \cap \mathcal{V}_T$, we construct left–right queries by sampling two co-visible 3D points and projecting them into both images (using the same intrinsics, extrinsics, and visibility checks as above). Let $u_A^S, u_B^S$ and $u_A^T, u_B^T$ denote the $u$-coordinates of the two keypoints (A and B) in the source and target views, respectively. We retain a pair only if
\begin{align}
    (u_A^S - u_B^S)\,(u_A^T - u_B^T) < 0
    \quad\text{and}\quad
    |u_A^T - u_B^T| \ge \tau,    
\end{align}
with $\tau = 50$ pixels to avoid near-vertical alignments. Thus, we keep only examples where the left–right relation flips between views and is sufficiently separated in the target, ensuring that the correct answer genuinely depends on adopting the target viewpoint.

\paragraph{Instruction Generation.}
For each source-target pair, we convert the annotations into instruction–answer examples to be input to MLLMs. 
We render task-specific visual markers: alphabet labels for \texttt{ViewBench-Text}, geometric symbols for \texttt{ViewBench-Shape}, and a single red circular marker for \texttt{ViewBench-Object}.

For \texttt{ViewBench-Text} and \texttt{ViewBench-Shape}, we pose a binary left–right question about the two markers in the \emph{target} view, randomly ordering the options (\eg, \emph{``left, right''} vs.\ \emph{``right, left''}). The ground-truth label is computed deterministically from the $x$-coordinates of the two keypoints in the target image.  
For \texttt{ViewBench-Object}, we instead use a fixed open-ended template (\eg, \emph{``Can you describe the object or feature at the red point?''}) and treat the MLLM’s response on the oracle target image as the reference description.

After applying the full data-processing pipeline, we obtain:
\begin{itemize}
    \item \textbf{571} text questions (\texttt{ViewBench-Text}),
    \item \textbf{744} shape questions (\texttt{ViewBench-Shape}),
    \item \textbf{300} object-description samples (\texttt{ViewBench-Object}),
\end{itemize}
all validated using the target-view oracle and co-visibility constraints.

\subsection{Details on ViewBench-Object Evaluation}
\label{supp_subsec:llm_eval}
As noted in Sec.~4~\ofmain{}, to evaluate MLLM responses on the target-view object description task (\texttt{ViewBench-Object}), we use an LLM (Qwen2.5-14B-Instruct~\cite{bai2025qwen25vl}) as an evaluator, asking it to rate each response on a 1–10 scale. For this, we query the evaluator LLM with the following prompt template:

\begin{tcolorbox}[
  breakable,
  width=\linewidth,
  colback=blue!3!white,
  colframe=blue!60!black,
  boxrule=0.6pt,
  arc=1mm,
  left=1.2mm,
  right=1.2mm,
  top=0.8mm,
  bottom=0.8mm,
]

\ttfamily\small

You are an AI assistant who will help me to evaluate the response given the question and the correct answer.\par
To mark a response, you should output a single integer between 1 and 10 (including 1, 10).\par\medskip

- 10 means that the response is describing the same or similar scene as the answer.\par\medskip
- 1 means that the response is describing a completely different scene from the answer.\par\medskip

Question: \textbf{\{question\}}\par
Answer: \textbf{\{answer\}}\par
Response: \textbf{\{response\}}\par\medskip

Please output in format <score>...</score>.\par
\end{tcolorbox}

\end{document}